\documentclass[10pt]{article}
\usepackage{amsthm,amsmath,amssymb,bbm,bm}
\usepackage{multirow}
\usepackage{xr-hyper}
\usepackage[pdftex]{graphicx}
\usepackage{wrapfig}
\usepackage{CJK}
\usepackage{array}
\usepackage{url}
\usepackage{booktabs}
\usepackage{algorithm, longtable}
\usepackage[dvipsnames]{xcolor}
\usepackage{algorithmic}
\usepackage{placeins}
\usepackage{mathtools}
\usepackage{romannum}
\usepackage{mathrsfs}
\usepackage{dsfont}
\usepackage{natbib}
\usepackage{relsize}
\usepackage{rotating}
\usepackage{enumitem}
\usepackage{setspace}
\usepackage{tikz}
\usepackage{caption}
\usepackage{subcaption}
\usetikzlibrary{shapes,shapes.multipart}
\usetikzlibrary{arrows}
\usepackage[page,header]{appendix}
\usepackage{titletoc}
\usepackage{titling}
\usepackage{multibib}
\usepackage{style_tex}
\newcites{SM}{References}

\usepackage[colorlinks]{hyperref}
\hypersetup{
    colorlinks,
    linkcolor={red},
    filecolor={red}, %<----
    urlcolor={blue},
    citecolor={blue}
}

\makeatletter
\newcommand*{\addFileDependency}[1]{% argument=file name and extension
  \typeout{(#1)}
  \@addtofilelist{#1}
  \IfFileExists{#1}{}{\typeout{No file #1.}}
}
\makeatother

\usepackage{etoolbox}
\newcommand{\zerodisplayskips}{%
  \setlength{\abovedisplayskip}{6.5pt}%
  \setlength{\belowdisplayskip}{6.5pt}%
  \setlength{\abovedisplayshortskip}{6.5pt}%
  \setlength{\belowdisplayshortskip}{6.5pt}}
\appto{\normalsize}{\zerodisplayskips}
\appto{\small}{\zerodisplayskips}
\appto{\footnotesize}{\zerodisplayskips}

\newtheorem{asu}{Assumption}
\usepackage{chngcntr}
\counterwithin{equation}{section}
\definecolor{ao(english)}{rgb}{0.0, 0.5, 0.0}

\theoremstyle{plain}

\newtheorem{thm}{Theorem}[section]
\newtheorem{lemma}{Lemma}[section]
\newtheorem{proposition}{Proposition}[section]
\newtheorem{corollary}{Corollary}[section]

\counterwithin{theorem}{section}

\def\##1\#{\begin{align}#1\end{align}}
\def\$#1\${\begin{align*}#1\end{align*}}

\def\beq#1\eeq{\begin{equation}#1\end{equation}}
\def\baa#1\eaa{\begin{eqnarray}#1\end{eqnarray}}
\def\bal#1\eal{\begin{align}#1\end{align}}

\newcommand{\blind}{1}

\usepackage[margin=1.3in]{geometry}

\DeclareMathOperator*{\argmin}{arg\,min}
\DeclareMathOperator*{\argmax}{arg\,max}
\begin{document}

\setcounter{page}{1}
\pagenumbering{arabic}

%%%%%%%%%%%%%%%%%%%%%%%%%%%%%%%%%%%%%%%%%%%%%%%%%%%%%%%%%%%%%%%%%%%%%%%%%%%%%%

\if1\blind
\title{\normalfont Adaptive Budgeted Data Acquisition for Robust Off-Policy Evaluation from Multiple Loggers}
\author{
Amitakshar Biswas \thanks{Corresponding author: \href{mailto:biswas8@illinois.edu}{biswas8@illinois.edu}}
\quad Ruoqing Zhu
\\[6pt]
\normalsize Department of Statistics, University of Illinois at Urbana-Champaign
}\fi

\if0\blind
{
    \title{\bf Adaptive Budgeted Data Acquisition for Robust Off-Policy Evaluation from Multiple Loggers}
} \fi

\date{}

\maketitle
%\setstretch{1.15}
% \date{}
% \maketitle
% \setcounter{page}{1}
% \pagenumbering{arabic}
%\vspace{-10mm}
\begin{abstract}
We consider off-policy evaluation when additional data can be purchased
from several existing logging policies (or loggers) with heterogeneous costs per observation. While existing
work on multiple loggers presumes a fixed number of observations from each logger, we study how
to allocate a limited budget across them sequentially to evaluate a finite class of target
policies. We propose \textit{Adaptive Greedy}, an algorithm that purchases
observations sequentially from the affordable logger that has the most cost-normalized variance
reduction across the target policies. It turns out that each such purchase is an 'approximate conditional-gradient step' for the
corresponding design problem. Under mild assumptions, we establish that the empirical allocation
converges almost surely to the optimal design at an explicit rate. We also derive the asymptotic normality of the final pooled estimator and show that it attains the semiparametric efficiency bound.
Simulations and a semi-synthetic application based on the AIDS Clinical Trials Group Study 175 show
that Adaptive Greedy outperforms standard allocation strategies, remains accurate if stopped early,
and matches the final accuracy of repeated integer programming at a fraction of its
computational cost.
\end{abstract}

%\newpage
%\spacingset{1.15}
%\doublespacing
\vspace{1em}

\section{Introduction}\label{sec:introduction}

Contextual bandits are widely used in applications where an action must be chosen after observing
information about the current user or environment, including recommendation systems, online content
selection, and personalized interventions \citep{li2010,tewari2017ads}. In such applications a new
policy usually cannot be deployed simply to learn how well it performs, since online experimentation
is costly, risky, and slow. Off-policy evaluation avoids
this by estimating the value of a target policy from data collected under a different behavior
policy \citep{dudik2014,wang2017}.

In practice, the available data need not come from a single behavior policy. Deployed systems are
updated over time, several experiments may run in parallel, and different loggers may use different
decision rules. Methods for combining observations from several loggers generally take the logger
sample sizes as given \citep{agarwal2017,he2019,kallus2021}. In many applications, however, more
data can still be collected, but only from loggers that already exist, and these loggers differ both
in the actions they tend to take and in their per-observation prices. An inexpensive logger may
concentrate on a few actions, while a more costly logger may provide the coverage that a target
policy needs. The practical question is where the remaining observations should be purchased when
the reward distribution is not yet known. We study a fixed set of logging policies
$\mu_1,\ldots,\mu_K$ with known costs $c_1,\ldots,c_K$. The policies cannot be redesigned, and
acquisition chooses only which logger to sample from next, subject to a finite budget.

This problem differs from behavior-policy design, in which the logging or augmentation policy is
itself chosen to reduce the variance of a downstream estimator
\citep{hanna2024,tucker2023,mukherjee2024speed}. When the loggers are fixed, an allocation is a count vector \(N=(N_1,\ldots,N_K)\in\mathbb Z_+^K\) with \(c^\top N\le B\), where \(N_k\) is the number of observations purchased from logger \(k\) and $B$ is the total budget. Each purchase increments one coordinate of \(N\) and bears the corresponding logger cost.

A second feature is that several target policies may have to be evaluated from the same sample. We
consider a finite target class $\Pi=\{\pi_1,\ldots,\pi_M\}$ and seek a design that controls the largest
target-specific uncertainty over that class. The criterion retains each policy's target-specific
efficient variance and minimizes its maximum, so the objective can change as the target
attaining that maximum changes. Prior work on evaluating multiple policies jointly provides related
motivation, although it studies different sampling settings and objectives
\citep{dann2023,chen2025caesar}.

The criterion depends on the conditional reward mean and variance, which are unknown at the outset and must be estimated from the observations acquired along the way. We propose a sequential rule, which we
call \textit{Adaptive Greedy}, that re-estimates the criterion as data accrue and at each step purchases from
the affordable logger with the largest estimated reduction in the worst-target criterion per unit of
cost. An additional feature is that the rule does not plan toward a declared final budget. The budget enters only through which loggers remain affordable and when acquisition stops. This is useful if the available budget is unexpectedly reduced or the process must stop early, since the data already collected were chosen to improve the current design rather than to follow a plan optimized for a larger final budget. Planners that repeatedly solve for count vectors at a declared final budget behave differently, since their early purchases are chosen with that budget in mind.

For a fixed logger composition, weighting each observation by the overall mixture of the logging policies has no larger asymptotic variance than weighting it only by the logger that generated it. We therefore combine all acquired observations using this pooled estimator. Under adaptive data collection, however, the final logger composition depends on the rewards observed along the way, so the pooled denominator is itself random. We show that as the design converges, this random denominator can be replaced at first order by its deterministic limiting value. As a result, the pooled estimator remains asymptotically normal and its large-sample variance is determined by the allocation reached by the algorithm.

Our contributions are threefold. First, we formulate budgeted off-policy evaluation as a sequential allocation problem among loggers with heterogeneous costs and propose the Adaptive Greedy algorithm. Second, assuming overlap only for the pooled behavior distribution at the optimal design, rather than for every logger, we show that the resulting design converges almost surely to the continuous minimax design. Under stronger conditions on the plug-in estimates, we also derive an explicit polynomial rate. Third, we establish joint asymptotic normality and consistent covariance estimation for the pooled estimator, even though the logger choices depend on previously observed rewards. The variance advantage of pooled over logger-specific weighting is known for fixed designs \citep{agarwal2017,kallus2021}. Our result shows that pooled inference remains valid after adaptive data collection under the stated conditions.

\subsection{Related Work}\label{sec:related-work}

Off-policy evaluation from several loggers has so far been studied with the logger sample sizes
given in advance. \citet{agarwal2017} observe that weighting each observation by the pooled behavior
distribution never increases variance relative to weighting it by the logger that generated it.
\citet{he2019} study off-policy learning from data collected by multiple loggers, and
\citet{lai2020mixture} choose variance-minimizing weights for combining estimators computed from
several behavior policies. \citet{kallus2021}
characterize the semiparametric efficiency bound for the fixed-stratified multi-logger model and
construct an estimator that attains it. This pooled bound, viewed as a function of the logger
composition, is the population objective of our design problem. Related mixture-design problems have also been studied in importance sampling.
\citet{heowen2014} establish convexity of variance objectives for mixtures of
fixed proposal distributions and discuss minimizing the maximum variance
when estimating multiple integrals from shared samples. Our setting
instead concerns sequential purchases from fixed contextual-bandit loggers
with heterogeneous costs, where allocation decisions depend on the
observations acquired so far.

A second line of work designs or learns the data-collection policy itself for downstream
evaluation. Examples include behavior-policy search \citep{hanna2024,liu2024behavior}, experimental
design in linear heteroscedastic bandits \citep{mukherjee2024speed}, safe or robust data-collection
designs \citep{zhu2022,mukherjee2024saver,liu2025safe,chen2026robust}, and the optimization of
logging probabilities \citep{douglas2026}. In contextual bandits, \citet{tucker2023} design an
augmentation logging policy for an existing dataset and evaluate with a balanced estimator that
divides by the pooled behavior distribution of the original and augmentation data. In all of these
works the behavior or augmentation policy is a design variable, whereas in our setting the loggers
are fixed and only the number of purchases from each can be chosen.

Evaluating several target policies from shared data is a closely related problem. \citet{dann2023}
study high-confidence evaluation of multiple policies in episodic reinforcement learning,
\citet{jain2024} adaptively learn a behavior policy for evaluating several general value functions,
and \citet{chen2025caesar} optimize an offline sampling distribution for several policies in an
episodic MDP. \citet{liu2025multi} and \citet{russo2025} also study data collection when there are
several evaluation targets. These works motivate designing for a finite target class jointly, which
in our setting takes the form of a minimax criterion over the target-specific pooled efficiency
bounds.

Inference from data collected by outcome-adaptive rules has been developed for batched bandits,
adaptive experiments, and off-policy evaluation from contextual bandits
\citep{zhang2020,hadad2021,zhan2021,bibaut2021,waudbysmith2024}. In these works the weights and
normalizations entering the estimator are predictable, meaning that they are known when each
observation is collected, and martingale arguments apply to the resulting estimating equations. The
pooled estimator studied here divides by the final logger composition, which depends on rewards
observed along the acquisition path and is known only after acquisition ends. Design convergence
allows us to replace this random final mixture by its deterministic limit.

The cost-normalized feasible set and the minimax variance criterion connect the problem to classical
optimal design \citep{fedorov1972,pukelsheim2006}. The sequential update has a conditional-gradient
interpretation related to the Fedorov--Wynn design algorithms \citep{wynn1970,chernoff1959}, with
vertices fixed by the available loggers and a step size set by the cost of the logger purchased.
Section~\ref{sec:convergence} develops this connection.

\section{Preliminaries}\label{sec:prelim}

\subsection{Problem Setup}\label{sec:setup}

Let $\mathcal{X}$ be a measurable context space and let $\mathcal{A}$ be a finite action space. There are $K$ existing logging policies (or loggers) 
$\mu_1(a \mid x), \ldots, \mu_K(a \mid x)$, and acquiring one observation from logger $k$ costs
$c_k > 0$. If an observation is purchased from logger $k$, it is generated as
\begin{equation}\label{eq:dgp}
X \sim P_X, \qquad A \sim \mu_k(\cdot \mid X), \qquad R \sim P_R(\cdot \mid X, A).
\end{equation}
The context distribution $P_X$ and conditional reward distribution $P_R$ are common across loggers.
Write $q(x, a) = \mathbb{E}(R \mid X = x, A = a)$ and
$\sigma^2(x, a) = \mathrm{Var}(R \mid X = x, A = a)$ for the conditional reward mean and variance. Let
$\Pi=\{\pi_1,\ldots,\pi_M\}$ be the finite target policy class. Our estimand is the vector of policy values
$J = (J_1, \ldots, J_M)^{\top}$, which we estimate after acquisition is complete, where
$J_j = \mathbb{E}_X \{ q_j(X) \}$ and $q_j(x) = \sum_{a \in \mathcal{A}} \pi_j(a \mid x) q(x, a)$.
Only context-action pairs used by some target policy affect the policy values. Let $\mathcal S=\{(x,a):\pi_j(a\mid x)>0\text{ for some }j\}$ be the union of the target-policy
supports. For a function $f$ of $(x,a)$, we write $\operatorname*{ess\,inf}_{\mathcal S}f$ for the
infimum of $f(x,a)$ over $(x,a)\in\mathcal S$, after removing a $P_X$-null set of contexts.

Data are acquired sequentially, and each logger is chosen using only the data observed before that
choice. For observation $i$, let
$K_i\in\{1,\ldots,K\}$ denote the selected logger, and $A_i$ for the
contextual-bandit action generated by that logger. Let
\[
\mathcal F_i
=
\sigma\{K_1,X_1,A_1,R_1,\ldots,K_i,X_i,A_i,R_i\},
\]
with $\mathcal F_0$ trivial, be the history till time $i$. If external randomization is used to select a logger
(e.g., to break a tie between two or more potential loggers), the filtration is assumed to include that randomization before
the corresponding logger choice. Thus $K_i$ is $\mathcal F_{i-1}$-measurable. Conditional on $\mathcal F_{i-1}$ and
$K_i=k$, the observation $(X_i,A_i,R_i)$ is generated according to
\eqref{eq:dgp}. We impose the following structural assumption, which is standard in off-policy evaluation with multiple loggers; see \citet{dudik2014} and \citet{kallus2021}. 

\begin{asu}\label{ass:known}
The logging probabilities $\mu_1, \ldots, \mu_K$ and the costs $c_1, \ldots, c_K$ are known,
$0 < c_{\min} = \min_k c_k \le \max_k c_k = c_{\max} < \infty$, and $|R| \le R_{\max} < \infty$
almost surely.
\end{asu}

With total budget $B$, acquisition continues until no logger remains affordable. Let
$N_{t,k}=\sum_{i=1}^{t}\mathbf{1}\{K_i=k\}$ count the number of observations acquired from
logger $k$. Let $C_t=\sum_{k=1}^{K}c_kN_{t,k}$ denote the expenditure after $t$ observations. Write
$N_t=(N_{t,1},\ldots,N_{t,K})^\top$, so that $\sum_kN_{t,k}=t$. Acquisition stops at
\begin{equation*}
T_B = \min \{ t \ge 0 : c_{\min} > B - C_t \},
\end{equation*}
which is a stopping time with respect to $(\mathcal{F}_t)$. It is easy to see that $T_B$ is bounded between $\lfloor (B - c_{\min}) / c_{\max} \rfloor$ and
$\lceil B / c_{\min} \rceil$, and in particular $T_B \to \infty$ as $B \to \infty$. 

\subsection{Fixed Logger Sizes}\label{sec:pooled}

The purpose of this subsection is to obtain the variance criterion that we will later use for acquisition. Here we fix the sample size $t$ and the sequence of loggers \(k_1,\ldots,k_t\) with $k_i \in \{1,\ldots,K\}$. Let \(N_k=N_{t,k}\) be the resulting number of observations from logger \(k\), and \(N=(N_1,\ldots,N_K)^\top\). Thus \(\sum_k N_k=t\), the counts are deterministic, and no acquisition decision depends on the observed data. This is the fixed-stratified setting studied by \citet{kallus2021}.

Since all loggers share $P_X$ and $P_R$, we can weight every observation by the mixture of the
logging policies, as if the combined sample came from that mixture. This weighting never gives a
larger variance than weighting each observation by the logger that generated it
\citep{agarwal2017,kallus2021}. The mixture is the pooled behavior distribution
\begin{equation}
\bar{\mu}_N(a \mid x) = \frac{1}{t} \sum_{k=1}^{K} N_k \mu_k(a \mid x).
\label{eq:pooled-behavior}
\end{equation}
We define the pooled augmented inverse propensity weighted (AIPW) estimator of $J_j$ as
\begin{equation}\label{eq:pooled-estimator}
\widetilde{J}_{t,j}
= \frac{1}{t} \sum_{i=1}^{t}
\left[
q_j(X_i) + \frac{\pi_j(A_i \mid X_i)}{\bar{\mu}_N(A_i \mid X_i)}
\{ R_i - q(X_i, A_i) \}
\right].
\end{equation}
\citet[Theorem~1]{kallus2021} show that this weighting,
together with the true $q$, minimizes
variance within the estimator class in the fixed-stratified
setting.

For a fixed logger sequence $k_1,\ldots,k_t$ with $\operatorname*{ess\,inf}_{\mathcal S}\bar\mu_N>0$,
the pooled estimator is unbiased, $\mathbb E(\widetilde J_{t,j})=J_j$, and its
fixed-design variance is
\begin{equation}\label{eq:fixed-variance}
\operatorname{Var}(\widetilde J_{t,j})
=
\frac{\operatorname{Var}\{q_j(X)\}}{t}
+
\mathbb E_X
\sum_{a\in\mathcal A}
\frac{\pi_j(a\mid X)^2\sigma^2(X,a)}
{\sum_{k=1}^K N_k\mu_k(a\mid X)}.
\end{equation}
This variance formula follows from
\citet[Theorem~1]{kallus2021}. For notational convenience, let
$V_j = \mathrm{Var}\{ q_j(X) \}$ and 
$W_j(x, a) = \pi_j(a \mid x)^2 \sigma^2(x, a)$. Note that the variance does not depend on the order in which the loggers appear but only through the
count vector $N$. Write $\theta_k=N_k/t$ for the sample fractions, which lie in the simplex
$\mathcal S_K=\{\theta\in\mathbb R_+^K:\mathbf 1^\top\theta=1\}$, so that $\bar\mu_N=m_\theta$ with
$m_\theta(a\mid x)=\sum_{k=1}^K\theta_k\mu_k(a\mid x)$. Then \eqref{eq:fixed-variance} becomes
$\mathrm{Var}( \widetilde{J}_{t,j} ) = A_j(\theta) / t$ with
\begin{equation}\label{eq:Aj}
A_j(\theta)
= V_j + \mathbb{E}_X \sum_{a \in \mathcal{A}}
\frac{W_j(X, a)}{m_{\theta}(a \mid X)} .
\end{equation}
Thus, for a fixed logger composition $\theta$,
a smaller $A_j(\theta)$ means a smaller variance
for estimating $J_j$.

\citet[Theorem~4]{kallus2021} derive the efficient influence function and the semiparametric
efficiency bound for the fixed-stratified multi-logger model. Their bound for $\pi_j$ is exactly $A_j(\theta)$. Hence $A_j(\theta)$ is the smallest regular
asymptotic variance attainable for estimating $J_j$ under a fixed logger composition $\theta$ with $\operatorname*{ess\,inf}_{\mathcal S}m_\theta>0$.\footnote{\citet[Theorem~5]{kallus2021} further show that the efficiency bound is unchanged when
the logging policies are unknown.} This efficiency result is for the fixed-stratified setting and does not automatically extend to our
adaptive procedure, where the final logger composition depends on the observed data. We nevertheless use $A_j(\theta)$ to construct the acquisition objective,
and we establish valid post-acquisition inference in Section~\ref{sec:adaptive-inference} for the pooled estimator computed from
the acquired sample.

\section{Methodology}\label{sec:method}

In this section, we develop our sequential method, \textit{Adaptive Greedy}. We break it into three parts: Section~\ref{sec:criterion} defines the design criterion that we eventually minimize, Section~\ref{sec:selection} describes our greedy approach to selecting the loggers adaptively and how we estimate the unknown quantities, and finally, we state the complete algorithm in Section~\ref{sec:algorithm}. All proofs and derivations are provided in the appendices.

\subsection{Design Criterion}\label{sec:criterion}

To compare allocations and decide where the next observation is most useful, we use the variance formula \eqref{eq:fixed-variance} as our criterion, with constants added so that it is finite at $N=0$. Let $N=(N_1,\ldots,N_K)^\top$ be a count vector. For target policy $\pi_j$, we define
\begin{equation}\label{eq:Gj}
G_j(N)
=
\frac{V_j}{\mathbf 1^\top N+\eta_0}
+
\mathbb E_X
\sum_{a\in\mathcal A}
\frac{W_j(X,a)}{D_N(X,a)} ,
\end{equation}
where $D_N(x,a) = \eta+\sum_{k=1}^K N_k\mu_k(a\mid x)$ and $\eta,\eta_0>0$. In our experiments we use the simple
choice $\eta=\eta_0=1$ throughout. Since every target in $\Pi$ should be estimated accurately, we control the worst target-specific variance $G(N)=\max_{1\le j\le M}G_j(N)$. Other choices can also be considered, for example $M^{-1}\sum_{j=1}^M G_j(N)$, but the average can be small while one target is estimated poorly. Note that a new observation reduces the variance if it adds coverage for regions where $D_N(x,a)$ is small and $W_j(X,a)$ is large.

To compare allocations independently of how much has been spent, we normalize the counts by their cost. Define $z_N = N/C_N$, where $C_N = c^\top N$ is the total expenditure. Thus $z_{N,k}$ is the number of
observations purchased from logger $k$ per unit of spent budget. Since $c^\top z_N=1$, let
$\mathcal C = \{z\in\mathbb R_+^K:c^\top z=1\}$. Multiplying $G_j(N)$ by the spent budget $C_N$ removes the
decrease in variance that comes simply from spending more money. For a large spent budget,
$C_NG_j(N)\approx\mathcal G_j(z_N)$, where $\mathcal G_j, \mathcal G : \mathcal{C} \to \mathbb{R} \cup \{\infty\}$ are
\begin{equation}\label{eq:Fj}
\mathcal G_j(z)
=
\frac{V_j}{\mathbf 1^\top z}
+
\mathbb E_X
\sum_{a\in\mathcal A}
\frac{W_j(X,a)}
{\boldsymbol{\mu}(X,a)^\top z},
\qquad
\mathcal G(z)=\max_{1\le j\le M}\mathcal G_j(z),
\qquad
z\in\mathcal C,
\end{equation}
with
$\boldsymbol{\mu}(x,a)
=(\mu_1(a\mid x),\ldots,\mu_K(a\mid x))^\top$. For any fixed positive lower bound on the pooled coverage, Lemma~\ref{lem:budget-expansion} shows that there is a finite constant $K_G$ such that
\[
\left|
C_NG_j(N)-\mathcal G_j(z_N)
\right|
\le
\frac{K_G}{C_N}
\]
for every nonzero count vector with such coverage. We will work with $G_j(N)$ for our algorithm as it is more convenient and we will use $\mathcal G_j$ when we talk about convergence in Section \ref{sec:theory}.

We now define the optimal allocation we would like our algorithm to converge to. Since each $\mathcal G_j$ is convex in $\mathcal{C}$ whenever the denominator is positive (as \(V_j\ge0\), \(W_j(X,a)\ge0\), and \(x\mapsto 1/x\) is convex for \(x>0\)), $\mathcal G=\max_j\mathcal G_j$ is also convex. If $V_j$ and $W_j$ are known, define
\begin{equation}\label{eq:xstar}
z^\star
\in
\arg\min_{v\in\mathcal C}\mathcal G(v),
\qquad
\theta^\star
=
\frac{z^\star}{\mathbf 1^\top z^\star}.
\end{equation}
Thus $z^\star$ gives the allocation per unit of budget that minimizes the
largest target-specific variance, and $\theta^\star$ gives the corresponding fractions. For a fixed logger composition,
$A_j(\theta)$ is also the semiparametric efficiency bound discussed in
Section~\ref{sec:pooled}, so this choice has the same fixed-design
interpretation after accounting for the logger costs. We discuss the uniqueness of $z^\star$ in Section \ref{sec:convergence}.

The fixed-design variance \eqref{eq:fixed-variance} is finite under our bounded-reward assumption
when the pooled probabilities are bounded away from zero on the target-policy supports, even if
individual loggers are not. We measure this coverage by
\begin{equation}\label{eq:min-coverage}
\kappa(\theta)
=
\operatorname*{ess\,inf}_{(x,a)\in\mathcal S}
m_\theta(a\mid x).
\end{equation}
When the contexts take finitely many values, for example finitely many subgroups, $\kappa(\theta)$ is a minimum of finitely many linear functions of $\theta$ and can be computed exactly. For general context spaces, forced sampling can instead use a known lower bound on $\kappa(\theta)$, for example a finite minimum of nonnegative weighted sums of the $\theta_k$ (Appendix~\ref{app:coverage-computation}). We impose this condition only at an optimal design.

\begin{asu}\label{ass:pooled}
There is a minimizer $z^\star$ in \eqref{eq:xstar} whose composition
$\theta^\star$ satisfies
$\kappa(\theta^\star)\ge \kappa_\star$
for some $\kappa_\star>0$.
\end{asu}

This assumption allows individual loggers to assign zero probability
to target-relevant actions, provided their mixture at the optimal
design has sufficient coverage. \citet{agarwal2017} note that their balanced estimator
needs only that the average logging policy covers the target, and \citet{kallus2021} distinguish
this pooled overlap from the stronger requirement that every logger covers the target, which
logger-specific weights need \citep[see also][]{he2019}. Per-logger lower bounds remain common in
multi-logger evaluation \citep{lai2020mixture,gao2023hope}. If each logger is deployed with $\mu_k(a\mid x)\ge\mu_0$ on $\mathcal S$ for a constant $\mu_0>0$,
as \citet{agarwal2017} suggest for ensuring support, then
$m_\theta(a\mid x)=\sum_k\theta_k\mu_k(a\mid x)\ge\mu_0\sum_k\theta_k=\mu_0$ for every composition
$\theta$, so $\kappa(\theta)\ge\mu_0$ and Assumption~\ref{ass:pooled} holds with $\kappa_\star=\mu_0$. Assumption~\ref{ass:pooled} is a
quantitative form of pooled overlap, imposed only at $\theta^\star$, and individual loggers may give
probability zero to actions that the targets use. Since $\theta_k^\star\le1$, it implies
$\sum_k\mu_k(a\mid x)\ge \kappa_\star$ on $\mathcal S$, so the uniform mixture of the loggers also covers
the targets.

\subsection{Sequential Selection Rule}\label{sec:selection}

To choose the next logger, we compare the reduction in the worst-target variance $G$ that one more
observation from each logger would bring, per unit of cost. Initially, we start with budget $B$ and no data. At time $t$, the remaining budget is $B - C_t$ and we choose the next logger from the set $
\mathcal K_t
=
\{k:c_k\le B-C_t\}.$ Now, given the current allocation $N_t$, we want to purchase the next logger from $\mathcal K_t$ that provides the largest reduction in $G$. Therefore, for any count vector $N$, define the reduction from purchasing one observation
from logger $k$ by
\begin{equation}\label{eq:robust-gain}
\Delta_k(N)
=
G(N)-G(N+e_k).
\end{equation}
If the reward moments were known, the optimal allocation
could be obtained by solving
\begin{equation}\label{eq:static}
\min_{\substack{N\in\mathbb Z_+^K\\c^\top N\le B}}G(N).
\end{equation}
However, solving this integer optimization problem is
computationally expensive for large budgets. Moreover,
$V_j$ and $W_j$ are unknown in our setting.
We therefore consider the simpler greedy rule
\begin{equation}\label{eq:oracle-greedy}
K_{t+1}^{\mathrm{greedy}}
\in
\argmax_{k\in\mathcal K_t}
\frac{\Delta_k(N_t)}{c_k},
\end{equation}
which chooses the affordable logger giving the largest immediate reduction in
$G$ per unit cost.

The rule \eqref{eq:oracle-greedy} cannot be used directly, for two reasons. First, the reductions
are small and unknown. When the pooled coverage is bounded below, $G_j(N_t)$ is of order $t^{-1}$
while the target-specific reduction $\Delta_{j,k}(N_t)=G_j(N_t)-G_j(N_t+e_k)$ is of order $t^{-2}$
(Appendix~\ref{app:order}), so the reductions behave like derivatives of the criterion. They depend
on the unknown $V_j$ and $W_j$, and the rule must still rank them correctly across loggers. Second,
$G$ is a maximum over targets. Its one-step change $\Delta_k(N)$ is not a combination of the
target-specific changes, and $G$ is not smooth where the maximizing target switches.

We handle the second difficulty by replacing the maximum with a log-sum-exp approximation
\citep{nesterov2005}. This puts exponential weights on the target-specific reductions and
introduces a bias of at most $\rho\log M$, which vanishes as the smoothing parameter $\rho_t$
decreases. We handle the first by plugging estimates of the reward mean and variance into the
closed form of the reductions. Our theory needs these estimates to be accurate on the $t^{-2}$ scale
of the reductions, which is Assumption~\ref{ass:nuisance} in Section~\ref{sec:convergence}. The two
fixes interact: as $\rho_t$ decreases, the weights become more sensitive to estimation error, so
$\rho_t$ must decrease slowly relative to the estimation error (see
Theorem~\ref{thm:design-optimality}).

\paragraph{Estimation.} Writing $T=\mathbf 1^\top N$, the target-specific reduction has the closed
form (see Appendix~\ref{app:gain} for derivation)
\begin{equation}\label{eq:gain}
\Delta_{j,k}(N)
=
G_j(N)-G_j(N+e_k)
=
\mathbb E_X\sum_{a\in\mathcal A}
\frac{
W_j(X,a)\mu_k(a\mid X)
}{
D_N(X,a)\{D_N(X,a)+\mu_k(a\mid X)\}
}
+
\frac{V_j}{
(T+\eta_0)(T+1+\eta_0)
}.
\end{equation}
Let $\widehat G_{j,t}$ and $\widehat\Delta_{j,k,t}$ denote $G_j(N_t)$ and $\Delta_{j,k}(N_t)$ with $V_j$
and $W_j$ computed from estimates $\widehat q_t$ and $\widehat\sigma_t^2$ of the reward mean and variance,
and with $P_X$ replaced by the empirical distribution of the observed contexts
(Appendix~\ref{app:sufficient}). Any regression estimators of the reward mean and variance can be
used, and our experiments use ridge regression (Appendix~\ref{app-implementation}).
Lemma~\ref{lem:plugin} gives conditions on these estimators under which
Assumption~\ref{ass:nuisance} holds, and Corollary~\ref{cor:finite-contexts} verifies them for finite
context spaces.

\paragraph{Smoothing.} To address the nonsmoothness of the worst-target criterion
$\mathcal G(v)=\max_{1\le j\le M}\mathcal G_j(v)$,
we use a log-sum-exp approximation \citep{nesterov2005}. For $\rho>0$, define
\begin{equation}\label{eq:Phi}
\Phi_\rho(v)
=
\rho\log\left\{\sum_{j=1}^M\exp\bigl(\mathcal G_j(v)/\rho\bigr)\right\},
\qquad
\mathcal G(v)\le\Phi_\rho(v)\le \mathcal G(v)+\rho\log M .
\end{equation}
Hence $\Phi_\rho(v)\to \mathcal G(v)$ as $\rho\downarrow0$. Its gradient is
$
\nabla\Phi_\rho(v)
=
\sum_{j=1}^M
\omega_j(v;\rho)\nabla \mathcal G_j(v)$, with exponential weights
$
\omega_j(v;\rho)
=
\exp\{\mathcal G_j(v)/\rho\}/
\sum_{\ell=1}^M\exp\{\mathcal G_\ell(v)/\rho\}.
$
Therefore, policies with larger values of $\mathcal G_j(v)$ receive larger weights.
We now relate the gradient to the one-step gains. Recall that
$C_NG_j(N)\approx\mathcal G_j(z_N)$ for large budgets.
Hence, we have
$C_N^2\Delta_{j,k}(N)\approx-\partial_k\mathcal G_j(z_N)$.
Using the expression for $\nabla\Phi_\rho$, this gives
\[
-\left\langle
\nabla\Phi_\rho(z_N),\frac{e_k}{c_k}
\right\rangle
\approx
\frac{C_N^2}{c_k}
\sum_{j=1}^M
\omega_j(z_N;\rho)\Delta_{j,k}(N).
\]
Since $C_N^2$ is common to all loggers, this motivates
using the weighted gains for selecting the next logger. Lemmas~\ref{lem:budget-expansion} and
\ref{lem:oracle-LMO} make this idea precise. So that the bias $\rho_t\log M$ vanishes, we let the
smoothing parameter decrease with expenditure, $\rho_t=\rho_0C_t^{-\alpha}$ with $\rho_0>0$ and
$0<\alpha<1$, and estimate the weights by
$
\widehat\omega_{j,t}
=
\exp(C_t\widehat G_{j,t}/\rho_t)/
\sum_{\ell=1}^M\exp(C_t\widehat G_{\ell,t}/\rho_t).
$
The weights therefore become increasingly
concentrated on the policies with the largest estimated
criterion values. We define the acquisition index for logger $k$ as
\begin{equation}\label{eq:score}
\widehat S_{t,k}
=
\frac{1}{c_k}
\sum_{j=1}^M
\widehat\omega_{j,t}\widehat\Delta_{j,k,t}.
\end{equation}
The rule selects the affordable logger with the largest index. In our experiments we use
$\rho_0=0.5$ and $\alpha=0.4$. Sensitivity to these choices is
reported in Appendix~\ref{app-parameter-sensitivity}. The additional
condition on $\alpha$ needed for the convergence result is given in
Section~\ref{sec:convergence}.

\subsection{Algorithm}\label{sec:algorithm}

In this section we state our final algorithm. Combining the previous subsections, the main rule is to select the affordable logger $k$ at step $t$ with the largest index $\widehat S_{t,k}$ except certain times as stated below. The full procedure is stated in Algorithm~\ref{alg:acquisition}. The initialization gives each
logger a small number of observations before selection begins. The refit interval $m$
avoids refitting the nuisance models after every acquisition.

\paragraph{Forced sampling.} Although Assumption \ref{ass:pooled} controls coverage once the algorithm is near the optimal allocation, it does not control coverage early in the acquisition. If the early
estimates undervalue the loggers that cover some target actions, the rule may rarely purchase from them, and
the estimates for those actions are then slow to improve. Forced sampling of under-sampled arms or strata
is a standard remedy in adaptive allocation \citep{antos2010,etore2010,garivier2016}, and adaptive
designs often keep assignment probabilities above a floor that vanishes over time
\citep{hadad2021,bibaut2021,dai2023}. We use forced sampling of this kind, acting on pooled coverage.

Recall that $\kappa(\theta)$ is the essential infimum of $m_\theta(a\mid x)$ over the
context-action pairs used by at least one target policy. If $\kappa(\theta_t)$ falls below a
threshold $\chi_t$ that vanishes over time, the algorithm purchases an observation from the logger
with the smallest count. Otherwise, it selects the logger using the usual greedy rule. We take
\[
\chi_t=\chi_0t^{-\upsilon},
\qquad
0<\upsilon<1,
\]
where $0<\chi_0\le\frac{\kappa(\mathbf 1/K)}{8}.$
Our convergence results require
\begin{equation}\label{eq:tuning}
0<\alpha<1,
\qquad
0<\upsilon<\frac{1-\alpha}{4},
\qquad
0<\chi_0\le\frac{\kappa(\mathbf 1/K)}{8},
\qquad
\frac{n_0}{K}\ge4 .
\end{equation}

Lemma~\ref{lem:coverage} shows that forced sampling maintains
$\kappa(\theta_t)\ge\chi_t/2$ while all loggers are affordable.
It also shows that the number of forced purchases up to time
$t$ is $O(t^{1-\upsilon})$. Under the conditions of
Theorem~\ref{thm:design-optimality}, the allocation converges
to $\theta^\star$ and forced sampling eventually stops.

If forced sampling uses a lower bound $\widetilde\kappa\le\kappa$ in place of $\kappa$, the same
guarantees hold under the conditions given in Appendix~\ref{app:coverage-computation}.

\begin{algorithm}[t]
\caption{Adaptive Greedy for budgeted data acquisition from multiple loggers}
\label{alg:acquisition}
\begin{algorithmic}[1]
\STATE \textbf{Input}: loggers $\mu_1,\ldots,\mu_K$,
costs $c_1,\ldots,c_K$, target policies
$\Pi=\{\pi_1,\ldots,\pi_M\}$, budget $B$,
constants $\rho_0>0$ and $\alpha\in(0,1)$,
forced-sampling constants
$\chi_0\in(0,\kappa(\mathbf 1/K)/8]$
and $\upsilon\in(0,1)$,
refit interval $m$, total initialization size $n_0$,
assumed divisible by $K$ with $n_0/K\ge4$.

\STATE Acquire $n_0/K$ initial observations from each
logger, and set $t=n_0$.

\WHILE{$\mathcal K_t$ is nonempty}
    \IF{$t-n_0$ is a multiple of $m$}
        \STATE Update $\widehat q_t$ and
        $\widehat\sigma_t^2$ using the first $t$
        observations, and form $\widehat V_{j,t}$
        and $\widehat W_{j,t}$ as in
        \eqref{eq:nuisance-estimates}.
    \ENDIF

    \IF{$\kappa(N_t/t)<\chi_0t^{-\upsilon}$}
        \STATE Select
        $K_{t+1}\in
        \argmin_{k\in\mathcal K_t}N_{t,k}$
        (forced sampling).
    \ELSE
        \STATE Compute $\widehat G_{j,t}$ for $j=1,\ldots,M$.
        \STATE Set $\rho_t=\rho_0C_t^{-\alpha}$
        and compute $\widehat\omega_{j,t}$.
        \STATE Compute $\widehat\Delta_{j,k,t}$
        from \eqref{eq:gainhat} and
        $\widehat S_{t,k}$ from \eqref{eq:score}
        for all $k\in\mathcal K_t$.
        \STATE Select
        $K_{t+1}\in
        \argmax_{k\in\mathcal K_t}\widehat S_{t,k}$.
    \ENDIF

    \STATE Acquire one observation from logger $K_{t+1}$,
    and set
    $N_{t+1}=N_t+e_{K_{t+1}}$,
    $C_{t+1}=C_t+c_{K_{t+1}}$,
    and $t\leftarrow t+1$.
\ENDWHILE

\STATE Set $T_B=t$, fit the final reward regression
$\widehat q_B$ using the completed sample, and set
$\widehat\theta_B=N_{T_B}/T_B$.
\STATE Compute the final estimator $\widehat J_B$
from \eqref{eq:terminal}.
\STATE \textbf{Output}: the acquired sample and
$\widehat J_B$.
\end{algorithmic}
\end{algorithm}

\paragraph{Final estimation.}
When acquisition ends, we estimate $J$ using the entire
acquired sample and the realized logger composition
$\widehat\theta_B=N_{T_B}/T_B$.
The final pooled estimator replaces $q$ in \eqref{eq:pooled-estimator} by a reward regression
$\widehat q_B$ fitted on that sample, truncated to $[-R_{\max},R_{\max}]$:
\begin{equation}\label{eq:terminal}
\widehat J_{B,j}
=
\frac{1}{T_B}\sum_{i=1}^{T_B}
\left[
\widehat q_{B,j}(X_i)
+
\frac{\pi_j(A_i\mid X_i)}
{m_{\widehat\theta_B}(A_i\mid X_i)}
\{R_i-\widehat q_B(X_i,A_i)\}
\right],
\qquad j=1,\ldots,M ,
\end{equation}
where $\widehat q_{B,j}(x)=\sum_{a\in\mathcal A}\pi_j(a\mid x)\widehat q_B(x,a)$.
Unlike the fixed-design setting, the final logger composition
$\widehat\theta_B$ is random and depends on the adaptive selection
process. We study its effect on inference in Section~\ref{sec:adaptive-inference}.

\section{Theory}\label{sec:theory}

We divide this section into two subsections. We first prove that \textit{Adaptive Greedy} actually converges to the optimal minimax design under mild assumptions in Section \ref{sec:convergence}. We then establish the asymptotic normality of our estimator in Section \ref{sec:adaptive-inference}. All proofs are in
Appendices~\ref{app:theory}, \ref{app:inference}, and \ref{app:technical}.

\subsection{Convergence and Asymptotics}\label{sec:convergence}

In Section~\ref{sec:method}, we defined $z^\star$ (equation \eqref{eq:xstar}) which minimizes the criterion $\mathcal G$. In this subsection, we are interested in two questions. First is when $z^\star$ is unique and if it is, whether $z_{T_B}$ converges to $z^\star$ as $B$ increases. The second is the rate. Theorem~\ref{thm:design-optimality} answers both.

The main idea is that choosing the logger with the largest gain per unit cost amounts to choosing the vertex $e_k/c_k$ of $\mathcal C$ that minimizes the linear
approximation of the criterion at $z_t$. When the algorithm purchases one observation from logger $k$, $z_t=N_t/C_t$ changes according to
\begin{equation}\label{eq:FW-update}
z_{t+1}
=
(1-\gamma_t)z_t
+
\gamma_t\frac{e_k}{c_k},
\qquad
\gamma_t=\frac{c_k}{C_t+c_k},
\end{equation}
so the purchase moves $z_t$ toward that vertex by the step $\gamma_t$. This is similar to the Frank--Wolfe, or
conditional-gradient, method \citep{frank1956,jaggi2013}, used in optimal design by
\citet{fedorov1972} and \citet{wynn1970}. Such steps converge to the minimizer of a smooth convex
function when their lengths tend to zero and their sum diverges, as $\gamma_t$, being of order
$C_t^{-1}$, does.

Strong convexity of $\mathcal G$ serves two purposes. If $\mathcal G$ is $m$-strongly convex, its minimizer
$z^\star$ is unique and $\mathcal G(z)-\mathcal G(z^\star)\ge\tfrac m2\|z-z^\star\|_2^2$ for every $z\in\mathcal C$, so
convergence of $\mathcal G(z_t)$ to $\mathcal G(z^\star)$ forces $z_t$ to converge to $z^\star$. Recall that a function $f$
on $\mathcal C$ is $m$-strongly convex if $f(z)-\tfrac m2\|z\|_2^2$ is convex; for twice differentiable $f$
this holds exactly when, along every segment in $\mathcal C$ with direction $h$, the second derivative of
$f$ is at least $m\|h\|_2^2$. Any feasible direction $h$ satisfies $c^\top h=0$, and on such directions
$h^\top\nabla^2\mathcal G_j(z)h\ge 2c_{\min}^3h^\top Q_jh$ (see the proof of
Proposition~\ref{prop:strong-convexity} in Appendix~\ref{app:technical}), where
\begin{equation}\label{eq:Qj}
Q_j
=
V_j\mathbf 1\mathbf 1^\top
+
\mathbb E_X\sum_{a\in\mathcal A}
W_j(X,a)\boldsymbol\mu(X,a)
\boldsymbol\mu(X,a)^\top.
\end{equation}

We therefore impose the following condition.

\begin{asu}
\label{ass:identifiability}
For every target policy $\pi_j$,
$
\lambda_j
=
\inf_{h\neq0,\ c^\top h=0}
\frac{h^\top Q_jh}{\|h\|_2^2}>0.
$
\end{asu}

This assumption ensures sufficient curvature of $\mathcal G$ on
$\mathcal C$, which guarantees uniqueness of $z^\star$. The condition plays the role of the full-rank conditions of optimal design
\citep{pukelsheim2006}. Under
Assumptions~\ref{ass:known}, \ref{ass:pooled}, and
\ref{ass:identifiability}, $\mathcal G$ is strongly convex with
$m=2c_{\min}^3\min_j\lambda_j$, and thus, $z^\star$ is unique.
These results are
Proposition~\ref{prop:strong-convexity} and
Lemma~\ref{lem:quadratic-growth} in
Appendix~\ref{app:convergence}, and their proofs are in Appendix~\ref{app:technical}.

Now, we focus on the rate of convergence. Recall that our selection rule uses $\widehat G_{j,t}$ and $\widehat\Delta_{j,k,t}$ from
Section~\ref{sec:selection}. Since $G_j(N_t)$ is of order
$C_t^{-1}$ and $\Delta_{j,k}(N_t)$ is of order $C_t^{-2}$
when pooled coverage is bounded below, we need their estimation
errors to be small on these scales. We impose the following
assumption.

\begin{asu}\label{ass:nuisance}
There is a deterministic sequence $r_t \to 0$ such that,
almost surely for all sufficiently large $t$,
\begin{equation}\label{eq:acquisition-rates}
C_t\max_j
\left|\widehat G_{j,t}-G_j(N_t)\right|
\le r_t,
\qquad
C_t^2\max_{j,k}
\left|\widehat\Delta_{j,k,t}-\Delta_{j,k}(N_t)\right|
\le r_t.
\end{equation}
\end{asu}

The accuracy of these estimates depends on the estimation
of $V_j$, $W_j$, and $P_X$. Lemma~\ref{lem:plugin} in Appendix~\ref{app:convergence}
gives sufficient conditions for Assumption~\ref{ass:nuisance}.
We also verify these conditions for finite context spaces
in Corollary~\ref{cor:finite-contexts}.

We now make the Frank--Wolfe interpretation more precise. Write $z_t=z_{N_t}$, and recall from Section~\ref{sec:selection} that the rescaled index
$C_t^2\widehat S_{t,k}$ is approximately $-\langle\nabla\Phi_{\rho_t}(z_t),e_k/c_k\rangle$.

\begin{proposition}\label{prop:fw-step}
Under Assumptions~\ref{ass:known} and \ref{ass:nuisance}, there is a finite constant $D$, depending
only on $c_{\min}$, $c_{\max}$, $R_{\max}$, $\eta$, $\eta_0$, and $\rho_0$, such that, almost surely
for all sufficiently large $t$,
\begin{equation}\label{eq:score-gradient}
\max_{1\le k\le K}\Big|C_t^2\widehat S_{t,k}+\Big\langle\nabla\Phi_{\rho_t}(z_t),\frac{e_k}{c_k}\Big\rangle\Big|
\le\varepsilon_t,
\qquad
\varepsilon_t=D\big\{\kappa(\theta_t)^{-2}C_t^{\alpha}r_t+\kappa(\theta_t)^{-4}C_t^{\alpha-1}\big\}.
\end{equation}
Consequently, whenever all loggers are affordable and the step is not forced, the vertex
$v_t=e_{K_{t+1}}/c_{K_{t+1}}$ purchased by Algorithm~\ref{alg:acquisition} satisfies
\begin{equation}\label{eq:approx-LMO}
\langle\nabla\Phi_{\rho_t}(z_t),v_t\rangle
\le
\min_{v\in\mathcal C}\langle\nabla\Phi_{\rho_t}(z_t),v\rangle+\delta_t,
\qquad\delta_t=2\varepsilon_t .
\end{equation}
\end{proposition}
This proposition formalizes the connection between our selection
rule and the Frank--Wolfe method. Our algorithm chooses the logger
with the largest estimated index $\widehat S_{t,k}$ while a
Frank--Wolfe step would minimize the linear approximation of
$\Phi_{\rho_t}$ over $\mathcal C$. The first bound shows that
these two are close and the second shows that our
selection is a $\delta_t$-approximate conditional-gradient step. Note that the approximation error depends on both $r_t$ and pooled coverage $\kappa(\theta_t)$. Therefore, if coverage is small, the errors can become fairly large. As discussed in
Section~\ref{sec:algorithm}, forced sampling is introduced to
prevent exactly this. Lemma~\ref{lem:coverage} in
Appendix~\ref{app:convergence} shows that
$\kappa(\theta_t)\ge\chi_t/2$ and that the step sizes of the forced purchases are summable.
Together with the accuracy assumption, these bounds allow us
to establish convergence of the acquired allocation.

We now state our main result. Because the last purchases depend on the budget, we state the
estimation condition along a single sequence that does not depend on $B$: the infinite-budget
sequence, in which every logger remains affordable, with counts $\bar N_t$ and expenditure $\bar C_t$.
The relation between the budget-$B$ and infinite-budget sequences is
provided in Appendix~\ref{app:convergence}. 
The conditions in the theorem are needed to control the
errors in Proposition~\ref{prop:fw-step}. Recall that
$\upsilon$ determines how quickly the forced-sampling threshold
decreases. The condition $\upsilon<(1-\alpha)/4$ in \eqref{eq:tuning} ensures that the error terms involving the pooled
coverage decrease as more observations are collected. We also require
$t^{\alpha+2\upsilon}r_t\to0$ to control the error
from estimating the criterion and gains.

\begin{thm}\label{thm:design-optimality}
Suppose Assumptions~\ref{ass:known}, \ref{ass:pooled}, and \ref{ass:identifiability} hold,
Assumption~\ref{ass:nuisance} holds along the infinite-budget sequence, with $N_t$ and $C_t$ replaced by
$\bar N_t$ and $\bar C_t$, the tuning constants satisfy \eqref{eq:tuning}, and
$t^{\alpha+2\upsilon}r_t\to0$. Then
\begin{enumerate}[label=(\roman*),topsep=0pt,itemsep=2pt]
\item $z_{T_B}\to z^\star$ and
$\widehat\theta_B\to\theta^\star$ almost surely as $B\to\infty$, and forced sampling occurs only
finitely often along the infinite-budget sequence.
\item If, moreover, $r_t=O(t^{-\beta_r})$ for some $\beta_r>\alpha$, then almost surely
\[
\mathcal G(z_{T_B})-\mathcal G(z^\star)=O(B^{-\beta}),\qquad
\|\widehat\theta_B-\theta^\star\|_2=O(B^{-\beta/2}),
\]
where $\beta=\min\{\alpha,1-\alpha,\beta_r-\alpha\}$.
\end{enumerate}
\end{thm}

Part (i) shows that \textit{Adaptive Greedy} indeed converges to the optimal design $z^\star$. It also shows that forced sampling eventually stops once the allocation is sufficiently close to the optimum as the coverage is bounded away from zero and the usual greedy rule takes over. In part (ii) each term of $\beta$ comes from one source of error: smoothing gives $\alpha$, purchasing one observation at a time gives $1-\alpha$, and estimation gives $\beta_r-\alpha$. The first term increases in $\alpha$ and the other two decrease, so taking $\alpha=\min\{\beta_r,1\}/2$ gives the largest exponent, $\beta=\min\{\beta_r,1\}/2$.

\subsection{Inference}\label{sec:adaptive-inference}

In this subsection we derive the asymptotic normality of $\widehat J_B$. The main difficulty is that, unlike in the fixed-design setting, where the estimator is simply an average of independent terms with variance $A_j(\theta)/t$, each observation in \eqref{eq:terminal} is divided by the pooled behavior probability
$m_{\widehat\theta_B}$ and $\widehat\theta_B$ depends on the rewards
observed during purchasing. The denominator is therefore unknown when each observation is
collected. Nevertheless, we show that the estimator is
asymptotically normal and that its asymptotic variance for target $\pi_j$ is $A_j(\theta^\star)$, the
fixed-design bound at the composition to which the rule converges.

We compare $\widehat J_B$ with averages in which $q$ replaces $\widehat q_B$ and a fixed composition
$\theta$ with $\kappa(\theta)>0$ replaces $\widehat\theta_B$; their summand is the score
\begin{equation}\label{eq:score-vector}
\psi_{i}(\theta)
=
q_\Pi(X_i)
+
\frac{\pi(A_i\mid X_i)}{m_\theta(A_i\mid X_i)}
\{R_i-q(X_i,A_i)\} ,
\end{equation}
where $\pi(a\mid x)=\{\pi_1(a\mid x),\ldots,\pi_M(a\mid x)\}^\top$ and
$q_\Pi(x)=\{q_1(x),\ldots,q_M(x)\}^\top$.

The main idea involves three steps. First, consider $\widetilde J_B^\star
=
\frac{1}{T_B}\sum_{i=1}^{T_B}\psi_i(\theta^\star)$ which uses the limiting composition $\theta^\star$, which is unknown. Although the loggers are selected adaptively, $K_i$ is
chosen before observation $i$ is collected and since
$R_i-q(X_i,A_i)$ has conditional mean zero,
$\psi_i(\theta^\star)-J$ also has conditional mean
zero given the past. Thus, these scores form a
martingale difference sequence. Their average conditional
variance depends on $\widehat\theta_B$, which converges to $\theta^\star$
by Theorem~\ref{thm:design-optimality}. The limiting
variance is therefore $\Sigma_A(\theta^\star)$, where
\begin{equation}\label{eq:SigmaA}
\Sigma_A(\theta)
=
\mathrm{Cov}\{q_\Pi(X)\}
+
\mathbb E_X\sum_{a\in\mathcal A}
\frac{\sigma^2(X,a)\pi(a\mid X)\pi(a\mid X)^\top}{m_\theta(a\mid X)} ,
\end{equation}
and the $j$th diagonal entry of $\Sigma_A(\theta)$ is $A_j(\theta)$ in \eqref{eq:Aj}. We can then apply a
martingale central limit theorem \citep{hallheyde1980}. The second step is to replace $\theta^\star$ by $\widehat\theta_B$. By
Assumption~\ref{ass:pooled}, the pooled probabilities
remain bounded away from zero when the composition
is sufficiently close to $\theta^\star$. Since
$\widehat\theta_B\to\theta^\star$, we need to show
that using $\widehat\theta_B$ instead of $\theta^\star$
does not change the limiting distribution. We show
in Appendix~\ref{app:inference} that
\[
\frac{1}{T_B}\sum_{i=1}^{T_B}
\{\psi_i(\widehat\theta_B)
-\psi_i(\theta^\star)\}
=
o_p(T_B^{-1/2}).
\]

Finally, we need to account for the estimated reward
regression $\widehat q_B$. Replacing $q$ by
$\widehat q_B$ introduces the remainder
\begin{equation}\label{eq:terminal-remainder}
\mathcal R_{B,j}(\theta)
=
\frac1{T_B}\sum_{i=1}^{T_B}
\left[
\{\widehat q_{B,j}-q_j\}(X_i)
-
\frac{\pi_j(A_i\mid X_i)}
{m_\theta(A_i\mid X_i)}
\{\widehat q_B-q\}(X_i,A_i)
\right].
\end{equation}
We therefore impose the following assumption.

\begin{asu}\label{ass:agg-orth}
There is a deterministic sequence $r_B\downarrow0$ such that
\begin{equation}\label{eq:agg-orth}
\mathbb P\!\left(
\|\widehat\theta_B-\theta^\star\|_2\le r_B
\right)\to1,
\qquad
\max_{1\le j\le M}
\sup_{\|\theta-\theta^\star\|_2\le r_B}
|\mathcal R_{B,j}(\theta)|
=
o_p(T_B^{-1/2}).
\end{equation}
\end{asu}

This is the usual condition under which the estimated regression does not affect an augmented
estimator at first order, and in semiparametric inference it is commonly obtained by sample
splitting or cross-fitting \citep{chernozhukov2018}. In Appendix~\ref{app:blocksplit} we describe a block-split estimator, motivated by cross-fitting, that avoids reusing the same observations for regression. Related conditions for adaptively collected
data appear in \citet{hadad2021}, \citet{zhan2021}, and \citet{bibaut2021}. Because
$\widehat\theta_B$ is random, the remainder is required to be small over a neighborhood of
$\theta^\star$ that contains $\widehat\theta_B$ with probability tending to one. Since $r_B\to0$, this neighborhood eventually lies in
$\{\theta:\|\theta-\theta^\star\|_1\le \kappa_\star/2\}$, where $m_\theta\ge \kappa_\star/2$ on $\mathcal S$
by Assumption~\ref{ass:pooled}, so the remainder is well defined there. The first
condition in \eqref{eq:agg-orth} only limits how fast this neighborhood may shrink, and by
Theorem~\ref{thm:design-optimality} it holds for any $r_B$ that decreases slowly enough. The
assumption concerns only the final regression in \eqref{eq:terminal} and is separate from
Assumption~\ref{ass:nuisance}, which governs the estimates used while purchasing.

\begin{thm}\label{thm:feasible-terminal-clt}
Under the conditions of part (i) of Theorem~\ref{thm:design-optimality} and
Assumption~\ref{ass:agg-orth},
\begin{equation}\label{eq:main-clt}
\sqrt{T_B}\left(\widehat J_B-J\right)
\to
N\{0,\Sigma_A(\theta^\star)\}
\qquad\text{in distribution.}
\end{equation}
\end{thm}

For each target, the asymptotic variance of the estimate computed from the purchased sample is
therefore $A_j(\theta^\star)$, the fixed-design bound at the composition to which the rule
converges, which is the quantity whose largest value over $\Pi$ the rule was built to reduce.

We need to estimate the covariance as well. With $\widehat\psi_i$ the $M$-vector summand of
\eqref{eq:terminal}, we use
\begin{equation}\label{eq:Sigmahat}
\widehat\Sigma_{B}
=
\frac1{T_B}\sum_{i=1}^{T_B}
(\widehat\psi_i-\widehat J_B)(\widehat\psi_i-\widehat J_B)^\top .
\end{equation}

\begin{corollary}\label{cor:student}
Suppose the conditions of Theorem~\ref{thm:feasible-terminal-clt} hold and the terminal regression
is consistent in mean square over the sample,
$T_B^{-1}\sum_{i\le T_B}\sum_{a\in\mathcal A}\{\widehat q_B(X_i,a)-q(X_i,a)\}^2\to0$ in
probability. Then $\widehat\Sigma_B\to\Sigma_A(\theta^\star)$ in probability. Consequently, for every
target $\pi_j$ with $A_j(\theta^\star)>0$, the intervals
\begin{equation}\label{eq:wald}
\widehat J_{B,j}\pm z_{1-\gamma/2}\sqrt{(\widehat\Sigma_B)_{jj}/T_B},
\qquad j=1,\ldots,M,
\end{equation}
have asymptotic coverage $1-\gamma$ for $J_j$.
\end{corollary}

Because Theorem~\ref{thm:feasible-terminal-clt} is a joint statement over the target class,
simultaneous regions also follow. A Bonferroni band replaces $z_{1-\gamma/2}$ in \eqref{eq:wald} by
$z_{1-\gamma/(2M)}$, and a less conservative band uses the $1-\gamma$ quantile of $\max_j|Z_j|$,
where $Z\sim N(0,\widehat R_B)$ and $\widehat R_B$ is the correlation matrix associated with
$\widehat\Sigma_B$.

The pooled weighting also matters for efficiency. A logger-specific estimator, which divides each
observation by the propensity of the logger that generated it, has fixed-design covariance
$\Sigma_L(\theta)$, defined in \eqref{eq:SigmaL} of Appendix~\ref{app:aggregation}, and
\begin{equation}\label{eq:aggregate-local-order}
\Sigma_A(\theta)\preceq\Sigma_L(\theta)
\qquad
\text{for every }\theta\in\mathcal S_K ,
\end{equation}
an ordering known for fixed designs \citep{agarwal2017,kallus2021}, where the right side is infinite when a logger with $\theta_k>0$ gives probability zero to an action that a target policy uses. Theorem~\ref{thm:feasible-terminal-clt} shows that the pooled estimator variance $A_j(\theta^\star)$
is no larger than the fixed-design variance of the logger-specific estimator at the same
composition.

\section{Numerical Studies}\label{sec:experiments}

In this section, we evaluate \textit{Adaptive Greedy} (Algorithm~\ref{alg:acquisition}) in three synthetic
environments with known optimal designs and in a semi-synthetic application based on the AIDS
Clinical Trials Group Study 175 \citep{hammer1996,hammer1996uci}. Implementation details, the
complete data-generating process, and additional experiments are given in
Appendix~\ref{app:experiments}.

\subsection{Simulation Setup}\label{sec-simulation}

All three environments have $K=5$ loggers, five actions, $M=4$ target policies, and a
five-dimensional context drawn from a three-component Gaussian mixture. The reward is a clipped
normal variable,
\[
R=\min\{4,\max(-4,R^\circ)\},\qquad R^\circ\mid X=x,A=a\sim N\{q_0(x,a),s^2(x,a)\},
\]
where $q_0(x,a)$ is linear in $x$, with an additional interaction $x_1x_2$ for actions 3 to 5, and
the variance $s^2(x,a)$ is larger for actions 4 and 5 and increases with $x_3^2$. Loggers 1 to 4
concentrate on the action pairs $(1,2)$, $(2,3)$, $(3,4)$, and $(4,5)$, so each covers a different
part of the action space, while Logger 5 is close to uniform. The target policies are softmax
policies mixed with a uniform exploration floor. The environments differ as follows.
\begin{itemize}[topsep=2pt,itemsep=1pt]
\item \textbf{Environment A} has costs $c=(1,2,4,8,16)$ and optimal design
$\theta^\star\approx(0.784,0,0.085,0.132,0)$.
\item \textbf{Environment B} keeps the loggers and costs of Environment A but uses more concentrated target
policies. Its optimal design $\theta^\star\approx(0.715,0.074,0.138,0.073,0)$ includes Logger 2,
so a change in the targets alone changes which loggers are worth purchasing from.
\item \textbf{Environment C} keeps the policies of Environment A and changes the costs to
$c=(1,4,16,8,2)$. Its optimal design $\theta^\star\approx(0.257,0,0,0,0.743)$ puts most of the
sample on Logger 5 even though Logger 1 is cheaper, so the best allocation depends on statistical
value per unit cost rather than on cost alone.
\end{itemize}

We compare \textit{Adaptive Greedy} with five procedures. All of them use the same initialization
and, where they need them, the same estimators of the reward mean and variance.
\begin{itemize}[topsep=2pt,itemsep=1pt]
\item \textit{Reoptimized plug-in} re-estimates $V_j$ and $W_j$ every ten observations, solves a
continuous relaxation of the static allocation problem \eqref{eq:static} with these estimates,
rounds the resulting allocation to integer counts, and purchases toward that allocation until the next
update.
\item \textit{Estimated static} solves the same relaxed problem once, after the initialization, and
follows the resulting integer allocation to the end.
\item \textit{Oracle static} uses the same procedure with the true $V_j$ and $W_j$. It is not feasible
in practice and serves as a reference.
\item \textit{Equal expenditure} spends the same amount on each logger, so its sample fractions are
proportional to $1/c_k$.
\item \textit{Cheapest only} always purchases from the cheapest affordable logger.
\end{itemize}
Every procedure starts with four observations from each logger. For each action, we estimate the
reward mean by ridge regression and the reward variance from the squared residuals of that fit
(Appendix~\ref{app-implementation}). Both are refitted every ten observations. Adaptive Greedy
uses $\alpha=0.4$ and $\rho_0=0.5$ in all settings, and forced sampling uses $\chi_0=10^{-4}$ and
$\upsilon=0.01$. Every logger in our experiments gives every action a probability bounded below by
more than $\chi_0$, so forced sampling never occurs (Appendix~\ref{app-implementation}). After the
budget is spent, the policy values are estimated by the final estimator \eqref{eq:terminal}. Since the design
criterion is a worst case over the target class, our main accuracy measure is the worst-target root
mean squared error,
\[
\operatorname{WorstRMSE}=\max_{1\le j\le M}\Big\{\frac1R\sum_{r=1}^R(\widehat J_{r,j}-J_j)^2\Big\}^{1/2},
\]
where $\widehat J_{r,j}$ is the estimate of $J_j$ in replication $r$. To measure how close
the purchased design $\widehat z=z_{T_B}$ is to the optimal design, we also report
$\|\widehat\theta-\theta^\star\|_2$ and $\mathcal G(\widehat z)/\mathcal G(z^\star)$, which equals one at the
optimal design. The budgets are
$B\in\{500,1000,1500,3000,6000,12000,24000\}$, with $R=200$ replications in Environments A and B
and $R=100$ in Environment C. We use common random numbers across procedures. For RMSE quantities,
one Monte Carlo standard error is computed from $5000$ bootstrap resamples of the replications. For quantities averaged directly over replications, such as
$\mathcal G(\widehat z)/\mathcal G(z^\star)$ and $\|\widehat\theta-\theta^\star\|_2$, it is the sample standard
deviation divided by $\sqrt R$.

\subsection{Simulation Results}\label{sec-simulation-results}

\begin{figure}[t]
    \centering
    \includegraphics[width=\textwidth]
    {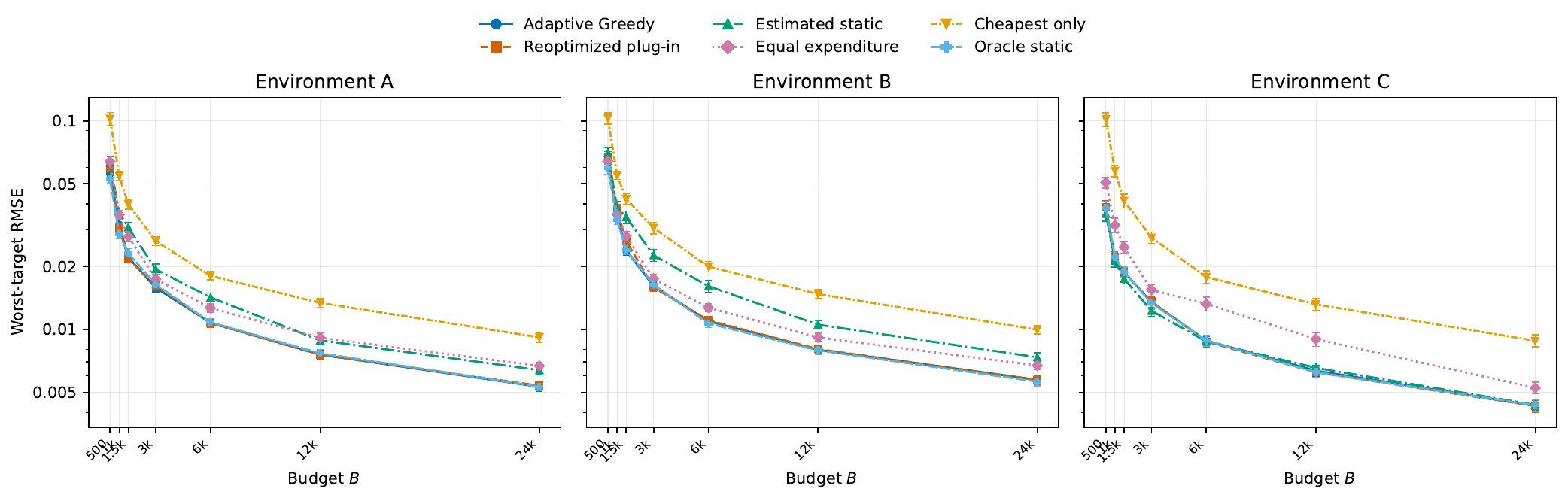}
    \caption{
Worst-target RMSE in the three synthetic environments.
Error bars show $\pm1$ Monte Carlo standard error, and the vertical axis is logarithmic.
}
    \label{fig-main-rmse}
\end{figure}

Figure~\ref{fig-main-rmse} and Table~\ref{tab-main-rmse-large} show that Adaptive Greedy is among
the most accurate feasible procedures in all three environments. Its accuracy also approaches that
of Oracle static as the budget grows. At $B=24000$, its worst-target RMSE is within about $2\%$ of
Oracle static in every environment. It is $17\%$ and $22\%$ lower than that of Estimated static in
Environments A and B, $15\%$ to $21\%$ lower than that of Equal expenditure, and $42\%$ to $51\%$
lower than that of Cheapest only. Reoptimized plug-in reaches almost the same accuracy as Adaptive
Greedy. In Environment C, Estimated static also performs well, while Cheapest only has about twice
the worst-target RMSE of Adaptive Greedy, which confirms that cost alone does not identify a good
design. At small budgets, the differences between procedures are smaller and their ordering varies.
Paired bootstrap comparisons separate Adaptive Greedy from Cheapest only at every budget, and from
Estimated static and Equal expenditure at most budgets in Environments A and B
(Appendix~\ref{app-paired-comparisons}).

\begin{table}[t]
\centering
\caption{
Worst-target RMSE ($\times10^{-3}$) at $B=24000$ in the synthetic environments, with Monte Carlo
standard errors ($\times10^{-3}$) in parentheses. Oracle static uses the true reward moments and is
not feasible in practice.
}
\label{tab-main-rmse-large}
\begin{tabular}{lccc}
\toprule
Method & Environment A & Environment B & Environment C \\
\midrule
Adaptive Greedy     & \textbf{5.30} (0.25) & \textbf{5.73} (0.27) & \textbf{4.31} (0.28) \\
Reoptimized plug-in & \textbf{5.37} (0.25) & \textbf{5.72} (0.27) & \textbf{4.29} (0.28) \\
Estimated static    & 6.38 (0.30) & 7.35 (0.37) & 4.35 (0.25) \\
Equal expenditure   & 6.69 (0.29) & 6.71 (0.28) & 5.24 (0.34) \\
Cheapest only       & 9.17 (0.48) & 9.97 (0.45) & 8.83 (0.59) \\
\textit{Oracle static}       & 5.31 (0.24) & 5.62 (0.25) & 4.33 (0.29) \\
\bottomrule
\end{tabular}
\end{table}

\begin{figure}[t]
    \centering
    \includegraphics[width=\textwidth]
    {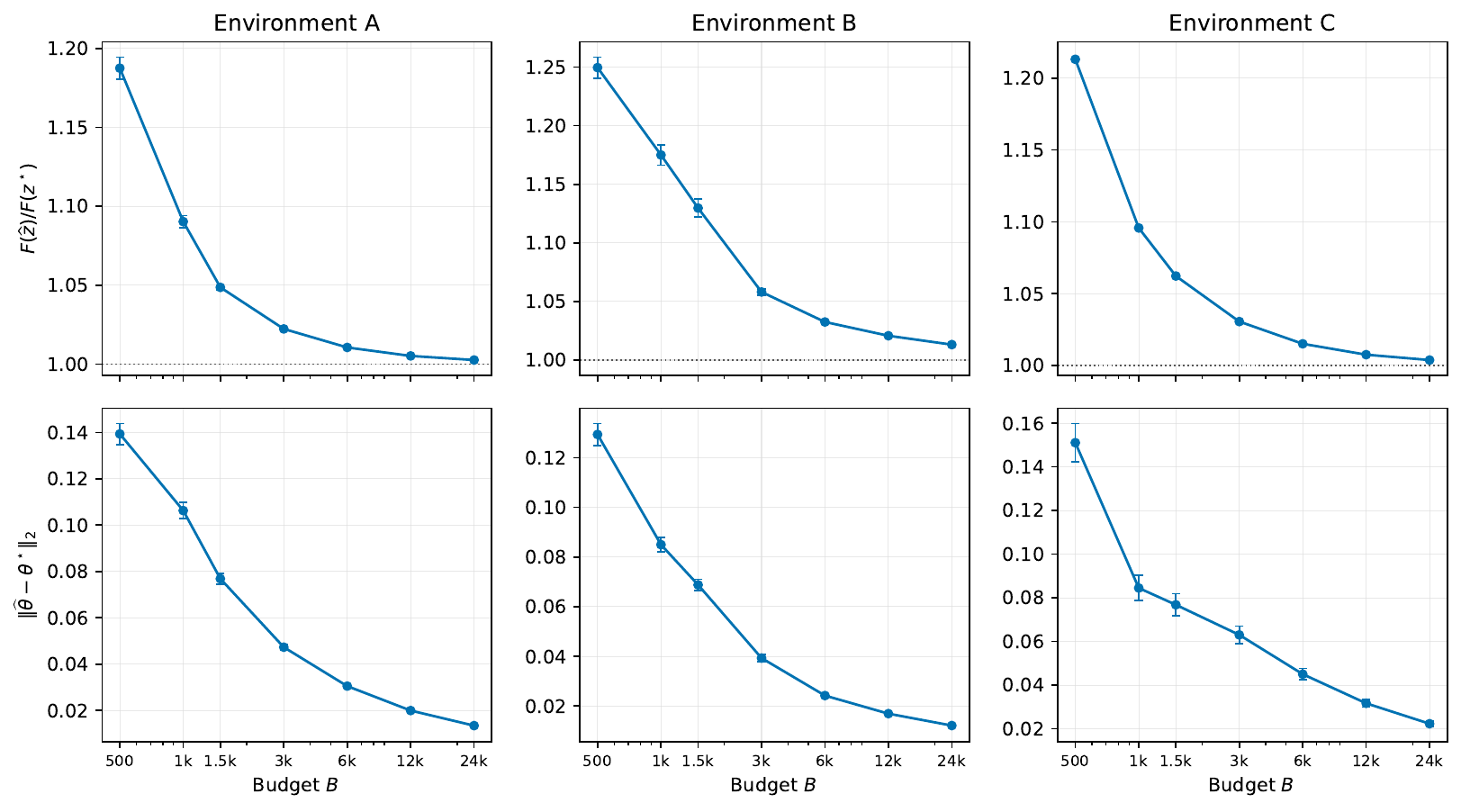}
    \caption{
Convergence of the Adaptive Greedy design. The top row shows $\mathcal G(\widehat z)/\mathcal G(z^\star)$ and the
bottom row shows $\|\widehat\theta-\theta^\star\|_2$. Error bars show $\pm1$ Monte Carlo standard
error, and the budget axis is logarithmic.
}
    \label{fig-design-convergence}
\end{figure}

Figure~\ref{fig-design-convergence} shows that the design purchased by Adaptive Greedy converges to
the optimal design in all three environments, as parts (i) and (ii) of
Theorem~\ref{thm:design-optimality} predict. Both $\mathcal G(\widehat z)/\mathcal G(z^\star)$ and
$\|\widehat\theta-\theta^\star\|_2$ decrease steadily with the budget.

The Wald intervals of Corollary~\ref{cor:student} are well calibrated. We use $R=500$ replications
in Environments A and B at $B=6000$ and $B=24000$. The ratio of the mean estimated standard error to
the empirical standard deviation lies between $0.989$ and $1.060$. Target-specific $95\%$ coverage
lies between $0.944$ and $0.966$, and simultaneous coverage lies between $0.950$ and $0.960$
(Appendix~\ref{app-inference-calibration}). The appendix reports three further findings. First,
Adaptive Greedy is robust to a misleading initialization, which degrades Estimated static much more
(Appendix~\ref{app-stress}). Second, its accuracy changes little over a range of $\alpha$ and
$\rho_0$ (Appendix~\ref{app-parameter-sensitivity}). Finally, the design criterion closely tracks
estimation accuracy, with a correlation of about $0.97$ at budgets of at least $3000$
(Appendix~\ref{app-f-rmse}).

\subsection{Comparison with Reoptimized Plug-in}\label{sec-anytime}

\begin{figure}[t]
    \centering
    \includegraphics[width=\textwidth]
    {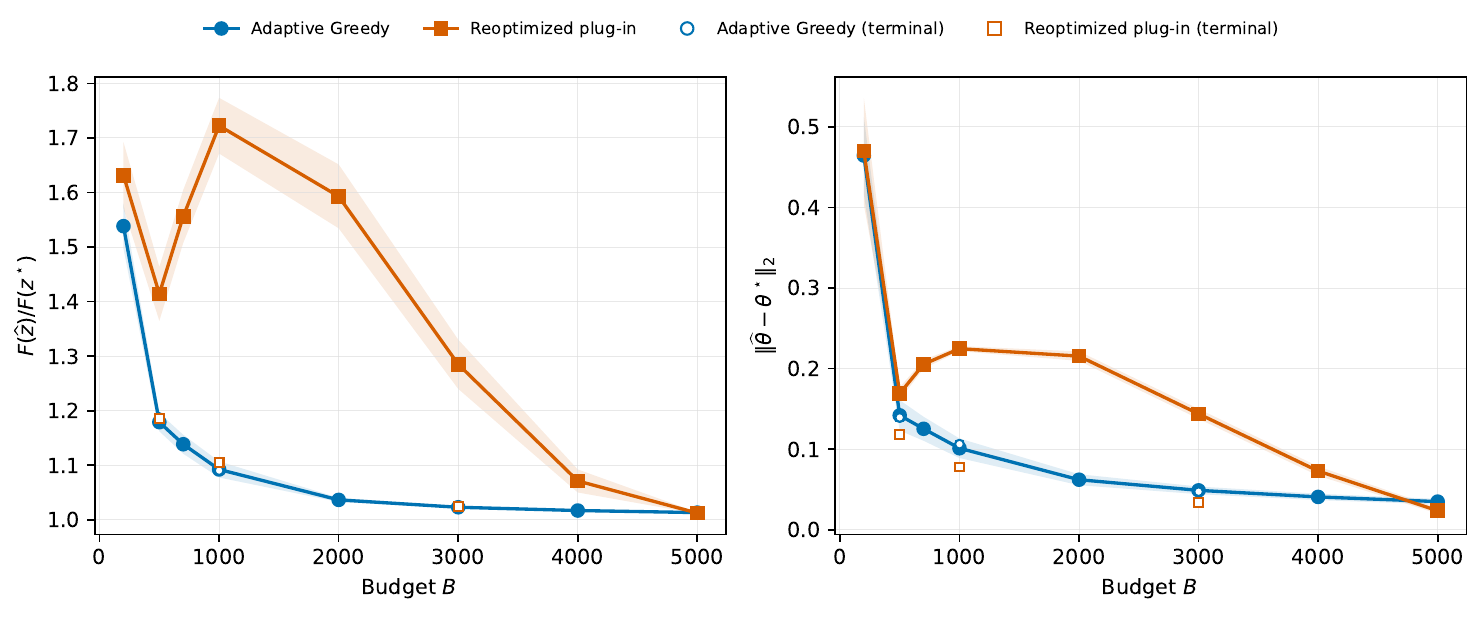}
    \caption{
Designs at intermediate budgets in Environment A. Solid curves follow runs with budget $5000$,
and open markers show separate runs whose budget equals that intermediate budget. Shaded bands and error bars
show Monte Carlo uncertainty.
}
    \label{fig-anytime}
\end{figure}

Adaptive Greedy and Reoptimized plug-in reach similar accuracy when the budget is spent, but they
get there differently. Reoptimized plug-in plans for the declared budget $B$. However, Adaptive
Greedy uses $B$ only to decide when to stop, so its purchases up to any point do not depend on $B$.
This matters when the final budget is uncertain or the run may end early. In Environment A, we ran
both procedures with budget $5000$ and recorded their designs at intermediate budgets. We also
ran separate runs whose budget equals each intermediate budget. Figure~\ref{fig-anytime} shows that the
Adaptive Greedy design at every intermediate budget is almost the same as that of a run planned for that
budget, so stopping early costs little. The Reoptimized plug-in design instead moves away from the
optimum early on, with $\mathcal G(\widehat z)/\mathcal G(z^\star)$ rising from $1.41$ at $B=500$ to $1.72$ at
$B=1000$, because its early purchases follow a plan made for $B=5000$
(Appendix~\ref{app-anytime-details}).

Adaptive Greedy is also much cheaper to run, since it compares $K$ indices at each step instead of
solving an allocation problem every ten observations. In Environment A, Reoptimized plug-in took
$8.7$ times as long as Adaptive Greedy at $B=12000$ ($6423$ versus $742$ seconds) and $9.4$ times
as long at $B=24000$ ($16442$ versus $1750$ seconds). Thus Adaptive Greedy matches the
accuracy of repeated reoptimization at about a tenth of its computing cost, and it provides a good
design at any stopping point.

\subsection{Application to ACTG 175 Data}\label{sec-actg}

The AIDS Clinical Trials Group Study 175 \citep{hammer1996} was a randomized trial in HIV-infected
adults that compared four treatments, zidovudine (ZDV), ZDV with didanosine (ddI), ZDV with
zalcitabine, and ddI alone. The public data \citep{hammer1996uci} contain $2139$ subjects with
complete records and have been used to study individualized treatment regimes \citep{fan2017}. We
take the four treatments as the actions, seven baseline covariates as the context (age, weight, CD4
count, CD8 count, Karnofsky score, symptomatic status, and prior antiretroviral therapy), and the
change in CD4 count from baseline to around week $20$, divided by $100$, as the reward. To make the
policy values and the optimal design known, we fit an arm-specific regression of the reward on the
covariates and generate rewards from it, with contexts resampled from the $2139$ subjects
(Appendix~\ref{app-actg-details}).

We construct five loggers with costs $c=(1,2,4,3,8)$. Loggers 1 to 4 each
favor a different pair of treatments, and Logger 5 is uniform. The loggers and costs are part of
the experimental design and do not describe the trial itself. Of the four target policies, one
favors the treatments with higher fitted rewards and three are softmax policies. The optimal design
$\theta^\star\approx(0.811,0,0.028,0.161,0)$ uses only Loggers 1, 3, and 4. Logger 1 covers the
first two treatments cheaply, Loggers 3 and 4 cover the other two, and the broad coverage of the
uniform logger does not justify its cost. We use budgets $500$, $1500$, $6000$, and $12000$, with
$R=200$ replications at each.

\begin{table}[t]
\centering
\caption{
Worst-target RMSE ($\times10^{-3}$) in the ACTG 175 application, with Monte Carlo standard errors
($\times10^{-3}$) in parentheses. As in Table~\ref{tab-main-rmse-large}, Oracle static uses the true
reward moments and is not feasible in practice.
}
\label{tab-actg-rmse}
\setlength{\tabcolsep}{5pt}
\begin{tabular}{lcccc}
\toprule
Method & $B=500$ & $B=1500$ & $B=6000$ & $B=12000$ \\
\midrule
Adaptive Greedy     & \textbf{113} (6.2) & \textbf{58.4} (2.6) & \textbf{28.3} (1.4) & \textbf{18.8} (0.85) \\
Reoptimized plug-in & 125 (9.7) & \textbf{56.1} (2.4) & \textbf{28.2} (1.4) & \textbf{18.6} (0.87) \\
Estimated static    & 134 (9.2) & 65.5 (7.2) & 31.0 (1.5) & 21.9 (1.6) \\
Equal expenditure   & \textbf{119} (5.7) & 63.1 (3.1) & 31.2 (1.5) & 22.8 (1.2) \\
Cheapest only       & 125 (7.2) & 60.9 (2.9) & 30.4 (1.5) & 21.2 (0.96) \\
\textit{Oracle static}       & 121 (11)  & 56.7 (2.6) & 27.9 (1.4) & 18.6 (0.84) \\
\bottomrule
\end{tabular}
\end{table}

\begin{figure}[t]
    \centering
    \includegraphics[width=\textwidth]
    {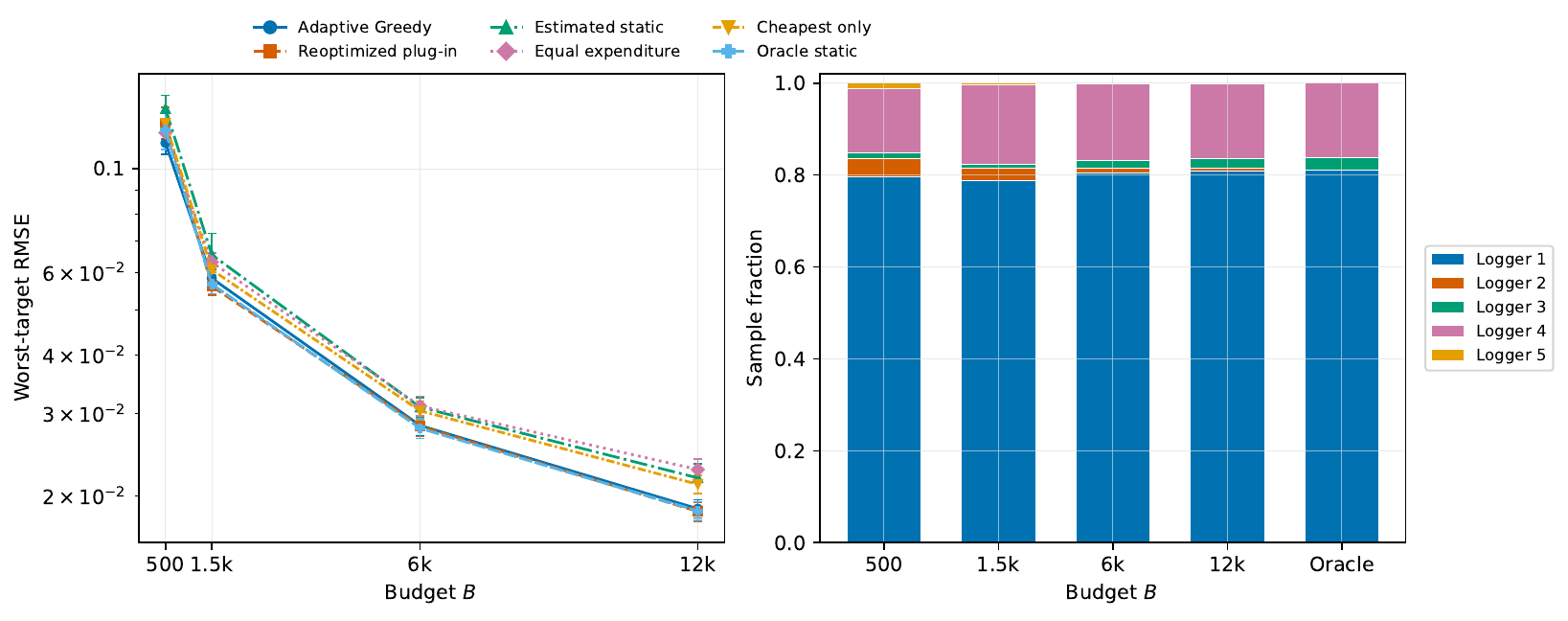}
    \caption{
ACTG 175 application. Left, worst-target RMSE with $\pm1$ Monte Carlo standard error. Right, the
mean allocation of Adaptive Greedy across the five loggers, compared with the optimal allocation.
}
    \label{fig-actg}
\end{figure}

Table~\ref{tab-actg-rmse} shows the same pattern as the simulations. At $B=12000$, Adaptive Greedy
matches Reoptimized plug-in and Oracle static, and its worst-target RMSE is $11\%$ to $18\%$ lower
than those of Estimated static, Equal expenditure, and Cheapest only. At the smallest budget,
$B=500$, it has the smallest reported worst-target RMSE among the procedures, including Oracle
static, although the Monte Carlo uncertainty is substantial at this budget.
Figure~\ref{fig-actg} shows the purchased allocation approaching the optimal design, with the mass
on Loggers 2 and 5 shrinking toward zero. Note that simple cost-based rules are far from optimal
here. The criterion ratio $\mathcal G(z_\theta)/\mathcal G^\star$, where $z_\theta=\theta/c^\top\theta$ is the design with
sample fractions $\theta$, is $2.50$ for equal sample counts, $1.40$ for Equal expenditure, and $1.21$
for purchasing only from the cheapest logger.

\section{Discussion}\label{sec:discussion}

In this paper, we studied budgeted data acquisition in off-policy evaluation, in
which observations are purchased sequentially from a fixed set of loggers with heterogeneous costs.
We derived the Adaptive Greedy rule from the pooled efficiency bound of \citet{kallus2021}, and
showed that each purchase is an approximate Frank--Wolfe step \citep{frank1956} on the worst-target
criterion. We also showed that the purchased design converges to the
minimax design. The pooled estimator computed from the purchased sample
remains asymptotically normal, with variance equal to the fixed-design bound at the selected
composition. Numerical studies show that the procedure matches the accuracy of repeated reoptimization
at a fraction of its computational cost.

Three extensions remain open. First, we assumed that the
logging probabilities are known. When they must be estimated, the estimation error enters both the
purchasing rule and the final estimator, although the efficiency bound itself is unchanged
\citep[Theorem~5]{kallus2021}. Second, inference for the final estimator relies on
Assumption~\ref{ass:agg-orth}, and establishing it for common regression methods under adaptive
sampling, in the spirit of \citet{hadad2021} and \citet{bibaut2021}, would be useful. Finally, we assumed fixed prices per
observation. When prices depend on the counts already purchased (variable pricing), the feasible set and its vertices
move with the design, and extending the Frank--Wolfe argument to this setting remains an open and interesting
problem.

\bibliographystyle{asa}
\bibliography{causal.bib}

@inproceedings{kallus2021,
  author    = {Kallus, Nathan and Saito, Yuta and Uehara, Masatoshi},
  title     = {Optimal Off-Policy Evaluation from Multiple Logging Policies},
  booktitle = {Proceedings of the 38th International Conference on Machine Learning},
  series    = {Proceedings of Machine Learning Research},
  volume    = {139},
  pages     = {5247--5256},
  year      = {2021},
  publisher = {PMLR}
}

@inproceedings{agarwal2017,
  author    = {Agarwal, Aman and Basu, Soumya and Schnabel, Tobias and Joachims, Thorsten},
  title     = {Effective Evaluation Using Logged Bandit Feedback from Multiple Loggers},
  booktitle = {Proceedings of the 23rd ACM SIGKDD International Conference on Knowledge Discovery and Data Mining},
  pages     = {687--696},
  year      = {2017},
  publisher = {ACM},
  doi       = {10.1145/3097983.3098155}
}

@inproceedings{he2019,
  author    = {He, Li and Xia, Long and Zeng, Wei and Ma, Zhi-Ming and Zhao, Yihong and Yin, Dawei},
  title     = {Off-policy Learning for Multiple Loggers},
  booktitle = {Proceedings of the 25th ACM SIGKDD International Conference on Knowledge Discovery and Data Mining},
  pages     = {1184--1193},
  year      = {2019},
  publisher = {ACM}
}

@article{hadad2021,
  author  = {Hadad, Vitor and Hirshberg, David A. and Zhan, Ruohan and Wager, Stefan and Athey, Susan},
  title   = {Confidence Intervals for Policy Evaluation in Adaptive Experiments},
  journal = {Proceedings of the National Academy of Sciences},
  volume  = {118},
  number  = {15},
  pages   = {e2014602118},
  year    = {2021},
  doi     = {10.1073/pnas.2014602118}
}

@inproceedings{zhan2021,
  author    = {Zhan, Ruohan and Hadad, Vitor and Hirshberg, David A. and Athey, Susan},
  title     = {Off-Policy Evaluation via Adaptive Weighting with Data from Contextual Bandits},
  booktitle = {Proceedings of the 27th ACM SIGKDD Conference on Knowledge Discovery and Data Mining},
  pages     = {2125--2135},
  year      = {2021}
}

@article{zhang2020,
  author  = {Zhang, Kelly W. and Janson, Lucas and Murphy, Susan A.},
  title   = {Inference for Batched Bandits},
  journal = {Advances in Neural Information Processing Systems},
  volume  = {33},
  pages   = {9818--9829},
  year    = {2020}
}

@article{dudik2014,
  author  = {Dud{\'i}k, Miroslav and Erhan, Dumitru and Langford, John and Li, Lihong},
  title   = {Doubly Robust Policy Evaluation and Optimization},
  journal = {Statistical Science},
  volume  = {29},
  number  = {4},
  pages   = {485--511},
  year    = {2014}
}

@book{hallheyde1980,
  author    = {Hall, Peter and Heyde, C. C.},
  title     = {Martingale Limit Theory and Its Application},
  publisher = {Academic Press},
  address   = {New York},
  year      = {1980}
}

@article{douglas2026,
  author  = {Douglas, Connor and Persson, Joel and Provost, Foster},
  title   = {Logging Policy Design for Off-Policy Evaluation},
  journal = {arXiv preprint arXiv:2605.15108},
  year    = {2026}
}

@article{nesterov2005,
  author  = {Nesterov, Yurii},
  title   = {Smooth Minimization of Non-Smooth Functions},
  journal = {Mathematical Programming},
  volume  = {103},
  pages   = {127--152},
  year    = {2005}
}

@article{gtoth2021lowerbounds,
  author  = {G.-T{\'o}th, B. and Casado, L. G. and Hendrix, E. M. T. and Messine, F.},
  title   = {On New Methods to Construct Lower Bounds in Simplicial Branch and Bound Based on Interval Arithmetic},
  journal = {Journal of Global Optimization},
  volume  = {80},
  pages   = {779--804},
  year    = {2021},
  doi     = {10.1007/s10898-021-01053-8}
}

@article{chernozhukov2018,
  author  = {Chernozhukov, Victor and Chetverikov, Denis and Demirer, Mert and Duflo, Esther and Hansen, Christian and Newey, Whitney and Robins, James},
  title   = {Double/Debiased Machine Learning for Treatment and Structural Parameters},
  journal = {The Econometrics Journal},
  volume  = {21},
  pages   = {C1--C68},
  year    = {2018}
}

@book{pukelsheim2006,
  author    = {Pukelsheim, Friedrich},
  title     = {Optimal Design of Experiments},
  publisher = {Society for Industrial and Applied Mathematics},
  address   = {Philadelphia, PA},
  series    = {Classics in Applied Mathematics},
  volume    = {50},
  year      = {2006},
  doi       = {10.1137/1.9780898719109},
  isbn      = {978-0-89871-604-7}
}

@inproceedings{li2010,
  author    = {Li, Lihong and Chu, Wei and Langford, John and Schapire, Robert E.},
  title     = {A Contextual-Bandit Approach to Personalized News Article Recommendation},
  booktitle = {Proceedings of the 19th International Conference on World Wide Web},
  pages     = {661--670},
  year      = {2010},
  publisher = {ACM},
  doi       = {10.1145/1772690.1772758}
}

@incollection{tewari2017ads,
  author    = {Tewari, Ambuj and Murphy, Susan A.},
  title     = {From Ads to Interventions: Contextual Bandits in Mobile Health},
  booktitle = {Mobile Health: Sensors, Analytic Methods, and Applications},
  editor    = {Rehg, James M. and Murphy, Susan A. and Kumar, Santosh},
  pages     = {495--517},
  publisher = {Springer},
  year      = {2017},
  doi       = {10.1007/978-3-319-51394-2_25}
}

@inproceedings{wang2017,
  author    = {Wang, Yu-Xiang and Agarwal, Alekh and Dud{\'i}k, Miroslav},
  title     = {Optimal and Adaptive Off-Policy Evaluation in Contextual Bandits},
  booktitle = {Proceedings of the 34th International Conference on Machine Learning},
  series    = {Proceedings of Machine Learning Research},
  volume    = {70},
  pages     = {3589--3597},
  year      = {2017},
  publisher = {PMLR}
}

@inproceedings{tucker2023,
  author    = {Tucker, Aaron D. and Joachims, Thorsten},
  title     = {Variance-Minimizing Augmentation Logging for Counterfactual Evaluation in Contextual Bandits},
  booktitle = {Proceedings of the Sixteenth ACM International Conference on Web Search and Data Mining},
  pages     = {967--975},
  year      = {2023},
  publisher = {ACM},
  doi       = {10.1145/3539597.3570452}
}

@article{hanna2024,
  author  = {Hanna, Josiah P. and Chandak, Yash and Thomas, Philip S. and White, Martha and Stone, Peter and Niekum, Scott},
  title   = {Data-Efficient Policy Evaluation Through Behavior Policy Search},
  journal = {Journal of Machine Learning Research},
  volume  = {25},
  number  = {313},
  pages   = {1--58},
  year    = {2024}
}

@inproceedings{mukherjee2024speed,
  author    = {Mukherjee, Subhojyoti and Xie, Qiaomin and Hanna, Josiah P. and Nowak, Robert},
  title     = {{SPEED}: Experimental Design for Policy Evaluation in Linear Heteroscedastic Bandits},
  booktitle = {Proceedings of the 27th International Conference on Artificial Intelligence and Statistics},
  series    = {Proceedings of Machine Learning Research},
  volume    = {238},
  pages     = {2962--2970},
  year      = {2024},
  publisher = {PMLR}
}

@inproceedings{mukherjee2024saver,
  author    = {Mukherjee, Subhojyoti and Hanna, Josiah P. and Nowak, Robert D.},
  title     = {{SaVeR}: Optimal Data Collection Strategy for Safe Policy Evaluation in Tabular {MDP}},
  booktitle = {Proceedings of the 41st International Conference on Machine Learning},
  series    = {Proceedings of Machine Learning Research},
  volume    = {235},
  pages     = {36531--36576},
  year      = {2024},
  publisher = {PMLR}
}

@inproceedings{liu2024behavior,
  author    = {Liu, Shuze and Zhang, Shangtong},
  title     = {Efficient Policy Evaluation with Offline Data Informed Behavior Policy Design},
  booktitle = {Proceedings of the 41st International Conference on Machine Learning},
  series    = {Proceedings of Machine Learning Research},
  volume    = {235},
  pages     = {32345--32368},
  year      = {2024},
  publisher = {PMLR}
}

@inproceedings{dann2023,
  author    = {Dann, Chris and Ghavamzadeh, Mohammad and Marinov, Teodor V.},
  title     = {Multiple-Policy High-Confidence Policy Evaluation},
  booktitle = {Proceedings of the 26th International Conference on Artificial Intelligence and Statistics},
  series    = {Proceedings of Machine Learning Research},
  volume    = {206},
  pages     = {9470--9487},
  year      = {2023},
  publisher = {PMLR}
}

@article{liu2025multi,
  author  = {Liu, Shuze Daniel and Chen, Claire and Zhang, Shangtong},
  title   = {Efficient Multi-Policy Evaluation for Reinforcement Learning},
  journal = {Proceedings of the AAAI Conference on Artificial Intelligence},
  volume  = {39},
  number  = {18},
  pages   = {18951--18959},
  year    = {2025},
  doi     = {10.1609/aaai.v39i18.34086}
}

@inproceedings{chen2025caesar,
  author    = {Chen, Yilei and Pacchiano, Aldo and Paschalidis, Ioannis},
  title     = {Multiple-Policy Evaluation via Density Estimation},
  booktitle = {Proceedings of the 42nd International Conference on Machine Learning},
  series    = {Proceedings of Machine Learning Research},
  volume    = {267},
  pages     = {8794--8820},
  year      = {2025},
  publisher = {PMLR}
}

@inproceedings{bibaut2021,
  author    = {Bibaut, Aur{\'e}lien and Dimakopoulou, Maria and Kallus, Nathan and Chambaz, Antoine and van der Laan, Mark},
  title     = {Post-Contextual-Bandit Inference},
  booktitle = {Advances in Neural Information Processing Systems},
  volume    = {34},
  pages     = {28548--28559},
  year      = {2021}
}

@article{chen2026robust,
  author        = {Chen, Claire and Liu, Shuze Daniel and Luo, Licheng and Chandra, Rohan and Jiang, Nan and Zhang, Shangtong},
  title         = {Robust Data-Collection Policy Learning for Low-Variance Online Policy Evaluation},
  journal       = {arXiv preprint arXiv:2608.24146},
  year          = {2026}
}

@inproceedings{jain2024,
  author    = {Jain, Arushi and Hanna, Josiah P. and Precup, Doina},
  title     = {Adaptive Exploration for Data-Efficient General Value Function Evaluations},
  booktitle = {Advances in Neural Information Processing Systems},
  volume    = {37},
  pages     = {18783--18814},
  year      = {2024},
  doi       = {10.52202/079017-2107}
}

@inproceedings{liu2025safe,
  author    = {Liu, Shuze Daniel and Chen, Claire and Zhang, Shangtong},
  title     = {Safe Online Policy Evaluation for Reinforcement Learning},
  booktitle = {International Conference on Learning Representations},
  year      = {2025}
}

@inproceedings{russo2025,
  author    = {Russo, Alessio and Pacchiano, Aldo},
  title     = {Adaptive Exploration for Multi-Reward Multi-Policy Evaluation},
  booktitle = {Proceedings of the 42nd International Conference on Machine Learning},
  series    = {Proceedings of Machine Learning Research},
  volume    = {267},
  pages     = {48931--48956},
  year      = {2025},
  publisher = {PMLR}
}

@article{waudbysmith2024,
  author    = {Waudby-Smith, Ian and Wu, Lili and Ramdas, Aaditya and Karampatziakis, Nikos and Mineiro, Paul},
  title     = {Anytime-Valid Off-Policy Inference for Contextual Bandits},
  journal   = {ACM/IMS Journal of Data Science},
  volume    = {1},
  number    = {3},
  articleno = {23},
  pages     = {1--36},
  year      = {2024},
  doi       = {10.1145/3643693}
}

@inproceedings{zhu2022,
  author    = {Zhu, Ruihao and Kveton, Branislav},
  title     = {Safe Optimal Design with Applications in Off-Policy Learning},
  booktitle = {Proceedings of the 25th International Conference on Artificial Intelligence and Statistics},
  series    = {Proceedings of Machine Learning Research},
  volume    = {151},
  pages     = {2436--2447},
  year      = {2022},
  publisher = {PMLR}
}

@article{hammer1996,
  author  = {Hammer, Scott M. and Katzenstein, David A. and Hughes, Michael D. and
             Gundacker, Holly and Schooley, Robert T. and Haubrich, Richard H. and
             Henry, W. Keith and Lederman, Michael M. and Phair, John P. and
             Niu, Manette and Hirsch, Martin S. and Merigan, Thomas C.},
  title   = {A Trial Comparing Nucleoside Monotherapy with Combination Therapy in
             {HIV}-Infected Adults with {CD4} Cell Counts from 200 to 500 per Cubic Millimeter},
  journal = {New England Journal of Medicine},
  volume  = {335}, number = {15}, pages = {1081--1090}, year = {1996},
  doi     = {10.1056/NEJM199610103351501}
}

@article{fan2017,
  author  = {Fan, Caiyun and Lu, Wenbin and Song, Rui and Zhou, Yong},
  title   = {Concordance-Assisted Learning for Estimating Optimal Individualized
             Treatment Regimes},
  journal = {Journal of the Royal Statistical Society: Series B (Statistical Methodology)},
  volume  = {79}, number = {5}, pages = {1565--1582}, year = {2017},
  doi     = {10.1111/rssb.12216}
}

@misc{hammer1996uci,
  author       = {Hammer, Scott M. and Katzenstein, David A. and Hughes, Michael D. and
                  Gundacker, Holly and Schooley, Robert T. and Haubrich, Richard H. and
                  Henry, W. Keith and Lederman, Michael M. and Phair, John P. and
                  Niu, Manette and Hirsch, Martin S. and Merigan, Thomas C.},
  title        = {{AIDS} Clinical Trials Group Study 175},
  year         = {1996},
  howpublished = {UCI Machine Learning Repository},
  doi          = {10.24432/C5ZG8F}
}

@book{fedorov1972,
  author    = {Fedorov, Valerii V.},
  translator= {Studden, W. J. and Klimko, E. M.},
  title     = {Theory of Optimal Experiments},
  publisher = {Academic Press},
  address   = {New York},
  year      = {1972},
  isbn      = {978-0-12-250750-2}
}

@article{wynn1970,
  author    = {Wynn, Henry P.},
  title     = {The Sequential Generation of {$D$}-Optimum Experimental Designs},
  journal   = {The Annals of Mathematical Statistics},
  volume    = {41},
  number    = {5},
  pages     = {1655--1664},
  year      = {1970},
  doi       = {10.1214/aoms/1177696809}
}

@article{chernoff1959,
  author    = {Chernoff, Herman},
  title     = {Sequential Design of Experiments},
  journal   = {The Annals of Mathematical Statistics},
  volume    = {30},
  number    = {3},
  pages     = {755--770},
  year      = {1959},
  doi       = {10.1214/aoms/1177706205}
}

@article{frank1956,
  author  = {Frank, Marguerite and Wolfe, Philip},
  title   = {An algorithm for quadratic programming},
  journal = {Naval Research Logistics Quarterly},
  volume  = {3},
  number  = {1--2},
  pages   = {95--110},
  year    = {1956}
}

@inproceedings{jaggi2013,
  author    = {Jaggi, Martin},
  title     = {Revisiting {Frank--Wolfe}: Projection-free sparse convex optimization},
  booktitle = {Proceedings of the 30th International Conference on Machine Learning (ICML)},
  publisher = {PMLR},
  volume    = {28},
  number    = {1},
  pages     = {427--435},
  year      = {2013}
}

@article{lai2020mixture,
  author  = {Lai, Jinlin and Zou, Lixin and Song, Jiaxing},
  title   = {Optimal Mixture Weights for Off-Policy Evaluation with Multiple Behavior Policies},
  journal = {arXiv preprint arXiv:2011.14359},
  year    = {2020},
  eprint  = {2011.14359},
  archivePrefix = {arXiv}
}

@inproceedings{gao2023hope,
  author    = {Gao, Ge and Ju, Song and Sanz-Ausin, Markel and Chi, Min},
  title     = {{HOPE}: Human-Centric Off-Policy Evaluation for E-Learning and Healthcare},
  booktitle = {Proceedings of the 22nd International Conference on Autonomous Agents and Multiagent Systems (AAMAS)},
  year      = {2023}
}

@article{antos2010,
  author  = {Antos, Andr{\'a}s and Grover, Varun and Szepesv{\'a}ri, Csaba},
  title   = {Active Learning in Heteroscedastic Noise},
  journal = {Theoretical Computer Science},
  volume  = {411},
  number  = {29--30},
  pages   = {2712--2728},
  year    = {2010},
  doi     = {10.1016/j.tcs.2010.04.007}
}

@inproceedings{garivier2016,
  author    = {Garivier, Aur{\'e}lien and Kaufmann, Emilie},
  title     = {Optimal Best Arm Identification with Fixed Confidence},
  booktitle = {Proceedings of the 29th Annual Conference on Learning Theory},
  series    = {Proceedings of Machine Learning Research},
  volume    = {49},
  pages     = {998--1027},
  year      = {2016},
  publisher = {PMLR}
}

@article{heowen2014,
  title        = {Optimal mixture weights in multiple importance sampling},
  author       = {He, Hera Y. and Owen, Art B.},
  journal = {arXiv preprint arXiv:1411.3954},
  year         = {2014},
  eprint       = {1411.3954},
  archivePrefix = {arXiv},
  primaryClass = {stat.CO},
  url          = {https://arxiv.org/abs/1411.3954}
}

@article{etore2010,
  author  = {{\'E}tor{\'e}, Pierre and Jourdain, Benjamin},
  title   = {Adaptive Optimal Allocation in Stratified Sampling Methods},
  journal = {Methodology and Computing in Applied Probability},
  volume  = {12},
  number  = {3},
  pages   = {335--360},
  year    = {2010},
  doi     = {10.1007/s11009-008-9108-0}
}

@inproceedings{dai2023,
  author    = {Dai, Jessica and Gradu, Paula and Harshaw, Christopher},
  title     = {{CLIP-OGD}: An Experimental Design for Adaptive Neyman Allocation in Sequential Experiments},
  booktitle = {Advances in Neural Information Processing Systems},
  volume    = {36},
  year      = {2023}
}

\appendix
\renewcommand{\thesection}{A\arabic{section}}

\section{Notation and Overview}\label{app:notation}

\begin{singlespace}
\small
\begin{longtable}{p{0.29\textwidth}p{0.63\textwidth}}
\toprule
Symbol & Meaning \\
\midrule
\endfirsthead
\toprule
Symbol & Meaning \\
\midrule
\endhead
\bottomrule
\endlastfoot
$K,M$ & Numbers of loggers and target policies. \\
$\mu_k,\pi_j$ & Logger $k$ and target policy $j$. \\
$J_j,q,\sigma^2$ & Value of target $j$, and the conditional mean and variance of the reward. \\
$c_k,\mathcal K_t$ & Cost of logger $k$, and the set of loggers affordable at step $t$. \\
$N_t,C_t,B,T_B$ & Counts and money spent after $t$ purchases, the budget, and the number of purchases when the budget is spent. \\
$\mathcal S_K,\theta$ & Probability simplex, and the sample fractions $\theta=N/t$ in it. \\
$m_\theta,\widehat\theta_B$ & Pooled behavior probability at composition $\theta$, and the final sample fractions. \\
$\mathcal S,\kappa(\theta)$ & Union of the target-policy supports, and the smallest pooled probability on it at composition $\theta$. \\
$\kappa_\star,\kappa_{\mathrm{all}}$ & Lower bound on $\kappa(\theta^\star)$ in Assumption~\ref{ass:pooled}, and the smallest value of $\sum_k\mu_k$ on $\mathcal S$. \\
$\chi_t,\chi_0,\upsilon$ & Forced-sampling threshold $\chi_t=\chi_0t^{-\upsilon}$. \\
$\mathcal C(e),\mathcal C_+$ & Designs whose coverage $\boldsymbol\mu^\top v$ is at least $e$ on $\mathcal S$, and their union over $e>0$. \\
$z_N,\mathcal C$ & Design normalized by cost, $N/c^\top N$, and the set $\{v\ge0:c^\top v=1\}$ that contains it. \\
$G_j,G$ & Criterion for target $j$ at counts $N$, and its maximum over targets. \\
$\Delta_{j,k},\Delta_k$ & Reduction in $G_j$ and in $G$ from one more purchase from logger $k$. \\
$\mathcal G_j,\mathcal G$ & Cost-normalized criterion for target $j$, and its maximum over targets. \\
$z^\star,\theta^\star$ & Minimizer of $\mathcal G$ over $\mathcal C$, and the corresponding sample fractions. \\
$\Phi_\rho,\rho_t,\rho_0,\alpha$ & Log-sum-exp approximation of $\mathcal G$, its smoothing parameter $\rho_t=\rho_0C_t^{-\alpha}$, and the constants that set it. \\
$\widehat\omega_{j,t},\widehat S_{t,k}$ & Estimated weights on the targets and estimated index of logger $k$. \\
$\omega_{j,t},S_{t,k}$ & The same weights and index computed from the true criterion. \\
$r_t,\beta_r$ & Accuracy of the estimated criterion (Assumption~\ref{ass:nuisance}) and its rate exponent. \\
$\bar N_t,\bar C_t,\bar z_t$ & Counts, money spent, and normalized design of the infinite-budget sequence, in which every logger is always affordable. \\
$\widehat J_B,\widehat\Sigma_B$ & Final estimator and its covariance estimate. \\
$\Sigma_A,\Sigma_L,L_j$ & Covariance of the pooled and of the logger-specific estimator at a fixed composition, and the $j$th diagonal entry of the latter. \\
$\psi_i(\theta),\widetilde J_B,\widetilde J_B^\star$ & Score of observation $i$ at composition $\theta$, and the score averages at $\widehat\theta_B$ and at $\theta^\star$. \\
$\mathcal R_{B,j},r_B$ & Error from using the estimated regression in the final estimator, and the radius in Assumption~\ref{ass:agg-orth}. \\
\end{longtable}
\end{singlespace}

The appendices are organized as follows. Appendix~\ref{app:prelim} contains the derivations used
in Section~\ref{sec:method}, and Appendices~\ref{app:convergence} and \ref{app:inference} prove the
results of Sections~\ref{sec:convergence} and \ref{sec:adaptive-inference}. Auxiliary results are
stated where they are first needed; those with longer proofs are proved in
Appendix~\ref{app:technical}, in the order of logical dependence. Appendix~\ref{app:blocksplit}
treats the block-split estimator, and Appendix~\ref{app:experiments} gives the details of the
numerical studies.

\section{Derivations for Section~\ref{sec:method}}\label{app:prelim}

\subsection{Derivation of \eqref{eq:gain}}\label{app:gain}

Fix a logger $k$, and write $T=\mathbf 1^\top N$. Then
\begin{align*}
\Delta_{j,k}(N)
&=
G_j(N)-G_j(N+e_k) \\
&= \frac{V_j}{T+\eta_0}
+
\mathbb E_X
\sum_{a\in\mathcal A}
\frac{W_j(X,a)}{D_N(X,a)} - \frac{V_j}{T+1+\eta_0}
-
\mathbb E_X
\sum_{a\in\mathcal A}
\frac{W_j(X,a)}{D_{N+e_k}(X,a)} \\
&= \frac{V_j}{(T+\eta_0)(T+1+\eta_0)}
+\mathbb E_X\sum_{a\in\mathcal A}
\frac{W_j(X,a)\mu_k(a\mid X)}{D_N(X,a)\{D_N(X,a)+\mu_k(a\mid X)\}},
\end{align*}
where the last line uses $D_{N+e_k}=D_N+\mu_k$ and the identity
$u^{-1}-(u+w)^{-1}=w\{u(u+w)\}^{-1}$, once with $u=T+\eta_0$ and $w=1$ and once with $u=D_N(x,a)$ and
$w=\mu_k(a\mid x)$. Taking the difference inside $\mathbb E_X$ is justified because $W_j\ge0$ and both
integrands are bounded by $W_j(X,a)/\eta$, which is integrable.

\subsection{Orders of $G_j$ and $\Delta_{j,k}$}\label{app:order}

Let $N\neq0$, $T=\sum_kN_k$, and $\theta=N/T$. Since $\mu_k\le1$ and $m_\theta\ge\kappa(\theta)$ on
$\mathcal S$,
\[
\eta+T\kappa(\theta)\le D_N(x,a)=\eta+T\,m_\theta(a\mid x)\le\eta+T
\qquad\text{for }(x,a)\in\mathcal S .
\]
Only pairs in $\mathcal S$ contribute to \eqref{eq:Gj} and \eqref{eq:gain}, because $W_j(x,a)=0$
whenever $\pi_j(a\mid x)=0$, and $\bar W_j=\mathbb E_X\sum_aW_j(X,a)$ is finite since
$\sigma^2\le R_{\max}^2$. Thus
\begin{equation*}
\frac{V_j}{T+\eta_0}+\frac{\bar W_j}{\eta+T}
\ \le\ G_j(N)\ \le\
\frac{V_j}{T+\eta_0}+\frac{\bar W_j}{\eta+T\kappa(\theta)},
\end{equation*}
and, since $D_N\{D_N+\mu_k\}\ge\{\eta+T\kappa(\theta)\}^2$ on $\mathcal S$,
\begin{equation*}
\Delta_{j,k}(N)\ \le\ \frac{\bar W_j}{\{\eta+T\kappa(\theta)\}^2}+\frac{V_j}{(T+\eta_0)(T+1+\eta_0)}.
\end{equation*}
Along any sequence of count vectors for which $\kappa(\theta)$ stays bounded away from zero, it
follows that $G_j(N)=O(T^{-1})$ and $\Delta_{j,k}(N)=O(T^{-2})$ as $T\to\infty$; the lower bound shows
that the first order is exact when $V_j+\bar W_j>0$.

\section{Proofs for Section~\ref{sec:convergence}}\label{app:theory}\label{app:convergence}

This appendix proves Theorem~\ref{thm:design-optimality} and Proposition~\ref{prop:fw-step}. We
first establish parts (i) and (ii) of the theorem along the infinite-budget sequence of
Section~\ref{sec:convergence}, in which every logger is always available, and then transfer them to
the budgeted algorithm. Auxiliary results are stated where they are first needed, and those not proved here are
proved in Appendix~\ref{app:technical}.

\subsection{Proof of Theorem~\ref{thm:design-optimality}}

Our argument goes through the infinite-budget sequence $(\bar K_t,\bar N_t,\bar C_t)_{t\ge n_0}$ of
Section~\ref{sec:convergence}. It follows Algorithm~\ref{alg:acquisition}, but keeps every logger available at
every step. Write $\mathcal T$ for its set of forced steps (those steps $t$ at which $\kappa(\bar\theta_t)<\chi_t$), $\bar z_t=z_{\bar N_t}$, and
$\bar\theta_t=\bar N_t/t$; by \eqref{eq:FW-update},
\[
\bar z_{t+1}=(1-\gamma_t)\bar z_t+\gamma_tv_t,
\qquad v_t=\frac{e_{\bar K_{t+1}}}{c_{\bar K_{t+1}}},
\qquad \gamma_t=\frac{c_{\bar K_{t+1}}}{\bar C_{t+1}} .
\]
Along this sequence, we call a purchase satisfying \eqref{eq:approx-LMO}, with $z_t$ replaced by
$\bar z_t$, a $\delta_t$-approximate conditional-gradient step. The proof rests on two results. The
first is an infinite-budget version of the theorem, proved in Appendix~\ref{app:pf-ipr}; the second
couples the two sequences, and is proved in Appendix~\ref{app:technical}. 
Unlike Theorem~\ref{thm:design-optimality}, which concerns the
actual budgeted algorithm stopped at $T_B$, the following
proposition establishes convergence and rates for the
infinite-budget sequence, in which every logger remains affordable.

\begin{proposition}\label{prop:infinite-path-rates}
Suppose the conditions of Theorem~\ref{thm:design-optimality} hold.
\begin{enumerate}[label=(\roman*),topsep=0pt,itemsep=2pt]
\item Almost surely,
\begin{equation}\label{eq:stability}
\bar z_t\to z^\star
\qquad\text{and}\qquad
\bar\theta_t\to\theta^\star ,
\end{equation}
and only finitely many steps of the infinite-budget sequence are forced.
\item Under the additional conditions of part (ii) of Theorem~\ref{thm:design-optimality}, there are, on an event of probability one, finite random constants
$\bar U_{\mathcal G} $, $\bar U_z$, and $\bar U_\theta$ such that, for all sufficiently large $t$,
\begin{equation}\label{eq:rate}
0\le \mathcal G(\bar z_t)-\mathcal G(z^\star)\le \bar U_{\mathcal G} \bar C_t^{-\beta},\qquad
\|\bar z_t-z^\star\|_2\le \bar U_z\bar C_t^{-\beta/2},\qquad
\|\bar\theta_t-\theta^\star\|_2\le \bar U_\theta\bar C_t^{-\beta/2}.
\end{equation}
\end{enumerate}
\end{proposition}

\begin{lemma}\label{lem:infinite-path-coupling}
Let $C_{n_0}$ be the cost of the initialization, suppose $B>C_{n_0}+c_{\max}$, and apply Algorithm~\ref{alg:acquisition} with budget $B$
to the same observations as the infinite-budget sequence. Define
$\tau_B=\min\{t\ge n_0:B-\bar C_t<c_{\max}\}$ and $L_0=\lceil c_{\max}/c_{\min}\rceil$. The two sequences
agree through time $\tau_B$, and
\[
0\le T_B-\tau_B\le L_0,
\qquad
\|N_{T_B}-\bar N_{T_B}\|_1\le2L_0,
\qquad
|C_{T_B}-\bar C_{T_B}|\le L_0c_{\max}.
\]
Consequently there are finite constants $K_z$ and $K_\theta$ such that, for all sufficiently large
$B$,
\[
\|z_{T_B}-\bar z_{T_B}\|_2\le K_zB^{-1},
\qquad
\left\|\widehat\theta_B-\frac{\bar N_{T_B}}{T_B}\right\|_2\le K_\theta B^{-1}.
\]
\end{lemma}

Given these two results, part (i) is immediate. Since $T_B\ge\lfloor(B-c_{\max})/c_{\max}\rfloor$
tends to infinity, \eqref{eq:stability} gives $\bar z_{T_B}\to z^\star$ and
$\bar N_{T_B}/T_B\to\theta^\star$ almost surely; by Lemma~\ref{lem:infinite-path-coupling},
$z_{T_B}$ and $\widehat\theta_B$ differ from these quantities by $O(B^{-1})$, so they converge
to the same limits.

For part (ii), we need to express the rates \eqref{eq:rate} in terms of $B$. By
Lemma~\ref{lem:infinite-path-coupling},
\[
\bar C_{T_B}\ge C_{T_B}-L_0c_{\max}>B-(L_0+1)c_{\max},
\]
hence $\bar C_{T_B}\ge B/2$ and $\bar C_{T_B}^{-\beta}\le2^{\beta}B^{-\beta}$ for all large $B$;
since $T_B\to\infty$, the bounds \eqref{eq:rate} also hold at $t=T_B$. Now split
\[
\mathcal G(z_{T_B})-\mathcal G(z^\star)
=\big[\mathcal G(z_{T_B})-\mathcal G(\bar z_{T_B})\big]+\big[\mathcal G(\bar z_{T_B})-\mathcal G(z^\star)\big].
\]
For large $B$, Lemma~\ref{lem:switch} gives $\kappa(\bar N_{T_B}/T_B)\ge\kappa_\star/2$. With
$e_\star=\min\{\kappa_\star/(4c_{\max}),1/2\}$, it follows that $\bar z_{T_B}\in\mathcal C(2e_\star)$. Because
$|\boldsymbol\mu(x,a)^\top(v-w)|\le\sqrt K\|v-w\|_2$, Lemma~\ref{lem:infinite-path-coupling} then puts
$z_{T_B}$ in $\mathcal C(e_\star)$ as well. Each $\mathcal G_j$ has a bounded gradient on the convex set
$\mathcal C(e_\star)$, so $\mathcal G$ is Lipschitz there with constant
$L_{\mathcal G}=\max_j\sup_{\mathcal C(e_\star)}\|\nabla \mathcal G_j\|_2$. With Lemma~\ref{lem:infinite-path-coupling},
this bounds the first bracket, and Proposition~\ref{prop:infinite-path-rates} bounds the second:
\[
\mathcal G(z_{T_B})-\mathcal G(\bar z_{T_B})\le\frac{L_{\mathcal G}K_z}B,
\qquad
\mathcal G(\bar z_{T_B})-\mathcal G(z^\star)\le\bar U_{\mathcal G}\bar C_{T_B}^{-\beta}\le2^{\beta}\bar U_{\mathcal G}B^{-\beta}.
\]
As $\beta\le1/2$,
\[
0\le \mathcal G(z_{T_B})-\mathcal G(z^\star)\le U_{\mathcal G} B^{-\beta},\qquad
U_{\mathcal G}=L_{\mathcal G}K_z+2^{\beta}\bar U_{\mathcal G} .
\]
For the sample fractions, Lemma~\ref{lem:infinite-path-coupling} and
Proposition~\ref{prop:infinite-path-rates} give
\[
\|\widehat\theta_B-\theta^\star\|_2
\le\Big\|\widehat\theta_B-\frac{\bar N_{T_B}}{T_B}\Big\|_2+\|\bar\theta_{T_B}-\theta^\star\|_2
\le\frac{K_\theta}{B}+2^{\beta/2}\bar U_\theta B^{-\beta/2}
\le U_\theta B^{-\beta/2},
\]
with $U_\theta=K_\theta+2^{\beta/2}\bar U_\theta$.

\subsection{Proof of Proposition~\ref{prop:infinite-path-rates}}\label{app:pf-ipr}

We first collect the results that the proof relies on: strong convexity of $\mathcal G$, five properties of
the smoothed criterion, control of the pooled coverage under forced sampling, and an elementary fact
about recursions. Their proofs are given in Appendix~\ref{app:technical}. The proof also uses
Lemma~\ref{lem:direction} of Appendix~\ref{app:pf-fw}, by which unforced purchases are approximate
conditional-gradient steps.

Recall that $\mathcal G=\max_j\mathcal G_j$, where
\[
\mathcal G_j(v)
=
\frac{V_j}{\mathbf 1^\top v}
+
\mathbb E_X
\sum_{a\in\mathcal A}
\frac{W_j(X,a)}
{\boldsymbol{\mu}(X,a)^\top v},
\qquad
\boldsymbol{\mu}(x,a)=(\mu_1(a\mid x),\ldots,\mu_K(a\mid x))^\top,
\qquad v\in\mathcal C .
\]
To bound the denominators, we restrict $v$ to designs with positive coverage on $\mathcal S$,
\[
\mathcal C_+
=\Big\{v\in\mathcal C:\operatorname*{ess\,inf}_{(x,a)\in\mathcal S}\boldsymbol\mu(x,a)^\top v>0\Big\}
=\bigcup_{e>0}\mathcal C(e),
\qquad
\mathcal C(e)
=\Big\{v\in\mathcal C:\operatorname*{ess\,inf}_{(x,a)\in\mathcal S}\boldsymbol\mu(x,a)^\top v\ge e\Big\}.
\]
These sets are convex because $\boldsymbol\mu(x,a)^\top v$ is linear in $v$. Moreover,
$\boldsymbol\mu(x,a)^\top v=(\mathbf 1^\top v)\,m_{v/\mathbf 1^\top v}(a\mid x)$ and
$\mathbf 1^\top v\ge c_{\max}^{-1}$ on $\mathcal C$, and so every $v\in\mathcal C$ with
$\kappa(v/\mathbf 1^\top v)>0$ lies in $\mathcal C(e)$ with $e=\kappa(v/\mathbf 1^\top v)/c_{\max}$.

\begin{proposition}\label{prop:strong-convexity}
Under Assumptions~\ref{ass:known} and \ref{ass:identifiability}, for every $v\in\mathcal C_+$ and
every $h$ with $c^\top h=0$,
\begin{equation}\label{eq:hessian-lower}
h^\top\nabla^2\mathcal G_j(v)h\ge2c_{\min}^3\lambda_j\|h\|_2^2 .
\end{equation}
Consequently each $\mathcal G_j$, and therefore $\mathcal G$, is $m$-strongly convex along every segment contained in
$\mathcal C_+$, with $m=2c_{\min}^3\min_j\lambda_j>0$.
\end{proposition}

Strong convexity enters the argument only through the following quadratic growth around $z^\star$,
which converts convergence of the criterion into convergence of the design.

\begin{lemma}\label{lem:quadratic-growth}
Under Assumptions~\ref{ass:known}, \ref{ass:pooled}, and \ref{ass:identifiability},
$\mathcal G(v)-\mathcal G(z^\star)\ge\tfrac m2\|v-z^\star\|_2^2$ for every $v\in\mathcal C$, so $z^\star$ is the unique
minimizer of $\mathcal G$ over $\mathcal C$.
\end{lemma}

The recursion below uses five properties of $\Phi_\rho$ in \eqref{eq:Phi}, where
\[
\omega_j(v;\rho)=\frac{\exp\{\mathcal G_j(v)/\rho\}}{\sum_\ell\exp\{\mathcal G_\ell(v)/\rho\}}.
\]

\begin{lemma}\label{lem:smoothed}
Under Assumption~\ref{ass:known}, parts (a), (b), (c), and (e) hold on $\mathcal C_+$, and part (d)
holds on $\mathcal C(e)$ for every $e\in(0,1]$.
\begin{enumerate}[label=(\alph*),topsep=0pt,itemsep=1pt]
\item $\mathcal G(v)\le\Phi_\rho(v)\le \mathcal G(v)+\rho\log M$.
\item $\nabla\Phi_\rho(v)=\sum_j\omega_j(v;\rho)\nabla \mathcal G_j(v)$.
\item $\Phi_\rho$ is convex.
\item $\nabla\Phi_\rho$ is Lipschitz on $\mathcal C(e)$ with constant at most $L_0(e)+L_1(e)/\rho$,
where $L_0(e)=2KR_{\max}^2(c_{\max}^3+e^{-3})$ and $L_1(e)=KR_{\max}^4(c_{\max}^2+e^{-2})^2$.
\item For fixed $v$, $\rho\mapsto\Phi_\rho(v)$ is nondecreasing, so $\rho'\le\rho$ implies
$\Phi_{\rho'}\le\Phi_\rho$.
\end{enumerate}
\end{lemma}

Next, we control the coverage under forced sampling. The two lemmas below concern any sequence
generated by Algorithm~\ref{alg:acquisition} during the steps at which every logger is affordable,
in particular the infinite-budget sequence. Write $\theta_t=N_t/t$, call step $t$ forced if $\kappa(\theta_t)<\chi_t$, and
let $\mathcal T$ be the set of forced steps. For the analysis, let
\[
\kappa_{\mathrm{all}}
=
K\kappa(\mathbf 1/K)
=
\operatorname*{ess\,inf}_{(x,a)\in\mathcal S}
\sum_{k=1}^K\mu_k(a\mid x),
\]
so that the condition $0<\chi_0\le\kappa(\mathbf 1/K)/8$ of Section~\ref{sec:algorithm} reads
$0<\chi_0\le\kappa_{\mathrm{all}}/(8K)$. By definition of $\kappa_{\mathrm{all}}$, for every
$\theta\in\mathcal S_K$ and every $(x,a)\in\mathcal S$,
\begin{equation}\label{eq:share-coverage}
m_\theta(a\mid x)
\ge
\left(\min_k\theta_k\right)
\sum_{k=1}^K\mu_k(a\mid x),
\qquad
\kappa(\theta)
\ge
\kappa_{\mathrm{all}}\min_k\theta_k.
\end{equation}

\begin{lemma}\label{lem:coverage}
Suppose $0<\chi_0\le\kappa_{\mathrm{all}}/(8K)$, $0<\upsilon<1$, and the
initialization purchases $n_0/K\ge4$ observations from each logger. Then, as long as every logger is
affordable,
\begin{enumerate}[label=(\alph*),topsep=0pt,itemsep=1pt]
\item $\kappa(\theta_t)\ge\chi_t/2$ for every $t\ge n_0$;
\item each logger receives at most $\chi_tt/\kappa_{\mathrm{all}}+1$ forced purchases at steps $s\le t$, and
$\sum_{t\in\mathcal T}\gamma_t<\infty$, where $\gamma_t=c_{K_{t+1}}/C_{t+1}$.
\end{enumerate}
\end{lemma}

Summing part (b) over the $K$ loggers, the number of forced purchases up to time $t$ is at most
$K(\chi_tt/\kappa_{\mathrm{all}}+1)=O(t^{1-\upsilon})$, as stated in Section~\ref{sec:algorithm}.

\begin{lemma}\label{lem:switch}
Under Assumption~\ref{ass:pooled} and the conditions of Lemma~\ref{lem:coverage}, if
$\theta_t\to\theta^\star$ along a sequence in which
every logger is always affordable, then only finitely many steps are forced, and
$\kappa(\theta_t)\ge \kappa_\star/2$ for all sufficiently large $t$.
\end{lemma}

\paragraph{Computing pooled coverage.}\label{app:coverage-computation}
When the context support is finite and known, $\kappa(\theta)$ can be computed by taking the minimum of $m_\theta(a\mid x)$ over all target-relevant context-action pairs.

For general context spaces, forced sampling may instead use
\[
\widetilde\kappa(\theta)
=
\operatorname*{ess\,inf}_{(x,a)\in\mathcal S}
\sum_{k=1}^K\theta_k\widetilde\mu_k(a\mid x),
\]
where the known functions $\widetilde\mu_k$ satisfy $0\le\widetilde\mu_k\le\mu_k$. For example, we can partition $\mathcal S$ into finitely many disjoint measurable regions $\mathcal S_1,\ldots,\mathcal S_R$ of positive measure. For each logger $k$ and region $r$, choose a known constant
\[
0\le\underline\mu_{k,r}
\le
\operatorname*{ess\,inf}_{(x,a)\in\mathcal S_r}
\mu_k(a\mid x),
\]
and set $\widetilde\mu_k(a\mid x)=\underline\mu_{k,r}$ for $(x,a)\in\mathcal S_r$. Then
$
\widetilde\kappa(\theta)
=
\min_{1\le r\le R}
\sum_{k=1}^K\theta_k\underline\mu_{k,r}
\le\kappa(\theta).
$
Thus, once the regional bounds are computed, evaluating $\widetilde\kappa(\theta)$ requires only a finite minimum of weighted sums, with $O(KR)$ operations. For structured logging policies on bounded regions, such bounds can be obtained using interval methods \citep{gtoth2021lowerbounds}. Individual regional bounds may be zero, so this construction does not require every logger to cover every target-relevant action.

This alternative requires $\widetilde\kappa(\theta^\star)>0$ and $0<\chi_0\le\widetilde\kappa(\mathbf 1/K)/8$.
Because $\widetilde\kappa$ is an infimum of the nonnegative linear functions
$\theta\mapsto\sum_k\theta_k\widetilde\mu_k(a\mid x)$, it has the three properties of $\kappa$ that the
coverage proofs use:
\[
\widetilde\kappa\Big(\frac{t\theta+e_k}{t+1}\Big)\ge\frac t{t+1}\widetilde\kappa(\theta),
\qquad
\widetilde\kappa(\theta)\ge K\widetilde\kappa(\mathbf 1/K)\min_k\theta_k,
\qquad
|\widetilde\kappa(\theta)-\widetilde\kappa(\theta')|\le\|\theta-\theta'\|_1,
\]
the last because $\widetilde\mu_k\le1$. The proofs of Lemmas~\ref{lem:coverage} and \ref{lem:switch}
therefore go through with three substitutions. A step is forced if $\widetilde\kappa(\theta_t)<\chi_t$; $K\widetilde\kappa(\mathbf 1/K)$ replaces $\kappa_{\mathrm{all}}$, and \eqref{eq:share-coverage} holds with $\widetilde\mu_k$ and $\widetilde\kappa$ in place of $\mu_k$ and $\kappa$. Finally, $\widetilde\kappa(\theta^\star)$ replaces $\kappa_\star$ in the proof of Lemma~\ref{lem:switch}. Since $\kappa\ge\widetilde\kappa$, the remaining arguments are unchanged.

Finally, we need an elementary fact about recursions.

\begin{lemma}\label{lem:recursion}
Let $h_t\ge0$ and $\gamma_t\in(0,1]$ with $\sum_t\gamma_t=\infty$, and let $\mathcal T$ be a set of
indices with $\sum_{t\in\mathcal T}\gamma_t<\infty$. Suppose that, for all sufficiently large $t$, $h_{t+1}\le(1-\gamma_t)h_t+\gamma_tq_t$ for $t\notin\mathcal T$, where $q_t\to0$, and that
$h_{t+1}\le(1+\gamma_t)h_t+\gamma_tb$ for $t\in\mathcal T$, where $b<\infty$. Then $h_t\to0$.
\end{lemma}

We start with part (i). Our main task is to show that the smoothed excess
$\bar h_t=\Phi_{\rho_t}(\bar z_t)-\mathcal G(z^\star)$ converges to zero, since the remaining claims then
follow quickly. By Lemma~\ref{lem:smoothed}(a), $0\le \mathcal G(\bar z_t)-\mathcal G(z^\star)\le\bar h_t$, hence
Lemma~\ref{lem:quadratic-growth} gives $\bar z_t\to z^\star$. Because $v\mapsto v/\mathbf 1^\top v$ is
Lipschitz on $\mathcal C$, where $\mathbf 1^\top v\ge c_{\max}^{-1}$, it follows that
$\bar\theta_t\to\theta^\star$; Lemma~\ref{lem:switch} then shows that only finitely many steps are
forced. All statements below hold pathwise on the event of probability one on which
Assumption~\ref{ass:nuisance} holds. Since $\bar C_t\le\bar C_{n_0}+c_{\max}t$, the condition
$t^{\alpha+2\upsilon}r_t\to0$ gives $\bar C_t^{\alpha+2\upsilon}r_t\to0$.

We show that $\bar h_t\to0$ by verifying the conditions of Lemma~\ref{lem:recursion}. Let
$e_t=\chi_t/(4c_{\max})$, which is at most $1/c_{\max}$ because $\chi_0\le\kappa_{\mathrm{all}}/(8K)\le1/8$.
Lemma~\ref{lem:coverage}(a) gives $\kappa(\bar\theta_t)\ge\chi_t/2$, so that $\bar z_t\in\mathcal C(2e_t)$;
since $\bar z_{t+1}\ge(1-\gamma_t)\bar z_t$ and $\gamma_t\le1/2$ for large $t$, the segment from
$\bar z_t$ to $\bar z_{t+1}$ lies in $\mathcal C(e_t)$. Write $L_t=L_0(e_t)+L_1(e_t)/\rho_t$ for the
Lipschitz constant of Lemma~\ref{lem:smoothed}(d) on $\mathcal C(e_t)$, and $d_{\mathcal C}$ for the
diameter of $\mathcal C$, so that $\|v_t-\bar z_t\|_2\le d_{\mathcal C}$. As $\bar z_{t+1}-\bar z_t=\gamma_t(v_t-\bar z_t)$, the usual descent
inequality for functions with a Lipschitz gradient gives
\begin{equation}
\label{eq:descent-step}
\Phi_{\rho_t}(\bar z_{t+1})
\le\Phi_{\rho_t}(\bar z_t)+\gamma_t\langle\nabla\Phi_{\rho_t}(\bar z_t),v_t-\bar z_t\rangle
+\tfrac12d_{\mathcal C}^2L_t\gamma_t^2 .
\end{equation}

The choice of logger enters only through the middle term. At an unforced step,
Lemma~\ref{lem:direction} and $z^\star\in\mathcal C$ bound the first part of that term, while
convexity of $\Phi_{\rho_t}$ on $\mathcal C_+$, which contains $\bar z_t$, $z^\star$, and the segment
between them, bounds the second:
\[
\langle\nabla\Phi_{\rho_t}(\bar z_t),v_t\rangle\le\langle\nabla\Phi_{\rho_t}(\bar z_t),z^\star\rangle+\delta_t,
\qquad
\langle\nabla\Phi_{\rho_t}(\bar z_t),z^\star-\bar z_t\rangle\le\Phi_{\rho_t}(z^\star)-\Phi_{\rho_t}(\bar z_t).
\]
Plugging both bounds into \eqref{eq:descent-step}, and using $\Phi_{\rho_{t+1}}\le\Phi_{\rho_t}$ (as $\rho_{t+1}\le\rho_t$) and
$\Phi_{\rho_t}(z^\star)\le \mathcal G(z^\star)+\rho_t\log M$ from Lemma~\ref{lem:smoothed}(e) and (a), we
obtain
\begin{equation}
\label{eq:h-recursion}
\bar h_{t+1}\le(1-\gamma_t)\bar h_t+\gamma_tq_t,
\qquad
q_t=\rho_t\log M+\delta_t+\tfrac12d_{\mathcal C}^2L_t\gamma_t,
\qquad t\notin\mathcal T .
\end{equation}
At a forced step, we only use that purchasing never increases the criterion to first order. The partial
derivatives of each $\mathcal G_j$ are nonpositive and $v_t\ge0$, so that
$\langle\nabla\Phi_{\rho_t}(\bar z_t),v_t\rangle\le0$; meanwhile, homogeneity of degree $-1$ gives
\[
-\langle\nabla\Phi_{\rho_t}(\bar z_t),\bar z_t\rangle
=\sum_j\omega_j(\bar z_t;\rho_t)\mathcal G_j(\bar z_t)
\le\Phi_{\rho_t}(\bar z_t).
\]
Thus \eqref{eq:descent-step} and Lemma~\ref{lem:smoothed}(e) give
\begin{equation}
\label{eq:h-recursion-forced}
\bar h_{t+1}\le(1+\gamma_t)\bar h_t+\gamma_t\big\{\mathcal G(z^\star)+\tfrac12d_{\mathcal C}^2L_t\gamma_t\big\},
\qquad t\in\mathcal T .
\end{equation}

It remains to check the hypotheses of Lemma~\ref{lem:recursion}. Lemma~\ref{lem:coverage}(b) gives
$\sum_{t\in\mathcal T}\gamma_t<\infty$, and, using $\bar C_t\ge c_{\min}t$ for the second bound,
\[
\gamma_t\ge\frac{c_{\min}}{\bar C_{n_0}+c_{\max}(t+1)},
\qquad
\chi_t^{-1}\le\chi_0^{-1}c_{\min}^{-\upsilon}\bar C_t^{\upsilon};
\]
in particular, $\sum_t\gamma_t=\infty$. Since $e_t\le1/c_{\max}$,
Lemma~\ref{lem:smoothed}(d) also gives
\[
L_0(e_t)\le4KR_{\max}^2e_t^{-3},
\qquad
L_1(e_t)\le4KR_{\max}^4e_t^{-4}.
\]
Combined with $\gamma_t\le c_{\max}/\bar C_t$ and $\rho_t=\rho_0\bar C_t^{-\alpha}$, these bounds
yield finite constants $D_4$ and $D_5$ such that
\[
L_t\gamma_t\le D_4\big(\bar C_t^{3\upsilon-1}+\bar C_t^{4\upsilon+\alpha-1}\big),
\qquad
\delta_t\le D_5\big(\bar C_t^{\alpha+2\upsilon}r_t+\bar C_t^{\alpha+4\upsilon-1}\big);
\]
for the second bound, Lemma~\ref{lem:direction} and Proposition~\ref{prop:nuisance} show that
$\delta_t$ is at most a constant multiple of
$\kappa(\bar\theta_t)^{-2}\bar C_t^{\alpha}r_t+\kappa(\bar\theta_t)^{-4}\bar C_t^{\alpha-1}$. Both
bounds tend to zero because $\upsilon<(1-\alpha)/4$ and $\bar C_t^{\alpha+2\upsilon}r_t\to0$, thus
$q_t\to0$. Finally, the additive term in \eqref{eq:h-recursion-forced} is at most $\gamma_tb$ with
$b=\mathcal G(z^\star)+\sup_t\tfrac12d_{\mathcal C}^2L_t\gamma_t<\infty$, and Lemma~\ref{lem:recursion}
gives $\bar h_t\to0$.

For part (ii), part (i) and Lemma~\ref{lem:switch} provide a finite random $T_0$ after which no step
is forced and $\kappa(\bar\theta_t)\ge\kappa_\star/2$. For $t\ge T_0$, then,
$\bar z_t\in\mathcal C(2e_\star)$ with $e_\star=\min\{\kappa_\star/(4c_{\max}),1/2\}$, and the segment
to $\bar z_{t+1}$ lies in $\mathcal C(e_\star)$ for large $t$. Thus \eqref{eq:h-recursion} holds with
constants $L_0(e_\star)$ and $L_1(e_\star)$ that do not depend on $t$, and we bound the three terms
of $q_t$ in turn. The smoothing bias is $\rho_0(\log M)\bar C_t^{-\alpha}$. For the other two terms,
Lemma~\ref{lem:direction}, applied with $\underline m=\kappa_\star/2$ and the condition $r_t=O(t^{-\beta_r})$ of part
(ii) of Theorem~\ref{thm:design-optimality}, and the bound $\gamma_t\le c_{\max}/\bar C_t$ give
\begin{align*}
\delta_t&\le D_2\big(\bar C_t^{\alpha-1}+\bar C_t^{\alpha-\beta_r}\big),\\
\tfrac12d_{\mathcal C}^2L_t\gamma_t
\le\tfrac12d_{\mathcal C}^2\Big\{L_0(e_\star)+\frac{L_1(e_\star)\bar C_t^{\alpha}}{\rho_0}\Big\}
\frac{c_{\max}}{\bar C_t}
&\le D_3\big(\bar C_t^{-1}+\bar C_t^{\alpha-1}\big).
\end{align*}
Since $\beta$ is no larger than $\alpha$, $1-\alpha$, $\beta_r-\alpha$, or $1$, each of these powers
is at most $\bar C_t^{-\beta}$ once $\bar C_t\ge1$; thus there is a finite $Q$ with
$q_t\le Q\bar C_t^{-\beta}$ for all large $t$.

To conclude, we solve the recursion with this bound. Because
$\gamma_t=(\bar C_{t+1}-\bar C_t)/\bar C_{t+1}$, multiplying \eqref{eq:h-recursion} by
$\bar C_{t+1}$ makes it telescope:
\[
\bar C_{t+1}\bar h_{t+1}\le\bar C_t\bar h_t+(\bar C_{t+1}-\bar C_t)q_t .
\]
Choose $t_0\ge T_0$ beyond which the bound on $q_t$ holds and with
$\bar C_{t_0}\ge\max\{1,c_{\max}\}$. Summing from $t_0$ to $t-1$,
\begin{equation}\label{eq:app-usum}
\bar C_t\bar h_t\le\bar C_{t_0}\bar h_{t_0}
+Q\sum_{s=t_0}^{t-1}(\bar C_{s+1}-\bar C_s)\bar C_s^{-\beta}.
\end{equation}
The sum is a Riemann sum for $\int u^{-\beta}du$: for $s\ge t_0$,
$\bar C_{s+1}\le\bar C_s+c_{\max}\le2\bar C_s$, and so
\[
(\bar C_{s+1}-\bar C_s)\bar C_s^{-\beta}
\le2(\bar C_{s+1}-\bar C_s)\bar C_{s+1}^{-\beta}
\le2\int_{\bar C_s}^{\bar C_{s+1}}u^{-\beta}du .
\]
The sum in \eqref{eq:app-usum} is therefore at most $2\bar C_t^{1-\beta}/(1-\beta)$, and dividing by
$\bar C_t$ gives
\[
\bar h_t\le\frac{\bar C_{t_0}\bar h_{t_0}}{\bar C_t}+\frac{2Q}{1-\beta}\,\bar C_t^{-\beta}
\le \bar U_{\mathcal G} \,\bar C_t^{-\beta},
\]
because $\beta\le1/2<1$. This is the first bound in \eqref{eq:rate}, since
$\mathcal G(\bar z_t)-\mathcal G(z^\star)\le\bar h_t$. Lemma~\ref{lem:quadratic-growth} then gives
$\|\bar z_t-z^\star\|_2^2\le(2\bar U_{\mathcal G}/m)\bar C_t^{-\beta}$, which is the second bound, and the
third follows because $v\mapsto v/\mathbf 1^\top v$ is Lipschitz on $\mathcal C$. The constants depend
on the realized path through $t_0$ and $\bar h_{t_0}$, but not on $t$.

\subsection{Proof of Proposition~\ref{prop:fw-step}}\label{app:pf-fw}

We need to show that each purchase of Algorithm~\ref{alg:acquisition} is a $\delta_t$-approximate
conditional-gradient step with $\delta_t\to0$. Throughout, $N_t$ and $C_t$ denote the counts and
cost of a generic sequence generated by the algorithm, and $z_t=z_{N_t}$. Writing
$g_k=\langle\nabla\Phi_{\rho_t}(z_t),e_k/c_k\rangle$ for the value of the linear approximation at the
vertex of logger $k$, the claim is that the purchased logger $\widehat k$ satisfies
\[
g_{\widehat k}\le\min_kg_k+\delta_t,\qquad\delta_t\to0 .
\]
The algorithm does not observe $g_k$; it purchases from the logger with the largest estimated index
$\widehat S_{t,k}$ of \eqref{eq:score}, so the claim holds if $C_t^2\widehat S_{t,k}$ is close to
$-g_k$ for every $k$. The index differs from $-g_k$ in two ways: it uses estimated rather than true
quantities, and the finite-count criterion rather than the continuous one. We handle the two
separately by passing through the index built from the true finite-count criterion,
\begin{equation}\label{eq:oracle-annealed-score}
\omega_{j,t}
=
\frac{\exp\{C_tG_j(N_t)/\rho_t\}}
{\sum_{\ell=1}^M\exp\{C_tG_\ell(N_t)/\rho_t\}},
\qquad
S_{t,k}
=
\frac{1}{c_k}\sum_{j=1}^M\omega_{j,t}\Delta_{j,k}(N_t).
\end{equation}
Every index weights the gains by exponential weights, so both comparisons rely on the following
sensitivity bound.

\begin{lemma}\label{lem:softmax}
For $\rho>0$ and $y\in\mathbb R^M$, let $\omega(y)_j=\frac{\exp(y_j/\rho)}{\sum_\ell\exp(y_\ell/\rho)}$. Then
$\|\omega(y)-\omega(y')\|_1\le2\rho^{-1}\|y-y'\|_\infty$ for all $y,y'\in\mathbb R^M$.
\end{lemma}

For the finite-count error, $S_{t,k}$ uses $G_j$ and $\Delta_{j,k}$ where the gradient of
$\Phi_{\rho_t}$ uses $\mathcal G_j$ and $\partial_k\mathcal G_j$. After rescaling by $C$ and $C^2$, these quantities
differ by $O(C^{-1})$; combined with Lemma~\ref{lem:softmax}, this bounds the distance between
$C_t^2S_{t,k}$ and $-g_k$.

\begin{lemma}\label{lem:budget-expansion}
Under Assumption~\ref{ass:known}, for every $0<e\le1/c_{\max}$, uniformly over nonzero $N$ with
$z_N\in\mathcal C(e)$, targets $j$, and loggers $k$, with $C=C_N$ and $z=z_N$,
\begin{equation}\label{eq:expansions}
|CG_j(N)-\mathcal G_j(z)|\le K_G(e)C^{-1},
\qquad
|C^2\Delta_{j,k}(N)+\partial_k\mathcal G_j(z)|\le K_\Delta(e)C^{-1},
\end{equation}
where $K_G(e)=R_{\max}^2(\eta_0c_{\max}^2+\eta e^{-2})$ and $K_\Delta(e)$ is given at the end of
the proof.
\end{lemma}

\begin{lemma}\label{lem:oracle-LMO}
Under Assumption~\ref{ass:known}, there is a finite constant $D_1$, depending only on $c_{\min}$,
$c_{\max}$, $R_{\max}$, $\eta$, $\eta_0$, and $\rho_0$, such that at every step with
$z_t\in\mathcal C(e)$ for some $0<e\le1/c_{\max}$, uniformly over loggers,
\begin{equation}\label{eq:LMO-error}
\left|
C_t^2S_{t,k}
+
g_k
\right| =
\left|
C_t^2S_{t,k}
+
\left\langle
\nabla\Phi_{\rho_t}(z_t),
\frac{e_k}{c_k}
\right\rangle
\right|
\le
D_1e^{-4}\left(C_t^{\alpha-1}+C_t^{-1}\right).
\end{equation}
\end{lemma}

For the estimation error, recall that the estimated index uses $\widehat G_{j,t}$ and
$\widehat\Delta_{j,k,t}$ where $S_{t,k}$ uses the true quantities; Assumption~\ref{ass:nuisance}
controls exactly these differences.

\begin{proposition}\label{prop:nuisance}
Under Assumptions~\ref{ass:known} and \ref{ass:nuisance}, there is a finite constant $D_0$, depending
only on $c_{\min}$, $c_{\max}$, and $R_{\max}$, such that, almost surely for all sufficiently large
$t$,
\begin{equation}\label{eq:annealed-score-error}
C_t^2\max_k
\left|\widehat S_{t,k}-S_{t,k}\right|
\le
D_0\left(
r_t+\frac{r_t}{\kappa(\theta_t)^2\rho_t}
\right).
\end{equation}
\end{proposition}

The last lemma combines the two errors.

\begin{lemma}\label{lem:direction}
Suppose Assumption~\ref{ass:known} holds, Assumption~\ref{ass:nuisance} holds along the sequence, every
logger is available at step $t$, and step $t$ is not forced. Then the purchase at step $t$ is a
$\delta_t$-approximate conditional-gradient step with
\[
\delta_t\le2C_t^2\max_k|\widehat S_{t,k}-S_{t,k}|
+2D_1c_{\max}^4\kappa(\theta_t)^{-4}\big(C_t^{\alpha-1}+C_t^{-1}\big),
\]
where $D_1$ is the constant of Lemma~\ref{lem:oracle-LMO}. If, in addition,
$\kappa(\theta_t)\ge\underline m>0$ and $r_t\le Rt^{-\beta_r}$ for all large $t$, then
$\delta_t\le D_2(C_t^{\alpha-1}+C_t^{\alpha-\beta_r})$ for a finite $D_2$.
\end{lemma}

We now prove the proposition. For the first claim, we insert the index $S_{t,k}$ built from the true
criterion. Since $\kappa(\theta_t)\le1$, Proposition~\ref{prop:nuisance} gives, once $C_t\ge1$,
\[
C_t^2\max_k|\widehat S_{t,k}-S_{t,k}|
\le D_0\big(r_t+\rho_0^{-1}\kappa(\theta_t)^{-2}C_t^{\alpha}r_t\big)
\le D_0(1+\rho_0^{-1})\kappa(\theta_t)^{-2}C_t^{\alpha}r_t ,
\]
while Lemma~\ref{lem:oracle-LMO}, applied with $e=\kappa(\theta_t)/c_{\max}$, gives
\[
\max_k|C_t^2S_{t,k}+g_k|\le2D_1c_{\max}^4\kappa(\theta_t)^{-4}C_t^{\alpha-1}.
\]
The triangle inequality yields \eqref{eq:score-gradient} with
$D=\max\{D_0(1+\rho_0^{-1}),2D_1c_{\max}^4\}$. The second claim then follows by the argument in the
proof of Lemma~\ref{lem:direction}, applied with this bound on the index error.

\subsection{Sufficient Conditions for Assumption~\ref{ass:nuisance}}\label{app:sufficient}

We now give sufficient conditions for Assumption~\ref{ass:nuisance}. They concern the estimated
quantities of Section~\ref{sec:selection}, which are defined as follows. Let
$\widehat q_t$ and $\widehat\sigma_t^2$ be estimates of the conditional reward mean and variance from the
first $t$ observations, truncated to $[-R_{\max},R_{\max}]$ and $[0,R_{\max}^2]$, respectively, and let
$\widehat P_tf=t^{-1}\sum_{i=1}^tf(X_i)$ be the empirical distribution of the observed contexts. For each
target policy $\pi_j$, define
\begin{align}
\widehat q_{j,t}(x)
&=\sum_{a\in\mathcal A}\pi_j(a\mid x)\widehat q_t(x,a),
\nonumber\\
\widehat W_{j,t}(x,a)
&=\pi_j(a\mid x)^2\widehat\sigma_t^2(x,a),
\nonumber\\
\widehat V_{j,t}
&=\widehat P_t\{\widehat q_{j,t}^2\}
-\{\widehat P_t\widehat q_{j,t}\}^2.
\label{eq:nuisance-estimates}
\end{align}
Substituting these estimates into \eqref{eq:Gj} and \eqref{eq:gain} gives
\begin{equation}\label{eq:Ghat}
\widehat G_{j,t}
=
\frac{\widehat V_{j,t}}{t+\eta_0}
+
\widehat P_t\sum_{a\in\mathcal A}
\frac{\widehat W_{j,t}(\cdot,a)}{D_{N_t}(\cdot,a)}
\end{equation}
and
\begin{equation}\label{eq:gainhat}
\widehat\Delta_{j,k,t}
=
\widehat P_t\sum_{a\in\mathcal A}
\frac{\widehat W_{j,t}(\cdot,a)\mu_k(a\mid\cdot)}
{D_{N_t}(\cdot,a)\{D_{N_t}(\cdot,a)+\mu_k(a\mid\cdot)\}}
+
\frac{\widehat V_{j,t}}
{(t+\eta_0)(t+1+\eta_0)}.
\end{equation}

Since $G_j$ and
$\Delta_{j,k}$ are linear in $V_j$ and $W_j$, the rate $r_t$ is inherited from the estimates of
these moments and from the replacement of $P_X$ by $\widehat P_t$. In the next lemma, the first two
conditions are rates for the estimated reward moments. The last two control the replacement of
$P_X$ by the empirical distribution of the purchased contexts, which are independent draws from
$P_X$ because each logger is chosen before its context is observed. Conditions of this type are
standard in semiparametric estimation \citep{chernozhukov2018}. Both results below are proved in
Appendix~\ref{app:technical}.

\begin{lemma}\label{lem:plugin}
Under Assumption~\ref{ass:known}, suppose there are deterministic sequences
$r_{V,t},r_{W,t},r_{P,t}\to0$ and $\underline m_t\in(0,1]$
such that, almost surely for all sufficiently large $t$,
$\kappa(\theta_t)\ge\underline m_t$,
\[
\max_j|\widehat V_{j,t}-V_j|\le r_{V,t},
\qquad
\max_j\sum_{a\in\mathcal A}
\|\widehat W_{j,t}(\cdot,a)-W_j(\cdot,a)\|_{L_1(P_X)}
\le r_{W,t},
\]
and, uniformly over $j$ and $k$,
\[
t\Bigg|
(\widehat P_t-P_X)
\sum_a
\frac{\widehat W_{j,t}(\cdot,a)}
{D_{N_t}(\cdot,a)}
\Bigg|
\le r_{P,t},
\]
\[
t^2\Bigg|
(\widehat P_t-P_X)
\sum_a
\frac{\widehat W_{j,t}(\cdot,a)\mu_k(a\mid\cdot)}
{D_{N_t}(\cdot,a)
\{D_{N_t}(\cdot,a)+\mu_k(a\mid\cdot)\}}
\Bigg|
\le r_{P,t}.
\]
Let
\[
a_t=r_{V,t}+r_{P,t}+\underline m_t^{-2}r_{W,t}.
\]
If $a_t\to0$, then Assumption~\ref{ass:nuisance} holds with
$r_t=K'\sup_{s\ge t}a_s$, where
$K'=\max\{c_{\max},c_{\max}^2\}$.
\end{lemma}

When the context space is finite, these conditions can be checked directly.

\begin{corollary}\label{cor:finite-contexts}
Suppose $\mathcal X$ is finite, $\widehat q_t(x,a)$ and
$\widehat\sigma_t^2(x,a)$ are the sample mean and sample
variance of the rewards observed at $(x,a)$, and
$\widehat P_t$ is the empirical distribution of
$X_1,\ldots,X_t$. Under Assumptions~\ref{ass:known}
and \ref{ass:pooled}, with the forced-sampling constants of
Section~\ref{sec:algorithm} and $\upsilon<1/6$,
Assumption~\ref{ass:nuisance} holds along the sequence generated by
Algorithm~\ref{alg:acquisition}, while every logger is
affordable, with
$r_t=K't^{3\upsilon}\sqrt{\log t/t}$, and, if
$\theta_t\to\theta^\star$, also with
$r_t=K'\sqrt{\log t/t}$, for a finite constant $K'$.
\end{corollary}

With the first of these rates, the condition $t^{\alpha+2\upsilon}r_t\to0$ of Theorem~\ref{thm:design-optimality} holds when $\alpha+5\upsilon<1/2$, which is satisfied by the values $\alpha=0.4$ and $\upsilon=0.01$ used in Section~\ref{sec:experiments}.

\section{Proofs for Section~\ref{sec:adaptive-inference}}\label{app:inference}

This appendix proves Theorem~\ref{thm:feasible-terminal-clt} and Corollary~\ref{cor:student}. We
compare $\widehat J_B$ with two averages that cannot be computed. Replacing $\widehat q_B$ by $q$ in
\eqref{eq:terminal} gives $\widetilde J_B=T_B^{-1}\sum_{i\le T_B}\psi_i(\widehat\theta_B)$, and also
replacing $\widehat\theta_B$ by $\theta^\star$ gives
$\widetilde J_B^\star=T_B^{-1}\sum_{i\le T_B}\psi_i(\theta^\star)$. The error then splits as
\[
\widehat J_B-J=(\widehat J_B-\widetilde J_B)+(\widetilde J_B-\widetilde J_B^\star)+(\widetilde J_B^\star-J).
\]
The last term is an average of martingale differences, and the middle term reflects the random
denominator. Theorem~\ref{thm:terminal-equivalence} below shows that the former is asymptotically
normal and the latter is $o_p(T_B^{-1/2})$. The first term is the regression remainder
$\mathcal R_B(\widehat\theta_B)$, which Assumption~\ref{ass:agg-orth} controls. Beyond
Assumptions~\ref{ass:known} and \ref{ass:pooled}, Theorem~\ref{thm:terminal-equivalence} uses only
predictable logger selection and convergence of the final composition, so it also applies to
other purchasing rules with these properties.

Throughout, for each budget $B$, $(\mathcal F_i)$ is the filtration of Section~\ref{sec:setup} for
the sequence with that budget, extended by $\mathcal F_i=\mathcal F_{T_B}$ for $i>T_B$. Since each
purchase costs at least $c_{\min}$ and the total cost is at most $B$, we have $T_B\le n_B$, where
$n_B=\lceil B/c_{\min}\rceil$. The event $\{i\le T_B\}$ equals $\{B-C_{i-1}\ge c_{\min}\}$, so it is
$\mathcal F_{i-1}$-measurable.

\subsection{Proof of Theorem~\ref{thm:feasible-terminal-clt}}\label{app:pf-clt}

The key step is the following result, which we prove in Appendix~\ref{app:pf-equivalence}.

\begin{thm}\label{thm:terminal-equivalence}
Under Assumptions~\ref{ass:known} and \ref{ass:pooled}, suppose $\widehat\theta_B\to\theta^\star$
in probability. Then
\begin{enumerate}[label=(\roman*),topsep=0pt,itemsep=0pt]
\item $\sqrt{T_B}(\widetilde J_B^{\star}-J)\to N\{0,\Sigma_A(\theta^\star)\}$ in distribution;
\item writing $\widetilde J_B=T_B^{-1}\sum_{i\le T_B}\psi_i(\widehat\theta_B)$,
\begin{equation}\label{eq:equivalence}
\sqrt{T_B}\left(\widetilde J_B-\widetilde J_B^{\star}\right)\to0
\end{equation}
in probability, and consequently $\sqrt{T_B}(\widetilde J_B-J)\to N\{0,\Sigma_A(\theta^\star)\}$ in
distribution.
\end{enumerate}
\end{thm}

Under the conditions of part (i) of Theorem~\ref{thm:design-optimality},
$\widehat\theta_B\to\theta^\star$ almost surely. Theorem~\ref{thm:terminal-equivalence} therefore
applies and gives $\sqrt{T_B}(\widetilde J_B-J)\to N\{0,\Sigma_A(\theta^\star)\}$, and it remains to
show that the estimated regression does not change this limit. By definition, the $j$th coordinate
of $\widehat J_B-\widetilde J_B$ is $\mathcal R_{B,j}(\widehat\theta_B)$. On the event
$\{\|\widehat\theta_B-\theta^\star\|_2\le r_B\}$, whose probability tends to one by
Assumption~\ref{ass:agg-orth},
\[
\max_j|\mathcal R_{B,j}(\widehat\theta_B)|
\le\max_j\sup_{\|\theta-\theta^\star\|_2\le r_B}|\mathcal R_{B,j}(\theta)|
=o_p(T_B^{-1/2}).
\]
Hence $\sqrt{T_B}(\widehat J_B-\widetilde J_B)\to0$ in probability, and Slutsky's theorem gives
\eqref{eq:main-clt}.

\subsection{Proof of Theorem~\ref{thm:terminal-equivalence}}\label{app:pf-equivalence}

We need two lemmas. The first describes the scores at a fixed composition.

\begin{lemma}\label{lem:fixed-theta-martingale}
Let $\theta\in\mathcal S_K$ be deterministic with $\kappa(\theta)>0$. Under Assumption~\ref{ass:known}, $\{\psi_i(\theta)-J\}_{i\le T_B}$ is a bounded martingale difference
sequence with respect to $(\mathcal F_i)$, and
\begin{equation}\label{eq:cond-cov}
\mathbb E\left[\{\psi_i(\theta)-J\}\{\psi_i(\theta)-J\}^\top\mid\mathcal F_{i-1}\right]
=\mathrm{Cov}\{q_\Pi(X)\}
+\mathbb E_X\sum_a\mu_{K_i}(a\mid X)
\frac{\sigma^2(X,a)\pi(a\mid X)\pi(a\mid X)^\top}{m_\theta(a\mid X)^2}.
\end{equation}
\end{lemma}

To prove Lemma~\ref{lem:fixed-theta-martingale}, write $\varepsilon_i=R_i-q(X_i,A_i)$ and
$w_i=\pi(A_i\mid X_i)/m_\theta(A_i\mid X_i)$, with $w_i=0$ when $\pi(A_i\mid X_i)=0$, so that
\[
\psi_i(\theta)-J=\{q_\Pi(X_i)-J\}+w_i\varepsilon_i .
\]
Given $\mathcal F_{i-1}$, the logger $K_i$ is fixed, $X_i\sim P_X$,
$A_i\mid X_i\sim\mu_{K_i}(\cdot\mid X_i)$, and $R_i\mid X_i,A_i\sim P_R(\cdot\mid X_i,A_i)$. Thus
\[
\mathbb E(\varepsilon_i\mid\mathcal F_{i-1},X_i,A_i)=0,
\qquad
\mathbb E(\varepsilon_i^2\mid\mathcal F_{i-1},X_i,A_i)=\sigma^2(X_i,A_i),
\]
and, because $\theta$ is deterministic, $w_i$ depends on $(X_i,A_i)$ alone. Conditioning first on
$(\mathcal F_{i-1},X_i,A_i)$ therefore gives $\mathbb E(w_i\varepsilon_i\mid\mathcal F_{i-1})=0$,
while $\mathbb E\{q_\Pi(X_i)\mid\mathcal F_{i-1}\}=\mathbb E_X\{q_\Pi(X)\}=J$; so the scores have
conditional mean $J$. They are also bounded: since $m_\theta\ge\kappa(\theta)$ whenever
$\pi(A_i\mid X_i)\ne0$,
\[
\|q_\Pi(X_i)-J\|_2\le2\sqrt MR_{\max},
\qquad
|\varepsilon_i|\le2R_{\max},
\qquad
\|w_i\|_2\le\frac{\sqrt M}{\kappa(\theta)} .
\]

For the conditional covariance, the same conditioning shows that the cross term between
$q_\Pi(X_i)-J$ and $w_i\varepsilon_i$ has conditional mean zero. The first term contributes
$\mathrm{Cov}\{q_\Pi(X)\}$, and the second contributes
\[
\mathbb E(w_iw_i^\top\varepsilon_i^2\mid\mathcal F_{i-1})
=\mathbb E\{w_iw_i^\top\sigma^2(X_i,A_i)\mid\mathcal F_{i-1}\}
=\mathbb E_X\sum_a\mu_{K_i}(a\mid X)\frac{\sigma^2(X,a)\pi(a\mid X)\pi(a\mid X)^\top}{m_\theta(a\mid X)^2},
\]
where the last step averages over $A_i\sim\mu_{K_i}(\cdot\mid X_i)$ and then over $X_i\sim P_X$. This
gives \eqref{eq:cond-cov}.

The second lemma relates the number of purchases to the budget.

\begin{lemma}\label{lem:terminal-budget-ratio}
If $\widehat\theta_B\to\theta^\star$ in probability, then $T_B/B\to1/c^\top\theta^\star$ in
probability.
\end{lemma}

Indeed, at termination,
\[
0\le B-C_{T_B}<c_{\min},
\qquad
\frac{C_{T_B}}{T_B}=c^\top\widehat\theta_B\to c^\top\theta^\star\ \text{in probability},
\]
so that $C_{T_B}/B\to1$, and the ratio of the two is $T_B/B$.

We now prove part (i) of Theorem~\ref{thm:terminal-equivalence}. Since $\widetilde J_B^\star$ averages a
data-dependent number $T_B$ of terms, we work with a sum over the fixed range $i\le n_B$, normalized
by $B^{1/2}$, and convert to the normalization by $T_B$ at the end. For $i\le n_B$ let
\[
\xi_{B,i}=\mathbf 1\{i\le T_B\}\{\psi_i(\theta^\star)-J\},
\]
so that $\sum_{i\le n_B}\xi_{B,i}=T_B(\widetilde J_B^\star-J)$. Because $\{i\le T_B\}$ is
$\mathcal F_{i-1}$-measurable, Lemma~\ref{lem:fixed-theta-martingale} shows that
$(\xi_{B,i})_{i\le n_B}$ is a bounded martingale difference array. By the Cram\'er--Wold device, it
suffices to show that, for each fixed $u\in\mathbb R^M$,
\[
S_B(u)=\sum_{i=1}^{n_B}Z_{B,i},\qquad Z_{B,i}=B^{-1/2}u^\top\xi_{B,i},
\]
converges in distribution to $N(0,s_u^2)$, where $s_u^2=u^\top\Sigma_A(\theta^\star)u/c^\top\theta^\star$.

For convenience, we recall the martingale central limit theorem in the form we use
\citep[Corollary~3.1]{hallheyde1980}. For each $n$, let $(X_{n,i},\mathcal H_{n,i})_{i\le k_n}$ be a
square-integrable martingale difference sequence, and let $\eta^2>0$ be a constant. If, for every
$\epsilon>0$,
\[
\sum_{i=1}^{k_n}\mathbb E(X_{n,i}^2\mid\mathcal H_{n,i-1})\to\eta^2
\quad\text{and}\quad
\sum_{i=1}^{k_n}\mathbb E\big(X_{n,i}^2\mathbf 1\{|X_{n,i}|>\epsilon\}\mid\mathcal H_{n,i-1}\big)\to0
\]
in probability, then $\sum_{i\le k_n}X_{n,i}\to N(0,\eta^2)$ in distribution. Because $\eta^2$ is
constant, no nesting condition on the $\sigma$-fields $\mathcal H_{n,i}$ is required.

Suppose first that $s_u^2>0$, and take $X_{B,i}=Z_{B,i}$, $k_B=n_B$, and $\eta^2=s_u^2$. The scores
are bounded, so $|Z_{B,i}|\le c_Z\|u\|_2B^{-1/2}$ for a finite $c_Z$, and the indicators in the second
condition vanish once $B$ is large. For the first condition, \eqref{eq:cond-cov} at
$\theta=\theta^\star$, together with the fact that the average of $\mu_{K_i}$ over $i\le T_B$ is
$m_{\widehat\theta_B}$, gives
\begin{align*}
V_B&=\sum_{i=1}^{n_B}\mathbb E(Z_{B,i}^2\mid\mathcal F_{i-1})=\frac{T_B}{B}\,u^\top\Gamma(\widehat\theta_B)u,\\
\Gamma(\theta)&=\mathrm{Cov}\{q_\Pi(X)\}
+\mathbb E_X\sum_am_\theta(a\mid X)
\frac{\sigma^2(X,a)\pi(a\mid X)\pi(a\mid X)^\top}{m_{\theta^\star}(a\mid X)^2}
\qquad(\theta\in\mathcal S_K).
\end{align*}
Here $\Gamma(\theta^\star)=\Sigma_A(\theta^\star)$ by \eqref{eq:SigmaA}. Moreover,
$m_{\theta^\star}\ge\kappa_\star$ on contributing actions, $\sigma^2\le R_{\max}^2$, and
$|\pi_j\pi_\ell|\le1$, while
\[
\sum_a|m_\theta(a\mid x)-m_{\theta'}(a\mid x)|\le\|\theta-\theta'\|_1
\]
because each $\mu_k(\cdot\mid x)$ sums to one. Every entry of $\Gamma$ therefore satisfies
\[
|\Gamma(\theta)_{j\ell}-\Gamma(\theta')_{j\ell}|\le R_{\max}^2\kappa_\star^{-2}\|\theta-\theta'\|_1 .
\]
Since $\widehat\theta_B\to\theta^\star$ in probability, $u^\top\Gamma(\widehat\theta_B)u\to
u^\top\Sigma_A(\theta^\star)u$, and Lemma~\ref{lem:terminal-budget-ratio} gives $V_B\to s_u^2$, both in
probability. The martingale central limit theorem now gives $S_B(u)\to N(0,s_u^2)$.

If $s_u^2=0$, then $V_B\to0$ in probability. Moreover, $T_B\le B/c_{\min}$, and every entry of
$\Gamma(\theta)$ is at most $R_{\max}^2(1+\kappa_\star^{-2})$ in absolute value. Hence
\[
V_B\le c_{\min}^{-1}R_{\max}^2(1+\kappa_\star^{-2})\|u\|_1^2 ,
\]
and bounded convergence gives $\mathbb E(V_B)\to0$. The cross products of the martingale differences $Z_{B,i}$ have mean zero, so
\[
\mathbb E\{S_B(u)^2\}=\sum_{i=1}^{n_B}\mathbb E(Z_{B,i}^2)=\mathbb E(V_B)\to0 ,
\]
and Chebyshev's inequality gives $S_B(u)\to0=N(0,0)$ in probability.

Combining over $u$, $B^{-1/2}T_B(\widetilde J_B^\star-J)\to N\{0,\Sigma_A(\theta^\star)/c^\top\theta^\star\}$
in distribution. Multiplying by $\sqrt{B/T_B}$, which converges to $\sqrt{c^\top\theta^\star}$ by
Lemma~\ref{lem:terminal-budget-ratio}, and applying Slutsky's theorem gives part (i).

For part (ii), the score differences $\psi_i(\theta)-\psi_i(\theta^\star)$ must be small uniformly over
$\theta$ near $\theta^\star$, because $\widehat\theta_B$ is random. We first record that the scores are
Lipschitz in $\theta$; only the residual term of \eqref{eq:score-vector} depends on the composition.
Let $\mathcal B^\star=\{\theta\in\mathcal S_K:\|\theta-\theta^\star\|_1\le\kappa_\star/2\}$. For
$\theta,\theta'\in\mathcal B^\star$, Assumption~\ref{ass:pooled} and
$|m_\theta-m_{\theta^\star}|\le\|\theta-\theta^\star\|_1$ give $m_\theta\wedge m_{\theta'}\ge\kappa_\star/2$
on contributing actions. Since also $|m_\theta-m_{\theta'}|\le\|\theta-\theta'\|_1$,
\begin{equation}\label{eq:lip-m}
\left|\frac{1}{m_\theta(a\mid x)}-\frac{1}{m_{\theta'}(a\mid x)}\right|
\le\frac{4|m_\theta(a\mid x)-m_{\theta'}(a\mid x)|}{\kappa_\star^{2}}
\le\frac{4\sqrt K}{\kappa_\star^{2}}\|\theta-\theta'\|_2 .
\end{equation}
The uniform bound is stated for the stopped process
\begin{equation}\label{eq:stopped-modulus-process}
\overline\Xi_B(\theta)
=
\frac1{\sqrt B}
\sum_{i=1}^{n_B}
\mathbf 1\{i\le T_B\}
\{\psi_i(\theta)-\psi_i(\theta^\star)\},
\end{equation}
and is proved in Appendix~\ref{app:technical}.

\begin{lemma}\label{lem:modulus}
Under Assumptions~\ref{ass:known} and \ref{ass:pooled}, there is a finite constant $c_6$, depending only on $K$, $M$, $R_{\max}$, $\kappa_\star$, and $c_{\min}$, such that for every $0<r\le \kappa_\star/(2K)$ and
$B\ge1$,
\begin{equation}\label{eq:modulus}
\mathbb E
\sup_{\|\theta-\theta^\star\|_2\le r}
\|\overline\Xi_B(\theta)\|_2
\le
c_6r .
\end{equation}
\end{lemma}

Both averages in \eqref{eq:equivalence} use the same observations, and so
\[
\sqrt{T_B}\big(\widetilde J_B-\widetilde J_B^\star\big)
=\frac1{\sqrt{T_B}}\sum_{i=1}^{T_B}\{\psi_i(\widehat\theta_B)-\psi_i(\theta^\star)\}
=\sqrt{\frac{B}{T_B}}\;\overline\Xi_B(\widehat\theta_B).
\]
Since $C_{T_B}\le c_{\max}T_B$ and $C_{T_B}>B-c_{\max}$, for $B\ge2c_{\max}$ we have
\[
T_B>\frac B{c_{\max}}-1\ge\frac B{2c_{\max}},
\qquad\text{and hence}\qquad
\sqrt{B/T_B}\le L=\sqrt{2c_{\max}} .
\]
Fix $\varepsilon>0$ and $\zeta>0$,
choose $0<r\le\kappa_\star/(2K)$ with $Lc_6r/\zeta<\varepsilon/2$, and let
$E_r=\{\|\widehat\theta_B-\theta^\star\|_2\le r\}$. On $E_r$, $\|\overline\Xi_B(\widehat\theta_B)\|_2$ is
at most the supremum in \eqref{eq:modulus}. Splitting according to $E_r$, Markov's inequality and
Lemma~\ref{lem:modulus} give
\[
\mathbb P\Big\{\sqrt{T_B}\big\|\widetilde J_B-\widetilde J_B^\star\big\|_2>\zeta\Big\}
\le\mathbb P\big(\|\widehat\theta_B-\theta^\star\|_2>r\big)+\frac{L}{\zeta}c_6r
\le\mathbb P\big(\|\widehat\theta_B-\theta^\star\|_2>r\big)+\frac\varepsilon2 .
\]
The first term tends to zero because $\widehat\theta_B\to\theta^\star$ in probability, which proves
\eqref{eq:equivalence}. The final claim follows from part (i) and Slutsky's theorem.

\subsection{Proof of Corollary~\ref{cor:student}}\label{app:pf-student}

Recall that the summand of \eqref{eq:terminal} is
\[
\widehat\psi_i=\widehat q_{\Pi,B}(X_i)
+\frac{\pi(A_i\mid X_i)\{R_i-\widehat q_B(X_i,A_i)\}}{m_{\widehat\theta_B}(A_i\mid X_i)},
\]
where $\widehat q_{\Pi,B}(x)=\{\widehat q_{B,1}(x),\ldots,\widehat q_{B,M}(x)\}^\top$. Let
$\mathcal E_B=\{\|\widehat\theta_B-\theta^\star\|_1\le\kappa_\star/2\}$, whose probability tends to one.
Restricting to such events does not affect convergence in probability, so we work on $\mathcal E_B$,
where $m_{\widehat\theta_B}\ge\kappa_\star/2$ on contributing actions. Since $\widehat q_B$ is truncated
to $[-R_{\max},R_{\max}]$, $|R_i|\le R_{\max}$, and $\pi_j\le1$,
\[
\|\widehat\psi_i\|_2\le\|\widehat q_{\Pi,B}(X_i)\|_2
+\Big\|\frac{\pi(A_i\mid X_i)}{m_{\widehat\theta_B}(A_i\mid X_i)}\Big\|_2|R_i-\widehat q_B(X_i,A_i)|
\le\sqrt MR_{\max}+\frac{4\sqrt MR_{\max}}{\kappa_\star}=K_\psi ,
\]
and the same bound holds for $\psi_i(\theta^\star)$.

We first show that $\widehat\Sigma_B\to\Sigma_A(\theta^\star)$ in probability. We compare
$\widehat\Sigma_B$ with the empirical second moment of the scores at the limiting composition,
\[
\Sigma_B^{\star}=\frac1{T_B}\sum_{i\le T_B}\{\psi_i(\theta^\star)-J\}\{\psi_i(\theta^\star)-J\}^\top,
\]
and write
\begin{equation}\label{eq:app-student-split}
\widehat\Sigma_B-\Sigma_A(\theta^\star)
=
\underbrace{\left(\widehat\Sigma_B-\Sigma_B^{\star}\right)}_{\text{(I)}}
+
\underbrace{\left\{\Sigma_B^{\star}-\Sigma_A(\theta^\star)\right\}}_{\text{(II)}} .
\end{equation}

Term (II) is the average of a martingale part and a predictable part. Let $D_i$ be the outer product
$\{\psi_i(\theta^\star)-J\}\{\psi_i(\theta^\star)-J\}^\top$ minus its conditional expectation given
$\mathcal F_{i-1}$. Because $\|\psi_i(\theta^\star)-J\|_2\le2K_\psi$, the matrices
$\mathbf 1\{i\le T_B\}D_i$, $i\le n_B$, form a martingale difference array bounded entrywise by
$8K_\psi^2$, hence their sum has second moment $O(n_B)=O(B)$. Dividing by $T_B\ge B/(2c_{\max})$ shows
that $T_B^{-1}\sum_{i\le T_B}D_i\to0$ in probability. The average of the conditional expectations is
$\Gamma(\widehat\theta_B)$ from the proof of part (i) of Theorem~\ref{thm:terminal-equivalence}, which
converges to $\Sigma_A(\theta^\star)$ in probability. Hence (II) tends to zero.

For term (I), the identity $uu^\top-vv^\top=u(u-v)^\top+(u-v)v^\top$ gives
\[
\|uu^\top-vv^\top\|\le(\|u\|+\|v\|)\|u-v\| .
\]
Both $\widehat\psi_i-\widehat J_B$ and
$\psi_i(\theta^\star)-J$ have norm at most $2K_\psi$, and so
\begin{equation}\label{eq:app-I-bound}
\left\|\widehat\Sigma_B-\Sigma_B^{\star}\right\|
\le
4K_\psi\left\{
\frac1{T_B}\sum_{i=1}^{T_B}\|\widehat\psi_i-\psi_i(\theta^\star)\|
+
\|\widehat J_B-J\|
\right\}.
\end{equation}
The second term in braces is $o_p(1)$ by Theorem~\ref{thm:feasible-terminal-clt}. For the first,
we split the score difference into a part from the denominator and a part from the regression,
\begin{equation}\label{eq:app-score-diff}
\widehat\psi_i-\psi_i(\theta^\star)
=
\underbrace{
\pi(A_i\mid X_i)\left\{
\frac{1}{m_{\widehat\theta_B}(A_i\mid X_i)}-\frac{1}{m_{\theta^\star}(A_i\mid X_i)}
\right\}
\{R_i-\widehat q_B(X_i,A_i)\}
}_{(\mathrm a)}
+
\underbrace{\vphantom{\frac11}\zeta_i}_{(\mathrm b)},
\end{equation}
where
\[
\zeta_i=\{\widehat q_{\Pi,B}-q_\Pi\}(X_i)
-\frac{\pi(A_i\mid X_i)\{\widehat q_B-q\}(X_i,A_i)}{m_{\theta^\star}(A_i\mid X_i)} .
\]
For $(\mathrm a)$, \eqref{eq:lip-m} gives
$|1/m_{\widehat\theta_B}-1/m_{\theta^\star}|\le4\sqrt K\kappa_\star^{-2}\|\widehat\theta_B-\theta^\star\|_2$ on
contributing actions. With $|R_i-\widehat q_B(X_i,A_i)|\le2R_{\max}$ and $\|\pi(A_i\mid X_i)\|_2\le\sqrt M$,
every term of $(\mathrm a)$ has norm at most
\[
8\sqrt{MK}R_{\max}\kappa_\star^{-2}\|\widehat\theta_B-\theta^\star\|_2 ,
\]
and so its average over $i\le T_B$ tends to zero in probability. For $(\mathrm b)$, since $\pi_j\le1$ and
$m_{\theta^\star}\ge\kappa_\star$ on contributing actions, Jensen's inequality gives
\[
\zeta_{i,j}^2\le2(1+\kappa_\star^{-2})\sum_a\{\widehat q_B-q\}^2(X_i,a).
\]
The mean-square consistency condition then yields $\max_jT_B^{-1}\sum_{i\le T_B}\zeta_{i,j}^2=o_p(1)$,
and the Cauchy--Schwarz inequality gives
\[
\frac1{T_B}\sum_{i=1}^{T_B}\|\zeta_i\|
\le\sqrt M\max_j\Big\{\frac1{T_B}\sum_{i=1}^{T_B}\zeta_{i,j}^2\Big\}^{1/2}=o_p(1).
\]
The first term in \eqref{eq:app-I-bound} is therefore $o_p(1)$, so (I) tends to zero, and
$\widehat\Sigma_B\to\Sigma_A(\theta^\star)$ in probability.

For the intervals, fix $j$ with $A_j(\theta^\star)>0$. By Theorem~\ref{thm:feasible-terminal-clt},
$\sqrt{T_B}(\widehat J_{B,j}-J_j)\to N\{0,A_j(\theta^\star)\}$, and $(\widehat\Sigma_B)_{jj}\to
A_j(\theta^\star)$ in probability, since $A_j(\theta^\star)$ is the $j$th diagonal entry of
$\Sigma_A(\theta^\star)$. Slutsky's theorem gives
\[
\frac{\sqrt{T_B}(\widehat J_{B,j}-J_j)}{(\widehat\Sigma_B)_{jj}^{1/2}}\to N(0,1),
\]
which is the coverage statement for \eqref{eq:wald}. Finally, $A_j(\theta^\star)>0$ holds whenever
$V_j>0$ or $\mathbb E_X\sum_{a\in\mathcal A}W_j(X,a)>0$.

\subsection{Pooled Efficiency}\label{app:aggregation}

The logger-specific estimator divides each observation by the propensity of the logger that
generated it. At a fixed logger composition $\theta$, its covariance is
\begin{equation}\label{eq:SigmaL}
\Sigma_L(\theta)
=
\mathrm{Cov}\{q_\Pi(X)\}
+
\mathbb E_X\sum_{a\in\mathcal A}
\sigma^2(X,a)\pi(a\mid X)\pi(a\mid X)^\top
\sum_{k=1}^K\frac{\theta_k}{\mu_k(a\mid X)} ,
\end{equation}
whose $j$th diagonal entry is
\begin{equation}\label{eq:Lj}
L_j(\theta)
=
V_j+
\mathbb E_X\sum_{a\in\mathcal A}
W_j(X,a)
\sum_{k=1}^K\frac{\theta_k}{\mu_k(a\mid X)} .
\end{equation}

To derive \eqref{eq:aggregate-local-order}, fix $u\in\mathbb R^M$. The context terms of
$\Sigma_A(\theta)$ and $\Sigma_L(\theta)$ coincide, so it suffices to compare the residual terms. For
each $(x,a)$, Jensen's inequality for the convex function $y\mapsto1/y$, with $1/0=\infty$, gives
\[
\frac{1}{m_\theta(a\mid x)}=\frac{1}{\sum_k\theta_k\mu_k(a\mid x)}
\le\sum_{k=1}^K\frac{\theta_k}{\mu_k(a\mid x)} .
\]
Multiplying by $\sigma^2(x,a)\{u^\top\pi(a\mid x)\}^2\ge0$, summing over $a$, and integrating over
$P_X$ gives $u^\top\Sigma_A(\theta)u\le u^\top\Sigma_L(\theta)u$.

Taking $u=e_j$ gives $A_j(\theta)\le L_j(\theta)$. For a fixed contrast $u^\top J$, equality holds
if and only if the logger propensities with $\theta_k>0$ agree on the context-action regions where
$\sigma^2(x,a)\{u^\top\pi(a\mid x)\}^2>0$, which is the equality condition of Jensen's inequality.
In particular, $A_j(\theta^\star)\le L_j(\theta^\star)$ for each target.

\section{A Block-Split Estimator}\label{app:blocksplit}

In the final estimator \eqref{eq:terminal}, $\widehat q_B$ is fitted on the same observations over
which it is averaged, and the denominator $m_{\widehat\theta_B}$ is random. The block-split estimator
divides the sample into blocks of geometrically growing size, and evaluates each block with a
regression fitted on earlier blocks and with its own realized logger frequencies in the
denominator.

Fix $\bar n\ge1$ and $\varkappa>1$ with $(\varkappa-1)\bar n\ge1$, let $\bar\tau_0=0$ and
$\bar\tau_\ell=\lceil\varkappa^{\ell-1}\bar n\rceil$ for $\ell\ge1$, and let
\begin{equation}\label{eq:block-partition}
L_B=\max\{\ell\ge1:\bar\tau_\ell\le T_B\},
\qquad
\tau_0=0,\qquad
\tau_\ell=\bar\tau_\ell\ (1\le\ell<L_B),
\qquad
\tau_{L_B}=T_B .
\end{equation}
The condition $(\varkappa-1)\bar n\ge1$ gives $\varkappa^{\ell}\bar n-\varkappa^{\ell-1}\bar n\ge1$, so
the $\bar\tau_\ell$ strictly increase and every block is nonempty. Set
$\mathcal B_\ell=\{i:\tau_{\ell-1}<i\le\tau_\ell\}$ and $n_\ell=|\mathcal B_\ell|$. For $\ell\ge2$, let
$\widehat q^{(\ell)}$ be fitted using only observations before block $\ell$, and let
$\widehat\theta^{(\ell)}_k=n_\ell^{-1}\sum_{i\in\mathcal B_\ell}\mathbf 1\{K_i=k\}$ be the logger
proportions within block $\ell$. The block-split estimator is
\begin{equation}\label{eq:blocksplit}
\widehat J^{\mathrm{BS}}_{B,j}
=
\frac1{T_B-\tau_1}\sum_{\ell=2}^{L_B}\sum_{i\in\mathcal B_\ell}
\left[
\widehat q^{(\ell)}_j(X_i)
+
\frac{\pi_j(A_i\mid X_i)}
{m_{\widehat\theta^{(\ell)}}(A_i\mid X_i)}
\{R_i-\widehat q^{(\ell)}(X_i,A_i)\}
\right].
\end{equation}

The regression for block $\ell$ is fixed before the block begins, and the denominator uses the
block's own logger frequencies. The construction is designed so that the regression error enters
only through a martingale term.

\section{Proofs of Technical Results}\label{app:technical}

The proofs below use the notation of Appendices~\ref{app:convergence} and \ref{app:inference}. We
present them in the order of logical dependence rather than the order in which the results are
stated.

\subsection{Proof of Proposition~\ref{prop:strong-convexity}}

It suffices to show that the Hessian of $\mathcal G_j$ dominates $2c_{\min}^3Q_j$ along feasible directions,
since Assumption~\ref{ass:identifiability} bounds the quadratic form of $Q_j$ below by
$\lambda_j\|h\|_2^2$. On each $\mathcal C(e)$, both denominators in \eqref{eq:Fj} are bounded away
from zero, by $c_{\max}^{-1}$ and $e$, and the integrands are bounded; thus the Hessian exists and is
continuous there, and differentiating \eqref{eq:Fj} twice gives
\begin{equation*}
h^\top\nabla^2\mathcal G_j(v)h
=\frac{2V_j(\mathbf 1^\top h)^2}{(\mathbf 1^\top v)^3}
+2\,\mathbb E_X\sum_{a\in\mathcal A}
\frac{W_j(X,a)\{\boldsymbol\mu(X,a)^\top h\}^2}{\{\boldsymbol\mu(X,a)^\top v\}^3}.
\end{equation*}
The numerators are the two terms of $h^\top Q_jh$. Meanwhile, $1=c^\top v\ge c_{\min}\mathbf 1^\top v$
and $\mu_k\le1$ give
\[
\boldsymbol\mu(x,a)^\top v\le\mathbf 1^\top v\le c_{\min}^{-1},
\]
and so each reciprocal cube is at least $c_{\min}^3$. By Assumption~\ref{ass:identifiability},
\[
h^\top\nabla^2\mathcal G_j(v)h\ge2c_{\min}^3h^\top Q_jh\ge2c_{\min}^3\lambda_j\|h\|_2^2,
\]
which is \eqref{eq:hessian-lower}.

To pass to strong convexity, take $y,y'\in\mathcal C$ such that the segment between them lies in
$\mathcal C_+$, and put $h=y'-y$, so that $c^\top h=0$. If $y\in\mathcal C(e)$ and
$y'\in\mathcal C(e')$, linearity of the coverage puts the whole segment in
$\mathcal C(\min\{e,e'\})$; hence $s\mapsto \mathcal G_j(y+sh)$ is twice differentiable on $[0,1]$, with second
derivative at least $m_j\|h\|_2^2$, where $m_j=2c_{\min}^3\lambda_j$. With $m=\min_jm_j$, each
$\mathcal G_j-\tfrac m2\|\cdot\|_2^2$ is therefore convex along the segment, and so is their maximum
$\mathcal G-\tfrac m2\|\cdot\|_2^2$.

\subsection{Proof of Lemma~\ref{lem:quadratic-growth}}

If $\mathcal G(v)=\infty$ there is nothing to prove; otherwise, write $h=v-z^\star$ and
$\phi(s)=\mathcal G(z^\star+sh)$ for $s\in[0,1]$. By Assumption~\ref{ass:pooled},
$z^\star\in\mathcal C(\kappa_\star/c_{\max})$, and because
$\boldsymbol\mu^\top(z^\star+sh)\ge(1-s)\boldsymbol\mu^\top z^\star$, the segment from $z^\star$ to
$z^\star+s'h$ lies in $\mathcal C\{(1-s')\kappa_\star/c_{\max}\}$ for each $s'<1$.
Proposition~\ref{prop:strong-convexity} therefore gives, for $0<s<s'<1$,
\[
\phi(s)\le\Big(1-\frac s{s'}\Big)\phi(0)+\frac s{s'}\phi(s')
-\frac m2\,\frac s{s'}\Big(1-\frac s{s'}\Big)s'^2\|h\|_2^2 .
\]
Since $\mathcal G$ is convex on $\mathcal C$, $\phi(s')\le(1-s')\phi(0)+s'\phi(1)$, and letting $s'\uparrow1$
yields
\[
\phi(s)\le(1-s)\phi(0)+s\phi(1)-\tfrac m2s(1-s)\|h\|_2^2 .
\]
The left side is at least $\phi(0)=\mathcal G(z^\star)$; rearranging and dividing by $s$ gives
\[
\mathcal G(v)-\mathcal G(z^\star)\ge\tfrac m2(1-s)\|h\|_2^2 ,
\]
and letting $s\downarrow0$ proves the claim.

\subsection{Proof of Lemma~\ref{lem:smoothed}}

Parts (a) to (c) are standard. Part (a) follows from
\[
\exp\{\mathcal G(v)/\rho\}\le\sum_j\exp\{\mathcal G_j(v)/\rho\}\le M\exp\{\mathcal G(v)/\rho\},
\]
part (b) is the chain rule, and (c) holds because the
log-sum-exp function is convex and increasing in each argument, while each $\mathcal G_j$ is convex. For (d),
differentiating (b) once more gives
\[
\nabla^2\Phi_\rho=\sum_j\omega_j\nabla^2\mathcal G_j
+\frac1\rho\Big\{\sum_j\omega_j\nabla \mathcal G_j\nabla \mathcal G_j^\top
-\Big(\sum_j\omega_j\nabla \mathcal G_j\Big)\Big(\sum_j\omega_j\nabla \mathcal G_j\Big)^{\!\top}\Big\},
\]
a weighted average of the Hessians of the $\mathcal G_j$ plus $\rho^{-1}$ times the covariance matrix of the
target gradients under the weights $\omega_j$. For $v\in\mathcal C(e)$, the bounds
$\|\mathbf 1\mathbf 1^\top\|_{\mathrm{op}}=K$,
$\|\boldsymbol\mu\boldsymbol\mu^\top\|_{\mathrm{op}}\le K$, $V_j\le R_{\max}^2$, and
$\mathbb E_X\sum_aW_j(X,a)\le R_{\max}^2$ give
\[
\|\nabla^2\mathcal G_j(v)\|_{\mathrm{op}}\le2KR_{\max}^2(c_{\max}^3+e^{-3}),
\qquad
|\partial_k\mathcal G_j(v)|\le R_{\max}^2(c_{\max}^2+e^{-2}).
\]
The covariance term therefore has operator norm at most
$\max_j\|\nabla \mathcal G_j(v)\|_2^2\le KR_{\max}^4(c_{\max}^2+e^{-2})^2$, and since $\mathcal C(e)$ is convex,
these bounds give the Lipschitz constant. Finally, for (e), a direct calculation gives
\[
\partial_\rho\Phi_\rho=\log\sum_j\exp(\mathcal G_j/\rho)-\sum_j\omega_j\mathcal G_j/\rho=-\sum_j\omega_j\log\omega_j,
\]
the entropy of the weights, which is nonnegative.

\subsection{Proof of Lemma~\ref{lem:coverage}}

The argument rests on two facts. First, one purchase can lower the pooled coverage by at most a
factor $t/(t+1)$, whichever logger is bought: since $\mu_{K_{t+1}}\ge0$,
\begin{equation}
\label{eq:coverage-one-step}
m_{\theta_{t+1}}=\frac{t\,m_{\theta_t}+\mu_{K_{t+1}}}{t+1}\ge\frac{t}{t+1}\,m_{\theta_t}.
\end{equation}
Second, a forced purchase raises a smallest count, and at most $K$ loggers share it; along
consecutive forced purchases, the smallest count therefore grows by at least one every $K$
purchases.

For (a), the initialization gives $\theta_{n_0,k}=1/K$, and so \eqref{eq:share-coverage} yields
\[
\kappa(\theta_{n_0})\ge\frac{\kappa_{\mathrm{all}}}K\ge8\chi_0>\chi_{n_0}.
\]
Unforced steps have
$\kappa(\theta_t)\ge\chi_t$ by definition, so it only remains to consider a maximal block of
forced steps $t_0,t_0+1,\ldots$. Here $t_0>n_0\ge4K$, and the unforced step $t_0-1$ has
$\kappa(\theta_{t_0-1})\ge\chi_{t_0}$. Let $n=\lceil2\chi_{t_0}t_0/\kappa_{\mathrm{all}}\rceil$ and
$L=Kn$. If the block lasted $L$ steps, the smallest count would reach $n$, and
\eqref{eq:share-coverage} would give
\[
\kappa(\theta_{t_0+L})\ge\frac{\kappa_{\mathrm{all}}\,n}{t_0+Kn}\ge\chi_{t_0}\ge\chi_{t_0+L},
\]
where the middle inequality is equivalent to $n(\kappa_{\mathrm{all}}-K\chi_{t_0})\ge\chi_{t_0}t_0$
and so holds because $K\chi_{t_0}\le\kappa_{\mathrm{all}}/8$ and $n\ge2\chi_{t_0}t_0/\kappa_{\mathrm{all}}$.
The block therefore lasts at most
\[
L\le\frac{2K\chi_{t_0}t_0}{\kappa_{\mathrm{all}}}+K\le\frac{t_0}4+\frac{t_0}4=\frac{t_0}2
\]
steps, and repeated use of \eqref{eq:coverage-one-step} gives, within the block,
\[
\kappa(\theta_t)\ge\kappa(\theta_{t_0-1})\,\frac{t_0-1}{t_0+L}
\ge\chi_{t_0}\,\frac{t_0-1}{3t_0/2}\ge\frac{\chi_{t_0}}2\ge\frac{\chi_t}{2},
\]
using $t_0\ge4$ and that $\chi_t$ is nonincreasing.

For (b), at a forced step $s$, \eqref{eq:share-coverage} gives
\[
\frac{\kappa_{\mathrm{all}}\min_kN_{s,k}}s\le\kappa(\theta_s)<\chi_s ,
\]
and so the logger bought, which has the smallest
count, has fewer than $\chi_ss/\kappa_{\mathrm{all}}$ observations. Because
$\chi_ss=\chi_0s^{1-\upsilon}$ is nondecreasing, a logger is forced at steps $s\le t$ only while its
count is below $\chi_tt/\kappa_{\mathrm{all}}$, hence at most $\chi_tt/\kappa_{\mathrm{all}}+1$ times.
Summing over loggers, the $p$th forced step $s_p$ satisfies
$p\le K(\chi_0s_p^{1-\upsilon}/\kappa_{\mathrm{all}}+1)$, and so
\[
s_p\ge\Big\{\frac{p\,\kappa_{\mathrm{all}}}{2K\chi_0}\Big\}^{1/(1-\upsilon)}
\qquad\text{for }p\ge2K .
\]
Since $\gamma_s\le1$ and $\gamma_s\le c_{\max}/C_{s+1}\le c_{\max}/(c_{\min}s)$,
\[
\sum_{t\in\mathcal T}\gamma_t\le2K+\frac{c_{\max}}{c_{\min}}\sum_{p\ge2K}
\Big\{\frac{2K\chi_0}{p\,\kappa_{\mathrm{all}}}\Big\}^{1/(1-\upsilon)}<\infty,
\]
where the series converges because $1/(1-\upsilon)>1$.

\subsection{Proof of Lemma~\ref{lem:switch}}

Because $0\le\mu_k\le1$, we have $|m_\theta(a\mid x)-m_{\theta^\star}(a\mid x)|\le\|\theta-\theta^\star\|_1$
for all $(x,a)$, and so
\[
\kappa(\theta_t)\ge\kappa(\theta^\star)-\|\theta_t-\theta^\star\|_1\ge\frac{\kappa_\star}2
\]
once $\|\theta_t-\theta^\star\|_1\le\kappa_\star/2$. Since $\chi_t\to0$, eventually
$\chi_t<\kappa_\star/2\le\kappa(\theta_t)$, after which no step is forced.

\subsection{Proof of Lemma~\ref{lem:recursion}}

We first show that $h_t$ is bounded. Choose $t_1$ beyond which both recursions hold and $q_t\le1$
for $t\notin\mathcal T$, and let $g_t=\max\{h_t,b,1\}$. For $t\ge t_1$ outside $\mathcal T$, $h_{t+1}$
is at most a weighted average of $h_t$ and $q_t$, and so
\[
g_{t+1}\le g_t\quad(t\notin\mathcal T),
\qquad
g_{t+1}\le(1+2\gamma_t)g_t\quad(t\in\mathcal T).
\]
Thus, for $t\ge t_1$,
\[
h_t\le g_t\le g_{t_1}\exp\Big(2\sum_{s\in\mathcal T}\gamma_s\Big)=H<\infty .
\]
Given this bound, every step satisfies the first recursion up to a summable error. Writing
$(1+\gamma_t)h_t=(1-\gamma_t)h_t+2\gamma_th_t$ for $t\in\mathcal T$, we have, for all $t\ge t_1$,
\[
h_{t+1}\le(1-\gamma_t)h_t+\gamma_t\bar q_t+a_t,
\qquad
(\bar q_t,a_t)=
\begin{cases}
(q_t,\,0), & t\notin\mathcal T,\\
(0,\,\gamma_t(2H+b)), & t\in\mathcal T,
\end{cases}
\]
so that $\bar q_t\to0$ and $\sum_ta_t<\infty$. Fix $\epsilon>0$, and choose $t_2\ge t_1$ such that
$\bar q_t\le\epsilon$ for $t\ge t_2$ and $\sum_{t\ge t_2}a_t\le\epsilon$. Unrolling the recursion
from $t$ down to $t_2$, with $P_i=\prod_{s=i}^{t}(1-\gamma_s)$ and $P_{t+1}=1$,
\[
h_{t+1}\le P_{t_2}h_{t_2}+\sum_{i=t_2}^{t}\gamma_iP_{i+1}\bar q_i+\sum_{i=t_2}^{t}P_{i+1}a_i .
\]
The weights $\gamma_iP_{i+1}=P_{i+1}-P_i$ sum to $1-P_{t_2}\le1$, and $P_{i+1}\le1$; thus
$h_{t+1}\le P_{t_2}h_{t_2}+2\epsilon$. Finally,
$P_{t_2}\le\exp(-\sum_{s=t_2}^{t}\gamma_s)\to0$ because $\sum_s\gamma_s$ diverges, hence
$\limsup_th_t\le2\epsilon$ for every $\epsilon>0$, that is, $h_t\to0$.

\subsection{Proof of Lemma~\ref{lem:softmax}}

Fix $y$, write $p=\omega(y)$, and let $\mathcal J(y)$ denote the Jacobian of $y\mapsto\omega(y)$.
Differentiating $\log\omega_i(y)=y_i/\rho-\log\sum_\ell\exp(y_\ell/\rho)$ gives
$\partial\log\omega_i/\partial y_j=(\mathbf 1\{i=j\}-p_j)/\rho$, and so
\[
\mathcal J(y)=\frac1\rho\big\{\mathrm{diag}(p)-pp^\top\big\}.
\]
For any $u\in\mathbb R^M$, the $i$th entry of $\{\mathrm{diag}(p)-pp^\top\}u$ is $p_i(u_i-p^\top u)$, and
$|u_i-p^\top u|\le2\|u\|_\infty$ because $p^\top u$ is a weighted average of the $u_i$. Since the $p_i$
sum to one,
\[
\|\mathcal J(y)u\|_1=\frac1\rho\sum_{i=1}^Mp_i\,|u_i-p^\top u|\le\frac2\rho\|u\|_\infty ,
\]
with a constant that does not depend on $y$. The map $s\mapsto\omega\{y'+s(y-y')\}$ is continuously
differentiable on $[0,1]$, and integrating along the segment from $y'$ to $y$ gives
\[
\|\omega(y)-\omega(y')\|_1
=\Big\|\int_0^1\mathcal J\{y'+s(y-y')\}(y-y')\,ds\Big\|_1
\le\frac2\rho\|y-y'\|_\infty .
\]

\subsection{Proof of Lemma~\ref{lem:budget-expansion}}

Without the regularization constants $\eta$ and $\eta_0$, $G_j(N)$ would equal $\mathcal G_j(N)$, which is
$\mathcal G_j(z)/C$ by homogeneity of degree $-1$; we show that the shifts by $\eta$ and $\eta_0$ change the
rescaled quantities by $O(C^{-1})$. Write $N=Cz$, $s=\mathbf 1^\top z$, and
$u=\boldsymbol\mu(X,a)^\top z$, so that $T=Cs$ and $D_N(X,a)=\eta+Cu$. Throughout,
$s\ge c_{\max}^{-1}$ because $c^\top z=1$; only actions in $\mathcal S$ contribute, and on them
$u\ge e$ because $z\in\mathcal C(e)$; moreover, $C\ge c_{\min}$ because $N\ne0$, and
$V_j\le R_{\max}^2$ and $\bar W_j=\mathbb E_X\sum_aW_j(X,a)\le R_{\max}^2$ because $|R|\le R_{\max}$.
We use two elementary inequalities, valid for $s>0$ and $0\le a\le b$:
\begin{equation}\label{eq:recip-one}
\left|\frac1{s+a}-\frac1s\right|=\frac{a}{s(s+a)}\le\frac{a}{s^2},
\end{equation}
\begin{equation}\label{eq:recip-two}
\left|\frac1{(s+a)(s+b)}-\frac1{s^2}\right|
=\frac{s(a+b)+ab}{s^2(s+a)(s+b)}\le\frac{2sb+b^2}{s^4}.
\end{equation}

For the first bound, dividing numerator and denominator by $C$ gives
\[
CG_j(N)-\mathcal G_j(z)
=V_j\Big(\frac1{s+\eta_0/C}-\frac1s\Big)
+\mathbb E_X\sum_aW_j(X,a)\Big(\frac1{u+\eta/C}-\frac1u\Big).
\]
Applying \eqref{eq:recip-one} with $a=\eta_0/C$ to the first term, and with $s$ replaced by $u$ and
$a=\eta/C$ to the second, the absolute value is at most
\[
\frac{V_j\eta_0}{Cs^2}+\mathbb E_X\sum_aW_j(X,a)\frac{\eta}{Cu^2}
\le\frac{R_{\max}^2(\eta_0c_{\max}^2+\eta e^{-2})}{C}=\frac{K_G(e)}{C}.
\]

For the second bound, differentiating \eqref{eq:Fj} and rescaling \eqref{eq:gain} give, with the
arguments $(X,a)$ suppressed,
\[
-\partial_k\mathcal G_j(z)=\frac{V_j}{s^2}+\mathbb E_X\sum_a\frac{W_j\mu_k}{u^2},
\]
\[
C^2\Delta_{j,k}(N)
=\frac{V_j}{(s+\eta_0/C)\{s+(1+\eta_0)/C\}}
+\mathbb E_X\sum_a\frac{W_j\mu_k}{(u+\eta/C)\{u+(\eta+\mu_k)/C\}} .
\]
Applying \eqref{eq:recip-two} with $a=\eta_0/C$ and $b=(1+\eta_0)/C$ to the first terms, and using
$b^2\le(1+\eta_0)^2/(c_{\min}C)$,
\[
\left|\frac{V_j}{(s+\eta_0/C)\{s+(1+\eta_0)/C\}}-\frac{V_j}{s^2}\right|
\le V_j\Big\{\frac{2(1+\eta_0)}{s^3}+\frac{(1+\eta_0)^2}{c_{\min}s^4}\Big\}\frac1C
\le\frac{K_1}{C},
\]
with $K_1=R_{\max}^2\{2(1+\eta_0)c_{\max}^3+(1+\eta_0)^2c_{\max}^4/c_{\min}\}$. Likewise, applying
\eqref{eq:recip-two} with $s$ replaced by $u$, $a=\eta/C$, and $b=(\eta+\mu_k)/C\le(1+\eta)/C$ to the
second terms, and using $\mu_k\le1$,
\begin{align*}
\left|\mathbb E_X\sum_aW_j\mu_k\Big[\frac1{(u+\eta/C)\{u+(\eta+\mu_k)/C\}}-\frac1{u^2}\Big]\right|
&\le\mathbb E_X\sum_aW_j\Big\{\frac{2(1+\eta)}{u^3}+\frac{(1+\eta)^2}{c_{\min}u^4}\Big\}\frac1C\\
&\le\frac{K_2(e)}{C},
\end{align*}
with $K_2(e)=R_{\max}^2\{2(1+\eta)e^{-3}+(1+\eta)^2e^{-4}/c_{\min}\}$. Adding the two gives the
second bound with $K_\Delta(e)=K_1+K_2(e)$.

\subsection{Proof of Lemma~\ref{lem:oracle-LMO}}

By Lemma~\ref{lem:smoothed}(b) and the definition of the index,
\[
g_k=\frac1{c_k}\sum_j\omega_j(z_t;\rho_t)\,\partial_k\mathcal G_j(z_t),
\qquad
C_t^2S_{t,k}=\frac1{c_k}\sum_j\omega_{j,t}\,C_t^2\Delta_{j,k}(N_t).
\]
The two have the same form, with the gains in place of the derivatives and with weights computed
from $C_tG_j(N_t)$ instead of $\mathcal G_j(z_t)$. Thus
\[
\big|C_t^2S_{t,k}+g_k\big|
\le\frac1{c_{\min}}\sum_j\omega_{j,t}\big|C_t^2\Delta_{j,k}(N_t)+\partial_k\mathcal G_j(z_t)\big|
+\frac1{c_{\min}}\big\|\omega_t-\omega(z_t;\rho_t)\big\|_1\max_j|\partial_k\mathcal G_j(z_t)| .
\]
Since the weights sum to one, Lemma~\ref{lem:budget-expansion} bounds the first term by
$K_\Delta(e)/(c_{\min}C_t)$. For the second, the arguments of the exponential weights differ by at most $K_G(e)/C_t$, again by
Lemma~\ref{lem:budget-expansion}, so Lemma~\ref{lem:softmax} gives
\[
\|\omega_t-\omega(z_t;\rho_t)\|_1\le\frac{2K_G(e)}{\rho_tC_t}=\frac{2K_G(e)}{\rho_0}\,C_t^{\alpha-1};
\]
together with $|\partial_k\mathcal G_j(z_t)|\le R_{\max}^2(c_{\max}^2+e^{-2})$, the second term is at most
\[
\frac{2R_{\max}^2(c_{\max}^2+e^{-2})K_G(e)}{c_{\min}\rho_0}\,C_t^{\alpha-1}.
\]
Finally, because
$e\le1/c_{\max}$, both $K_\Delta(e)$ and $(c_{\max}^2+e^{-2})K_G(e)$ are at most constant multiples of
$e^{-4}$, which gives \eqref{eq:LMO-error}.

\subsection{Proof of Proposition~\ref{prop:nuisance}}

Adding and subtracting $c_k^{-1}\sum_j\widehat\omega_{j,t}\Delta_{j,k}(N_t)$ separates the effect of the
gains from that of the weights:
\[
\widehat S_{t,k}-S_{t,k}
=
\frac1{c_k}\sum_{j=1}^M\widehat\omega_{j,t}
\{\widehat\Delta_{j,k,t}-\Delta_{j,k}(N_t)\}
+
\frac1{c_k}\sum_{j=1}^M
(\widehat\omega_{j,t}-\omega_{j,t})\Delta_{j,k}(N_t).
\]
The first term averages the gain errors with weights that sum to one, so it is at most
$r_t/(c_{\min}C_t^2)$ by Assumption~\ref{ass:nuisance}. In the second, the arguments of the exponential weights
$C_t\widehat G_{j,t}$ and $C_tG_j(N_t)$ differ by at most $r_t$, again by
Assumption~\ref{ass:nuisance}, and Lemma~\ref{lem:softmax} gives
$\|\widehat\omega_t-\omega_t\|_1\le2r_t/\rho_t$. Meanwhile, \eqref{eq:gain}, the bound
$D_{N_t}\ge t\,\kappa(\theta_t)$ on contributing actions, and $C_t\le c_{\max}t$ give
\[
C_t^2\Delta_{j,k}(N_t)\le2c_{\max}^2R_{\max}^2\kappa(\theta_t)^{-2}.
\]
Multiplying by $C_t^2$ and adding the two bounds yields \eqref{eq:annealed-score-error}; note also
that $r_t/\rho_t=\rho_0^{-1}C_t^{\alpha}r_t$.

\subsection{Proof of Lemma~\ref{lem:direction}}

Let $\widehat k$ be the purchased logger, which maximizes $\widehat S_{t,k}$ over all loggers. A linear
function on $\mathcal C$ attains its minimum at a vertex, so the minimum in \eqref{eq:approx-LMO}
equals $\min_kg_k$, and we need to show that $g_{\widehat k}\le\min_kg_k+\delta_t$. Let
$\varepsilon_t=\max_k|C_t^2\widehat S_{t,k}+g_k|$. Inserting $S_{t,k}$ and applying
Lemma~\ref{lem:oracle-LMO} with $e=\kappa(\theta_t)/c_{\max}$,
\[
\varepsilon_t
\le C_t^2\max_k|\widehat S_{t,k}-S_{t,k}|+\max_k|C_t^2S_{t,k}+g_k|
\le C_t^2\max_k|\widehat S_{t,k}-S_{t,k}|+D_1c_{\max}^4\kappa(\theta_t)^{-4}\big(C_t^{\alpha-1}+C_t^{-1}\big).
\]
By definition of $\varepsilon_t$, $g_\ell-\varepsilon_t\le-C_t^2\widehat S_{t,\ell}\le g_\ell+\varepsilon_t$
for every logger $\ell$. Since $\widehat k$ maximizes the estimated index, it follows that, for any
logger $k$,
\[
g_{\widehat k}\le-C_t^2\widehat S_{t,\widehat k}+\varepsilon_t
\le-C_t^2\widehat S_{t,k}+\varepsilon_t
\le g_k+2\varepsilon_t .
\]
The purchase is therefore a $\delta_t$-approximate conditional-gradient step with
$\delta_t=2\varepsilon_t$, and the bound on $\varepsilon_t$ gives the first claim. Under the
additional conditions, $C_t\le c_{\max}t$ gives $r_t\le Rc_{\max}^{\beta_r}C_t^{-\beta_r}$, and
Proposition~\ref{prop:nuisance} bounds the estimation term by
$D_0(r_t+\rho_0^{-1}\underline m^{-2}C_t^{\alpha}r_t)\le D_0Rc_{\max}^{\beta_r}(1+\rho_0^{-1})\underline m^{-2}C_t^{\alpha-\beta_r}$
once $C_t\ge1$; since also $C_t^{-1}\le C_t^{\alpha-1}$, the last claim follows.

\subsection{Proof of Lemma~\ref{lem:infinite-path-coupling}}

Before $\tau_B$, at least $c_{\max}$ of the budget remains, so every logger is affordable in both sequences.
The sequences then share counts, estimates, indices, and tie-breaking, so they make the same choices;
by induction, they agree through $\tau_B$. Afterwards, less than $c_{\max}$ remains and each purchase
costs at least $c_{\min}$, so the budget-$B$ sequence makes at most $L_0$ further purchases. At the common
time $T_B$ both count vectors total $T_B$, hence each sequence makes $T_B-\tau_B\le L_0$ purchases after
$\tau_B$. Write $a$ and $b$ for the counts that the budget-$B$ and the infinite-budget sequences add, so that
$\|a\|_1=\|b\|_1\le L_0$ and both costs $c^\top a$ and $c^\top b$ lie in $[0,L_0c_{\max}]$. Then
\[
\|N_{T_B}-\bar N_{T_B}\|_1=\|a-b\|_1\le2L_0,
\qquad
|C_{T_B}-\bar C_{T_B}|=|c^\top a-c^\top b|\le L_0c_{\max}.
\]

For the normalized designs, note that the budget-$B$ sequence stops only when no logger is affordable, so
that $C_{T_B}>B-c_{\max}\ge B/2$ for $B\ge2c_{\max}$. Adding and subtracting
$\bar N_{T_B}/C_{T_B}$,
\[
\frac{N_{T_B}}{C_{T_B}}-\frac{\bar N_{T_B}}{\bar C_{T_B}}
=\frac{N_{T_B}-\bar N_{T_B}}{C_{T_B}}
+\frac{\bar N_{T_B}}{\bar C_{T_B}}\cdot\frac{\bar C_{T_B}-C_{T_B}}{C_{T_B}} .
\]
The first term has $\ell_1$ norm at most $2L_0/C_{T_B}\le4L_0/B$. In the second,
$\|\bar N_{T_B}\|_1/\bar C_{T_B}\le c_{\min}^{-1}$, because each purchase costs at least $c_{\min}$, and
$|\bar C_{T_B}-C_{T_B}|/C_{T_B}\le2L_0c_{\max}/B$, so its norm is at most
$2L_0c_{\max}/(c_{\min}B)$; since the $\ell_2$ norm is at most the $\ell_1$ norm, the first claim holds
with $K_z=2L_0(2+c_{\max}/c_{\min})$. As for the sample fractions,
$\widehat\theta_B-\bar N_{T_B}/T_B=(N_{T_B}-\bar N_{T_B})/T_B$ has $\ell_1$ norm at most $2L_0/T_B$,
and $T_B\ge C_{T_B}/c_{\max}\ge B/(2c_{\max})$ because $C_{T_B}\le c_{\max}T_B$; this gives the second claim with
$K_\theta=4L_0c_{\max}$.

\subsection{Proof of Lemma~\ref{lem:plugin}}

By \eqref{eq:Ghat} and \eqref{eq:Gj},
\[
\widehat G_{j,t}-G_j(N_t)
=\frac{\widehat V_{j,t}-V_j}{t+\eta_0}
+(\widehat P_t-P_X)\sum_a\frac{\widehat W_{j,t}(\cdot,a)}{D_{N_t}(\cdot,a)}
+P_X\sum_a\frac{\widehat W_{j,t}(\cdot,a)-W_j(\cdot,a)}{D_{N_t}(\cdot,a)},
\]
and \eqref{eq:gainhat} and \eqref{eq:gain} give the same decomposition for
$\widehat\Delta_{j,k,t}-\Delta_{j,k}(N_t)$, with $1/D_{N_t}$ replaced by
$\mu_k/\{D_{N_t}(D_{N_t}+\mu_k)\}$ and $1/(t+\eta_0)$ by $1/\{(t+\eta_0)(t+1+\eta_0)\}$. On
contributing actions, $D_{N_t}\ge t\,\kappa(\theta_t)\ge t\underline m_t$, and so
\[
\frac t{D_{N_t}}\le\underline m_t^{-1},
\qquad
\frac{t^2\mu_k}{D_{N_t}(D_{N_t}+\mu_k)}\le\underline m_t^{-2},
\]
while the factors multiplying $\widehat V_{j,t}-V_j$, after rescaling by $t$ and $t^2$, are at most one.
Hence
\[
t\max_j|\widehat G_{j,t}-G_j(N_t)|\le r_{V,t}+r_{P,t}+\underline m_t^{-1}r_{W,t},
\qquad
t^2\max_{j,k}|\widehat\Delta_{j,k,t}-\Delta_{j,k}(N_t)|\le r_{V,t}+r_{P,t}+\underline m_t^{-2}r_{W,t}.
\]
Since $C_t\le c_{\max}t$ and $\underline m_t\le1$, the two bounds of Assumption~\ref{ass:nuisance}
hold with $K'a_t$ in place of $r_t$, where $K'=\max\{c_{\max},c_{\max}^2\}$. Finally,
$a_t\le\sup_{s\ge t}a_s$, and the right side is nonincreasing and tends to zero when $a_t\to0$; hence
Assumption~\ref{ass:nuisance} holds with $r_t=K'\sup_{s\ge t}a_s$.

\subsection{Proof of Corollary~\ref{cor:finite-contexts}}

We verify the conditions of Lemma~\ref{lem:plugin}, with rates that depend on a lower bound
$\underline m_t$ for the pooled coverage, chosen below. Every bound rests on one inequality: if
$(\xi_i)$ is a martingale difference sequence with $|\xi_i|\le b$, the Azuma--Hoeffding inequality
gives $\mathbb P(|\sum_{i\le t}\xi_i|>b\sqrt{6t\log t})\le2t^{-3}$, which is summable in $t$, thus,
by the Borel--Cantelli lemma,
\begin{equation}\label{eq:azuma-bc}
\Big|\sum_{i\le t}\xi_i\Big|\le b\sqrt{6t\log t}
\qquad\text{almost surely for all sufficiently large }t .
\end{equation}

Fix a target-relevant pair $(x,a)$, meaning $P_X(x)>0$ and $(x,a)\in\mathcal S$, and let
$\nu_i=\mathbf 1\{X_i=x,A_i=a\}$ and $n_t=\sum_{i\le t}\nu_i$. Given $\mathcal F_{i-1}$, the logger
$K_i$ is fixed, $X_i\sim P_X$, and $A_i\sim\mu_{K_i}(\cdot\mid X_i)$; thus
$\mathbb E(\nu_i\mid\mathcal F_{i-1})=P_X(x)\mu_{K_i}(a\mid x)$ and
\[
\sum_{i\le t}\mathbb E(\nu_i\mid\mathcal F_{i-1})=P_X(x)\,t\,m_{\theta_t}(a\mid x)\ge P_X(x)\,t\,\underline m_t .
\]
Here $\underline m_t=\chi_t/2$ is a lower bound for $\kappa(\theta_t)$ by Lemma~\ref{lem:coverage}(a);
if $\theta_t\to\theta^\star$, we may also take $\underline m_t=\kappa_\star/2$ for all large $t$, by
Lemma~\ref{lem:switch}. Writing $p_t=P_X(x)\underline m_t$, \eqref{eq:azuma-bc} applied to
$\nu_i-\mathbb E(\nu_i\mid\mathcal F_{i-1})$ gives, for all large $t$,
\[
n_t\ge p_tt-\sqrt{6t\log t}\ge\frac{p_tt}2 ,
\]
because $p_t\sqrt{t/\log t}\to\infty$ when $\upsilon<1/2$.

Next, with $\varepsilon_i=R_i-q(x,a)$, the variables $\nu_i\varepsilon_i$ and
$\nu_i\{\varepsilon_i^2-\sigma^2(x,a)\}$ are martingale differences bounded by $2R_{\max}$ and
$4R_{\max}^2$, since $\mathbb E(\varepsilon_i\mid\mathcal F_{i-1},X_i,A_i)=0$ and
$\mathbb E(\varepsilon_i^2\mid\mathcal F_{i-1},X_i,A_i)=\sigma^2(X_i,A_i)$ on $\{\nu_i=1\}$. By
\eqref{eq:azuma-bc} and the bound on $n_t$,
\[
|\widehat q_t(x,a)-q(x,a)|=\frac{|\sum_{i\le t}\nu_i\varepsilon_i|}{n_t}
\le\frac{4R_{\max}}{p_t}\sqrt{\frac{6\log t}{t}} .
\]
For the variance, write
\[
\widehat\sigma_t^2(x,a)=\frac{n_t}{n_t-1}\Big\{\frac1{n_t}\sum_{i\le t}\nu_i\varepsilon_i^2-(\widehat q_t-q)^2(x,a)\Big\}.
\]
By \eqref{eq:azuma-bc}, the average differs from $\sigma^2(x,a)$ by $O(p_t^{-1}\sqrt{\log t/t})$.
The squared term is $O(p_t^{-2}\log t/t)$ and $n_t/(n_t-1)=1+O\{(p_tt)^{-1}\}$, both of smaller order
because $p_t\sqrt{t/\log t}\to\infty$; hence
$|\widehat\sigma_t^2-\sigma^2|(x,a)=O(p_t^{-1}\sqrt{\log t/t})$. Truncation to $[0,R_{\max}^2]$
cannot increase this error, since $\sigma^2(x,a)\in[0,R_{\max}^2]$, and because $\mathcal X$ and
$\mathcal A$ are finite, these bounds hold simultaneously over all target-relevant pairs. Applying
\eqref{eq:azuma-bc} also to $\mathbf 1\{X_i=x\}-P_X(x)$, a bounded martingale difference since $X_i$ is
independent of $\mathcal F_{i-1}$, gives $\max_x|\widehat P_t(\{x\})-P_X(x)|\le\sqrt{6\log t/t}$, and so
\[
\big|(\widehat P_t-P_X)f\big|\le|\mathcal X|\sup|f|\sqrt{6\log t/t}
\qquad\text{for every bounded }f .
\]

We can now check the three conditions; only target-relevant pairs enter $W_j$ and $q_j$, because
$\pi_j(a\mid x)=0$ elsewhere. First,
\[
\sum_a\|\widehat W_{j,t}(\cdot,a)-W_j(\cdot,a)\|_{L_1(P_X)}
=\sum_{x,a}P_X(x)\pi_j(a\mid x)^2|\widehat\sigma_t^2-\sigma^2|(x,a)
=O\big(\underline m_t^{-1}\sqrt{\log t/t}\big).
\]
Second, $|\widehat q_{j,t}(x)-q_j(x)|\le\max_a|\widehat q_t-q|(x,a)$ over target-relevant $a$, and
$\widehat q_{j,t}$ and $q_j$ are bounded by $R_{\max}$, so that
\[
|\widehat V_{j,t}-V_j|
\le|\widehat P_t(\widehat q_{j,t}^2-q_j^2)|+|(\widehat P_t-P_X)q_j^2|
+2R_{\max}\big\{|\widehat P_t(\widehat q_{j,t}-q_j)|+|(\widehat P_t-P_X)q_j|\big\}
=O\big(\underline m_t^{-1}\sqrt{\log t/t}\big).
\]
Third, $t/D_{N_t}\le\underline m_t^{-1}$ on target-relevant actions, and $0\le\widehat W_{j,t}\le R_{\max}^2$
vanishes elsewhere. The two functions in the last condition of Lemma~\ref{lem:plugin} are therefore
bounded by $|\mathcal A|R_{\max}^2\underline m_t^{-1}$ and $|\mathcal A|R_{\max}^2\underline m_t^{-2}$,
and the bound on $\widehat P_t-P_X$ gives $r_{P,t}=O(\underline m_t^{-2}\sqrt{\log t/t})$.
Lemma~\ref{lem:plugin} now gives Assumption~\ref{ass:nuisance} with
$r_t=O(\underline m_t^{-3}\sqrt{\log t/t})$, which is $O(t^{3\upsilon}\sqrt{\log t/t})$ when
$\underline m_t=\chi_t/2$ and $O(\sqrt{\log t/t})$ when $\underline m_t=\kappa_\star/2$.

\subsection{Proof of Lemma~\ref{lem:modulus}}

Because $m_\theta$ is linear in $\theta$ and the score depends on $\theta$ only through $1/m_\theta$,
we expand $1/m_\theta$ around $\theta^\star$. Write $\delta=\theta^\star-\theta$ and
$\varepsilon_i=R_i-q(X_i,A_i)$, and suppose $\|\delta\|_2\le r$, so that $|\delta_k|\le r$ for every $k$.
On contributing actions, $m_{\theta^\star}\ge\kappa_\star$ and
\[
|m_{\theta^\star}-m_\theta|=\Big|\sum_k\delta_k\mu_k\Big|\le Kr\le\frac{\kappa_\star}2 ,
\]
so the geometric series converges and
\[
\frac1{m_\theta}-\frac1{m_{\theta^\star}}
=\sum_{p\ge1}\frac{(\sum_k\delta_k\mu_k)^p}{m_{\theta^\star}^{p+1}}
=\sum_{p\ge1}\sum_{|\alpha|=p}\binom p\alpha\frac{\delta^\alpha\mu^\alpha}{m_{\theta^\star}^{p+1}},
\]
where $\alpha$ ranges over multi-indices in $\mathbb Z_+^K$, $\delta^\alpha=\prod_k\delta_k^{\alpha_k}$,
and $\mu^\alpha=\prod_k\mu_k^{\alpha_k}$. Substituting into \eqref{eq:stopped-modulus-process}, and
exchanging the finite sum over $i$ with the absolutely convergent series,
\[
\overline\Xi_{B,j}(\theta)=\sum_{p\ge1}\sum_{|\alpha|=p}\binom p\alpha\delta^\alpha G_{B,j,\alpha},
\qquad
G_{B,j,\alpha}=\frac1{\sqrt B}\sum_{i=1}^{n_B}\mathbf 1\{i\le T_B\}
\frac{\pi_j(A_i\mid X_i)\mu(A_i\mid X_i)^\alpha}{m_{\theta^\star}(A_i\mid X_i)^{p+1}}\,\varepsilon_i .
\]

The variables $G_{B,j,\alpha}$ do not depend on $\theta$, and each is a sum of martingale differences:
$\mathbf 1\{i\le T_B\}$ is $\mathcal F_{i-1}$-measurable, the factors other than $\varepsilon_i$ are
functions of $(X_i,A_i)$, and $\mathbb E(\varepsilon_i\mid\mathcal F_{i-1},X_i,A_i)=0$. Since $\pi_j\le1$,
$\mu^\alpha\le1$, $|\varepsilon_i|\le2R_{\max}$, and $m_{\theta^\star}\ge\kappa_\star$, each summand, without
the factor $B^{-1/2}$, is at most $2R_{\max}\kappa_\star^{-(p+1)}$. Martingale differences are
uncorrelated, so with $n_B\le B/c_{\min}+1$,
\[
\mathbb E|G_{B,j,\alpha}|\le\{\mathbb E(G_{B,j,\alpha}^2)\}^{1/2}
\le\Big(\frac{n_B}B\Big)^{1/2}2R_{\max}\kappa_\star^{-(p+1)}\le c\,\kappa_\star^{-(p+1)},
\qquad c=2R_{\max}(1+c_{\min}^{-1})^{1/2}.
\]
Since $|\delta^\alpha|\le r^p$ and $\sum_{|\alpha|=p}\binom p\alpha=K^p$, it follows that
\[
\mathbb E\sup_{\|\theta-\theta^\star\|_2\le r}|\overline\Xi_{B,j}(\theta)|
\le\sum_{p\ge1}r^p\sum_{|\alpha|=p}\binom p\alpha\mathbb E|G_{B,j,\alpha}|
\le\frac c{\kappa_\star}\sum_{p\ge1}\Big(\frac{Kr}{\kappa_\star}\Big)^p\le\frac{2cK}{\kappa_\star^{2}}\,r ,
\]
where the last step uses $Kr/\kappa_\star\le1/2$. Summing over the $M$ coordinates, with
$\|\cdot\|_2\le\sum_j|\cdot_j|$, gives \eqref{eq:modulus} with $c_6=2cKM\kappa_\star^{-2}$.

\section{Additional Experimental Details and Results}\label{app:experiments}

In this appendix, we give the implementation details, the complete data-generating mechanisms, and
the additional experiments referred to in Section~\ref{sec:experiments}. Unless stated otherwise,
all policy values are estimated by the final estimator \eqref{eq:terminal}.

\subsection{Implementation}\label{app-implementation}

All experiments were implemented in Python 3.11.11 with NumPy 2.4.6, SciPy 1.17.1, pandas 3.0.5,
scikit-learn 1.9.0, matplotlib 3.11.1, and PyYAML 6.0.3. The continuous optimal design is computed
with SciPy's SLSQP solver, with \texttt{maxiter}$=500$ and \texttt{ftol}$=10^{-12}$. The static
planners solve an epigraph form of the continuous relaxation of \eqref{eq:static} with SLSQP, using
\texttt{maxiter}$=250$ and \texttt{ftol}$=10^{-12}$, and retry with \texttt{maxiter}$=1000$ and
\texttt{ftol}$=10^{-9}$ if the first attempt fails. The continuous solution is then rounded to
integer counts. The rounding takes the integer part of each logger's planned count, examines the
neighboring counts that differ by one in each coordinate, and spends the remaining budget on the
loggers with the largest reduction in the worst-target criterion per unit cost.

Adaptive Greedy and Oracle greedy break ties deterministically. Indices within a relative tolerance
of $10^{-12}$ and an absolute tolerance of $10^{-14}$ of the largest index are treated as tied, and
the logger with the smallest index is chosen. The rounding step of the static planners breaks ties
in the same way.

In the synthetic environments, the reward mean is estimated separately for each action by ridge
regression with penalty one on the context and the interaction $x_1x_2$, once the action has at
least six observations. With fewer, we use the action's sample mean, or the pooled mean if the
action has no observations. The prediction before any data is zero.
The reward variance is estimated separately for each action from the squared in-sample
residuals of the same mean fit, each multiplied by the degrees-of-freedom correction
$\min\{3,n_a/\max(n_a-p,1)\}$, where $n_a$ is the number of observations of action $a$ and
$p=7$ is the number of parameters of the mean model. With at least $25$ observations, these
corrected squared residuals are regressed by ridge regression with penalty one on all
monomials of the context of degree at most two. With $6$ to $24$ observations their mean is
used, with $2$ to $5$ the sample variance of the action's rewards, and with fewer the pooled
sample variance of all rewards, or $0.1$ if fewer than two rewards are available. Predicted
variances are truncated to $[0.001,16]$. In the ACTG application, the mean regression uses
the basis of Appendix~\ref{app-actg-details} with penalty $10^{-8}$, and the variance is
constant within each treatment arm, equal to the mean of the corrected squared residuals
with $p=10$, with the same fallbacks for small samples.
During purchasing, each fit uses only the observations collected up to that point, and we do not
use cross-fitting. Theorems~\ref{thm:design-optimality} and \ref{thm:feasible-terminal-clt} rely on
Assumptions~\ref{ass:nuisance} and \ref{ass:agg-orth}, which we do not verify for this ridge
implementation. Instead, the calibration study in Appendix~\ref{app-inference-calibration} examines
its behavior empirically.

Every procedure begins with four observations from each logger, collected in turn, for an
initialization of $n_0=20$ observations. The reward estimates are refitted every ten observations,
counting the initialization, and once more on all data for the final estimator. Reoptimized
plug-in reoptimizes at the same times. Adaptive Greedy uses
$\alpha=0.4$, $\rho_0=0.5$, and $\eta=\eta_0=1$, and the reward prediction before the first fit is
zero. Forced sampling in Algorithm~\ref{alg:acquisition} uses $\chi_0=10^{-4}$ and $\upsilon=0.01$. Each
logger mixes its policy with the uniform distribution, so every logger gives every action
probability at least $0.004$ in Environments A to C, $0.002$ in the weak-overlap experiment of
Appendix~\ref{app-stress}, and $0.0125$ in the ACTG application. Hence
$\kappa(\theta)\ge0.002>\chi_0\ge\chi_t$ for every composition $\theta$. Thus forced sampling never
occurs, and every reported run is a run of Algorithm~\ref{alg:acquisition}. These floors also give
$\kappa_{\mathrm{all}}\ge0.01$, so that
$\chi_0\le\kappa_{\mathrm{all}}/(8K)=\kappa(\mathbf 1/K)/8$ with $K=5$. The initialization of four
observations per logger gives $n_0/K=4$.

For target $\pi_j$, $\operatorname{RMSE}_j=\{R^{-1}\sum_r(\widehat J_{r,j}-J_j)^2\}^{1/2}$, the
worst-target RMSE is $\max_j\operatorname{RMSE}_j$, and the average RMSE is
$M^{-1}\sum_j\operatorname{RMSE}_j$. The worst-target RMSE is the largest of the target-specific
RMSEs, not the square root of the average of the largest squared errors. Its Monte Carlo standard
error is the standard deviation over $5000$ bootstrap resamples of the replications. In each
resample, the target-specific RMSEs are recomputed before the maximum is taken, and the same
resampled replications are used for all procedures, which preserves the common random numbers. For
quantities that are averages over replications, such as $\mathcal G(\widehat z)/\mathcal G(z^\star)$ and
$\|\widehat\theta-\theta^\star\|_2$, the Monte Carlo standard error is the sample standard deviation
divided by $\sqrt R$. The main simulations, the trajectory experiment, and the ACTG application use
the seeds $20260804$, $20260829$, and $20260911$, respectively. Table~\ref{tab-replication-inventory}
lists the number of replications in each experiment.

\begin{table}[t]
\centering
\caption{Number of replications in each experiment.}
\label{tab-replication-inventory}
\begin{tabular}{lc}
\toprule
Experiment & Replications \\
\midrule
Environments A and B & 200 \\
Environment C & 100 \\
Misleading initialization & 100 per initialization \\
Misspecified reward model & 100 \\
Weak overlap & 100 \\
Sensitivity to $\alpha$ and $\rho_0$ & 50 per setting \\
Inference calibration & 500 per environment and budget \\
Block-split estimator & 200 \\
Trajectory experiment & 50 \\
ACTG 175 & 200 \\
\bottomrule
\end{tabular}
\end{table}

\subsection{Comparison Methods}\label{app-methods}

Let $\mathcal P_B(\widetilde V,\widetilde W)$ denote the count plan for budget $B$ produced by the
static planner of Appendix~\ref{app-implementation} when it is given the reward moments
$\widetilde V$ and $\widetilde W$.

\textit{Adaptive Greedy} is Algorithm~\ref{alg:acquisition}. At step $t$, it purchases from the affordable logger
with the largest index
$\widehat S_{t,k}=c_k^{-1}\sum_j\widehat\omega_{j,t}\widehat\Delta_{j,k,t}$, where
$\widehat\Delta_{j,k,t}$ are the estimated one-step gains and $\widehat\omega_{j,t}$ are the
weights of Section~\ref{sec:selection}.

\textit{Reoptimized plug-in} computes, at each update time $t$, the plan
$\widetilde N_B^{(t)}=\mathcal P_B(\widehat V_t,\widehat W_t)$ for the declared budget $B$. Until the
next update, it purchases from the loggers whose current counts fall furthest short of this plan. Each
plan solves a static problem, but the procedure is adaptive because the plan is recomputed as data
accumulate.

\textit{Estimated static} follows the single plan $\mathcal P_B(\widehat V_{n_0},\widehat W_{n_0})$
computed after the initialization and never updates it. \textit{Oracle static} follows
$\mathcal P_B(V,W)$, which uses
the true reward moments, and serves only as a reference. Because both plans come from the rounded
relaxation, neither is guaranteed to be the exact integer optimum.

\textit{Equal expenditure} spends the same amount on each logger, so that in the limit its sample
fractions are $\theta_k^{\mathrm{eq}}=c_k^{-1}/\sum_\ell c_\ell^{-1}$. These differ from equal sample
counts whenever the costs differ. \textit{Cheapest only} purchases from the cheapest affordable logger
after the initialization, which shows the effect of using cost while ignoring the logging
policies.

\textit{Oracle greedy} purchases from the affordable logger that maximizes the true gain per unit cost,
$\Delta_k(N_t)/c_k$. It is not feasible in practice and is reported as a reference in the robustness
experiments of Appendix~\ref{app-stress} and in the ACTG application.

\subsection{Synthetic Environments}\label{app-synthetic-settings}

All three environments have a five-dimensional context, five actions, five loggers, and four target
policies. Environments A and B share the context distribution, the reward model, and the loggers,
and Environment C differs from Environment A only in its costs.

\subsubsection*{Context and reward}

The context follows the Gaussian mixture
\[
X\sim0.5\,N(m_1,I_5)+0.3\,N(m_2,0.7I_5)+0.2\,N(m_3,0.5I_5),
\]
with $m_1=(0,0,0,0,0)^\top$, $m_2=(1.5,-1,0,0,0)^\top$, and $m_3=(-1,1.5,1,0,0)^\top$. Let
\[
B_q=
\begin{pmatrix}
0.25 & -0.10 & 0.05 & 0 & 0\\
0.10 & 0.25 & -0.05 & 0.08 & 0\\
-0.10 & 0.15 & 0.20 & 0 & 0.05\\
-0.15 & -0.05 & 0.25 & 0.10 & 0\\
0.05 & -0.15 & 0.10 & -0.05 & 0.20
\end{pmatrix}.
\]
For action $a$, the reward before clipping is normal with mean
$q_0(x,a)=(B_qx)_a+0.3x_1x_2\mathbf 1\{a\in\{3,4,5\}\}$ and variance $s^2(x,a)=0.05$ for
$a\in\{1,2,3\}$ and $s^2(x,a)=0.05+0.02(1+0.5x_3^2)$ for $a\in\{4,5\}$. The observed reward is
clipped to $[-4,4]$, and the design calculations use the exact mean and variance of the clipped
normal distribution.

\subsubsection*{Loggers}

Loggers 1 to 4 favor the action pairs $(1,2)$, $(2,3)$, $(3,4)$, and $(4,5)$, with logit gaps
$4.5$, $3.5$, $3.5$, and $4.5$ and exploration floors $0.02$, $0.04$, $0.04$, and $0.04$. Their
logits do not depend on the context, so they differ only in which actions they cover. Logger 5 is
close to uniform, with a diagonal context coefficient of $0.01$ and exploration floor $0.10$. A
softmax policy $p(a\mid x)$ with exploration floor $\epsilon$ becomes
$(1-\epsilon)p(a\mid x)+\epsilon/5$. The costs in Environments A and B are $c=(1,2,4,8,16)$.

\subsubsection*{Target policies}

All four target policies share the context coefficient matrix
\[
\Gamma=0.025
\begin{pmatrix}
1 & -1 & 0 & 0 & 0\\
-1 & 1 & 0 & 0 & 0\\
0 & 1 & -1 & 0 & 0\\
0 & 0 & 1 & -1 & 0\\
0 & 0 & 0 & 1 & -1
\end{pmatrix}.
\]
In Environment A, their logit intercepts are $b_1=(0,0,0,0,0)$, $b_2=(-0.4,0.4,0.4,-0.4,-0.4)$,
$b_3=(-0.55,-0.55,-0.55,0.55,0.55)$, and $b_4=(-0.475,-0.475,0.475,0.475,-0.475)$, and their
exploration floors are $(0.12,0.06,0.05,0.05)$. Environment B uses the floors
$(0.09,0.045,0.0375,0.0375)$ and the fourth intercept $b_4=(-0.9,-0.9,0.9,0.9,-0.9)$, which has the
same sign pattern as in Environment A but a larger magnitude. Environment C uses the policies of
Environment A and the costs $c=(1,4,16,8,2)$. Table~\ref{tab-oracle-designs} lists the optimal
designs.

\begin{table}[t]
\centering
\caption{Optimal designs in the synthetic environments.}
\label{tab-oracle-designs}
\begin{tabular}{lcc}
\toprule
Environment & $\theta^\star$ & $\mathcal G^\star$ \\
\midrule
A & $(0.783722,0,0.084511,0.131766,0)$ & $0.742341$ \\
B & $(0.714882,0.073948,0.137919,0.073251,0)$ & $0.809153$ \\
C & $(0.256545,0,0,0,0.743455)$ & $0.437430$ \\
\bottomrule
\end{tabular}
\end{table}

\subsection{Additional Synthetic Results}\label{app-synthetic-results}

\subsubsection*{Average RMSE}\label{app-average-rmse}

\begin{figure}[t]
    \centering
    \includegraphics[width=\textwidth]
    {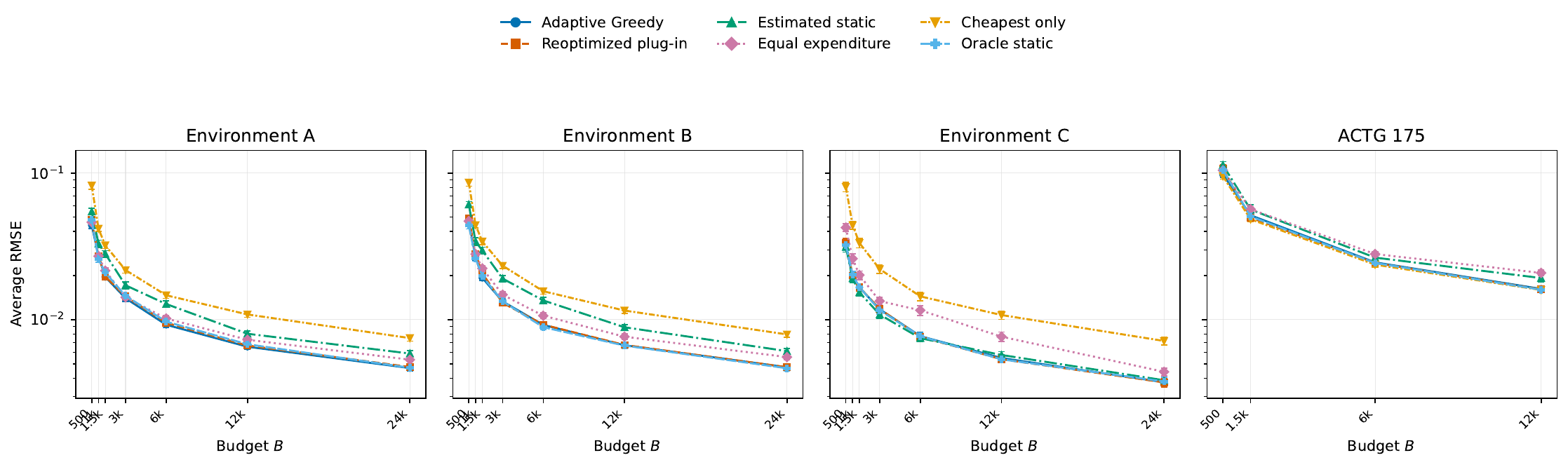}
    \caption{
Average RMSE over targets in the synthetic environments and the ACTG application. Error bars show
$\pm1$ Monte Carlo standard error.
}
    \label{fig-app-average-rmse}
\end{figure}

Figure~\ref{fig-app-average-rmse} reports the average RMSE over the four targets in the three
synthetic environments and the ACTG application. The comparisons are similar to those for the
worst-target RMSE in Section~\ref{sec:experiments}.

\subsubsection*{Paired comparisons}\label{app-paired-comparisons}

Because common random numbers are used, we also compare procedures pairwise, resampling the same
replications for both procedures in each pair. The paired intervals separate Adaptive Greedy from
Cheapest only at every budget. In Environments A and B, they also separate Adaptive Greedy from
Estimated static and Equal expenditure at most budgets. In Environment C, the intervals comparing
Adaptive Greedy with Estimated static contain zero at every budget, in line with their similar
accuracy in Figure~\ref{fig-main-rmse}. Adaptive Greedy and Reoptimized plug-in are much closer, and
only two of the twenty-one comparisons between them exclude zero.

\subsubsection*{Design criterion and estimation error}\label{app-f-rmse}

\begin{figure}[t]
    \centering
    \includegraphics[width=0.75\textwidth]
    {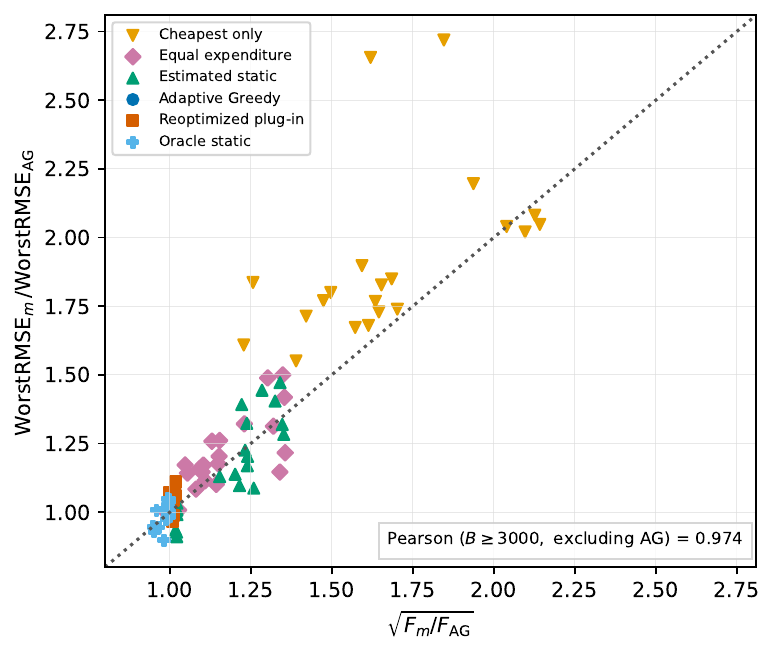}
    \caption{
Relative worst-target RMSE against the square root of the relative criterion $\mathcal G$, with Adaptive
Greedy as the reference. The diagonal line is the identity.
}
    \label{fig-app-f-rmse}
\end{figure}

Figure~\ref{fig-app-f-rmse} plots, for each procedure, environment, and budget, the worst-target
RMSE relative to that of Adaptive Greedy against the square root of the criterion $\mathcal G$ relative to
that of Adaptive Greedy. The square root puts $\mathcal G$, a variance, on the scale of the RMSE. For budgets
of at least $3000$, the correlation between the two is about $0.974$, excluding the reference
points of Adaptive Greedy itself. Thus the design criterion closely tracks estimation accuracy. The
points are not independent, so the figure is descriptive.

\subsection{Robustness Checks and Sensitivity}\label{app-stress}

\begin{figure}[t]
    \centering
    \includegraphics[width=\textwidth]
    {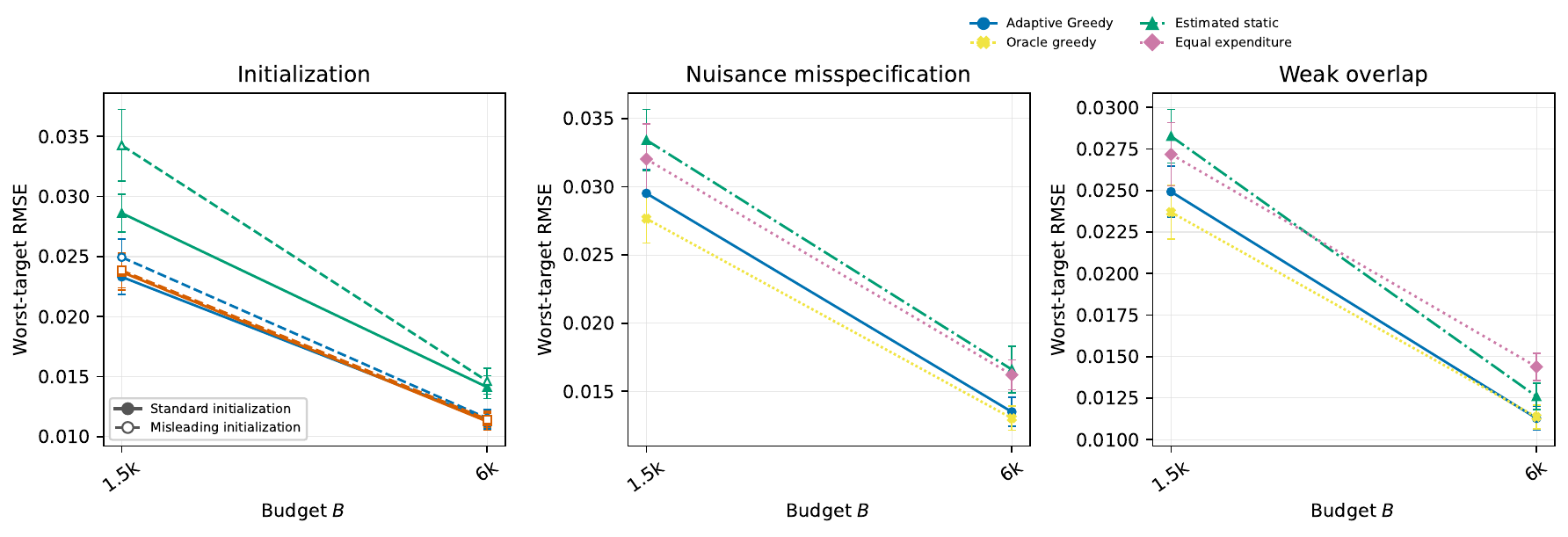}
    \caption{
Robustness experiments with a misleading initialization, a misspecified reward model, and weak overlap.
Error bars show $\pm1$ Monte Carlo standard error. Only the procedures run in each experiment are
shown.
}
    \label{fig-app-stress}
\end{figure}

We examine three departures from the standard setting in Environment A, each with budgets $1500$
and $6000$ and $R=100$ replications. The results are shown in Figure~\ref{fig-app-stress}.

The first experiment uses a misleading initialization, in which the third mixture component of the
context distribution is excluded from the initial observations only. At $B=1500$, the worst-target
RMSE is $0.025$ for Adaptive Greedy, $0.024$ for Reoptimized plug-in, and $0.034$ for Estimated
static, so the plan fixed after the initialization suffers most. Reoptimized plug-in is slightly
more accurate than Adaptive Greedy in this setting. At $B=6000$, the worst-target RMSE of Adaptive
Greedy is $0.0116$ under the misleading initialization and $0.0115$ under the standard
initialization. Its criterion ratio is about $1.011$, so the effect of the misleading start has
disappeared.

The second experiment misspecifies the reward model by adding $0.15x_3^2$ to the mean reward of
actions 3, 4, and 5, while the fitted regression omits this term. At $B=6000$, the worst-target
RMSE is $0.0135$ for Adaptive Greedy, $0.0130$ for Oracle greedy, $0.0166$ for Estimated static,
and $0.0162$ for Equal expenditure. Misspecification affects Adaptive Greedy only slightly, and it
remains more accurate than the static and cost-based procedures.

The third experiment weakens overlap by lowering the logger exploration floors to
$(0.01, \allowbreak 0.01, \allowbreak 0.01, \allowbreak 0.01, \allowbreak 0.025)$. At $B=6000$, the worst-target RMSE is $0.0113$ for Adaptive Greedy,
$0.0114$ for Oracle greedy, $0.0126$ for Estimated static, and $0.0144$ for Equal expenditure.

\subsubsection*{Sensitivity to $\alpha$ and $\rho_0$}\label{app-parameter-sensitivity}

\begin{figure}[t]
    \centering
    \includegraphics[width=\textwidth]
    {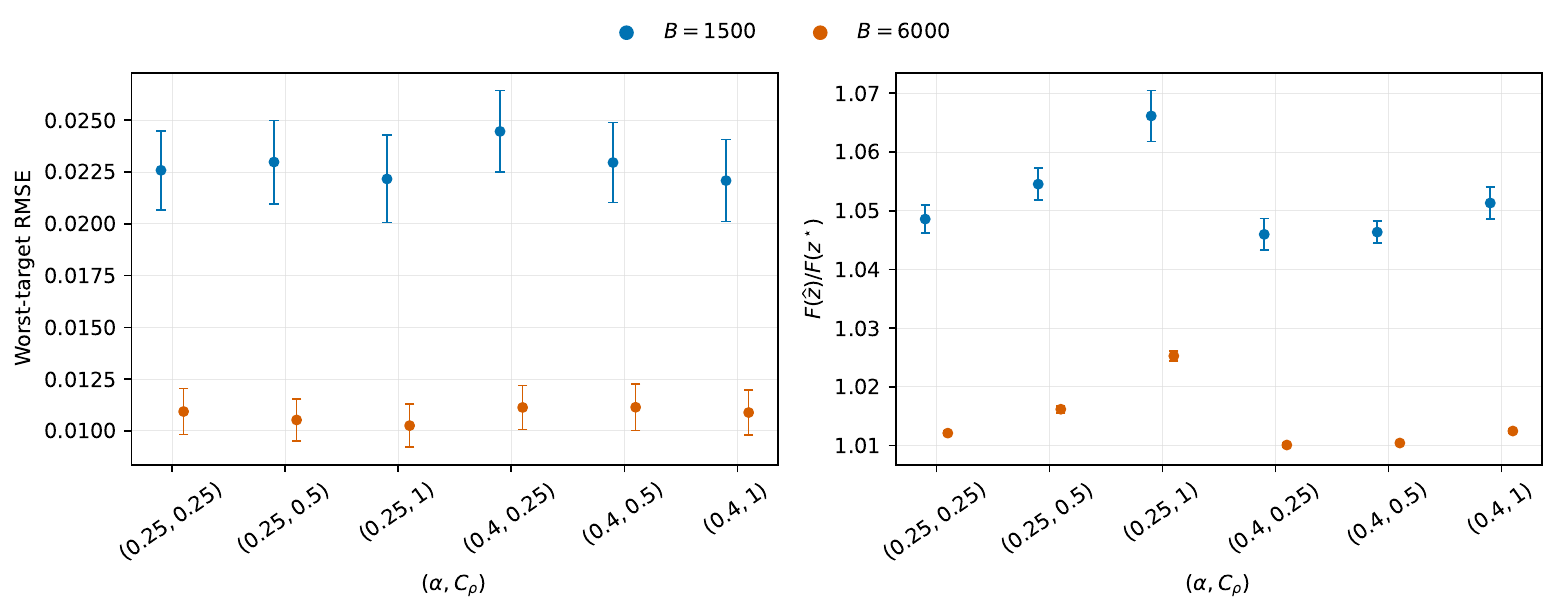}
    \caption{
Sensitivity of Adaptive Greedy to $\alpha$ and $\rho_0$, with $R=50$ replications per setting.
Error bars show $\pm1$ Monte Carlo standard error.
}
    \label{fig-app-parameter-sensitivity}
\end{figure}

We run Adaptive Greedy in Environment A with $\alpha\in\{0.25,0.4\}$ and $\rho_0\in\{0.25,0.5,1\}$,
at budgets $1500$ and $6000$ and with $R=50$ replications per setting. Over the six settings, the
worst-target RMSE at $B=1500$ lies between $0.0221$ and $0.0245$, and the criterion ratio at
$B=6000$ lies between $1.010$ and $1.025$ (Figure~\ref{fig-app-parameter-sensitivity}). The
procedure is therefore insensitive to these choices, and we use $\alpha=0.4$ and $\rho_0=0.5$ in all
other experiments without tuning.

\subsection{Inference}\label{app-inference}

\subsubsection*{Calibration}\label{app-inference-calibration}

For Environments A and B at budgets $6000$ and $24000$, the $200$ main replications are combined
with $300$ further replications drawn with different seeds, giving $R=500$ independent replications
for each environment and budget. Pointwise intervals use the normal $0.975$ quantile, and
simultaneous intervals use the $0.95$ quantile of the largest absolute coordinate of a normal vector
with the estimated correlation matrix, computed from $20000$ draws. Over the four settings, the
empirical standard deviations range from $0.0038$ to $0.0113$ and the mean estimated standard errors
from $0.0039$ to $0.0116$. Their ratio lies between $0.989$ and $1.060$, pointwise coverage lies
between $0.944$ and $0.966$, and simultaneous coverage lies between $0.950$ and $0.960$
(Tables~\ref{tab-app-inference-target} and \ref{tab-app-inference-simultaneous}).

\begin{table}[t]
\centering
\small
\caption{
Target-specific calibration of the Wald intervals, with $500$ replications per environment and
budget. The last column is the Monte Carlo standard error of the coverage.
}
\label{tab-app-inference-target}
\begin{tabular}{lrrrrrrr}
\toprule
Environment & Budget & Target & Empirical SD & Mean SE & SE/SD & Coverage & MCSE \\
\midrule
A & 6000 & 1 & 0.008227 & 0.008401 & 1.021 & 0.964 & 0.0083 \\
A & 6000 & 2 & 0.008563 & 0.008610 & 1.005 & 0.944 & 0.0103 \\
A & 6000 & 3 & 0.010898 & 0.011053 & 1.014 & 0.948 & 0.0099 \\
A & 6000 & 4 & 0.010511 & 0.010655 & 1.014 & 0.952 & 0.0096 \\
A & 24000 & 1 & 0.004153 & 0.004251 & 1.023 & 0.944 & 0.0103 \\
A & 24000 & 2 & 0.004227 & 0.004403 & 1.042 & 0.958 & 0.0090 \\
A & 24000 & 3 & 0.005463 & 0.005549 & 1.016 & 0.944 & 0.0103 \\
A & 24000 & 4 & 0.005107 & 0.005412 & 1.060 & 0.958 & 0.0090 \\
B & 6000 & 1 & 0.008090 & 0.008204 & 1.014 & 0.966 & 0.0081 \\
B & 6000 & 2 & 0.007739 & 0.007806 & 1.009 & 0.956 & 0.0092 \\
B & 6000 & 3 & 0.011265 & 0.011459 & 1.017 & 0.956 & 0.0092 \\
B & 6000 & 4 & 0.011278 & 0.011575 & 1.026 & 0.954 & 0.0094 \\
B & 24000 & 1 & 0.004091 & 0.004132 & 1.010 & 0.950 & 0.0097 \\
B & 24000 & 2 & 0.003789 & 0.003922 & 1.035 & 0.948 & 0.0099 \\
B & 24000 & 3 & 0.005846 & 0.005784 & 0.989 & 0.956 & 0.0092 \\
B & 24000 & 4 & 0.005648 & 0.005797 & 1.026 & 0.950 & 0.0097 \\
\bottomrule
\end{tabular}
\end{table}

\begin{table}[t]
\centering
\small
\caption{Simultaneous $95\%$ coverage of the Wald intervals.}
\label{tab-app-inference-simultaneous}
\begin{tabular}{lrrr}
\toprule
Environment & Budget & Coverage & MCSE \\
\midrule
A & 6000 & 0.956 & 0.0092 \\
A & 24000 & 0.950 & 0.0097 \\
B & 6000 & 0.960 & 0.0088 \\
B & 24000 & 0.958 & 0.0090 \\
\bottomrule
\end{tabular}
\end{table}

\subsubsection*{Pooled and logger-specific estimators}\label{app-pooled-local}

\begin{figure}[t]
    \centering
    \includegraphics[width=0.8\textwidth]
    {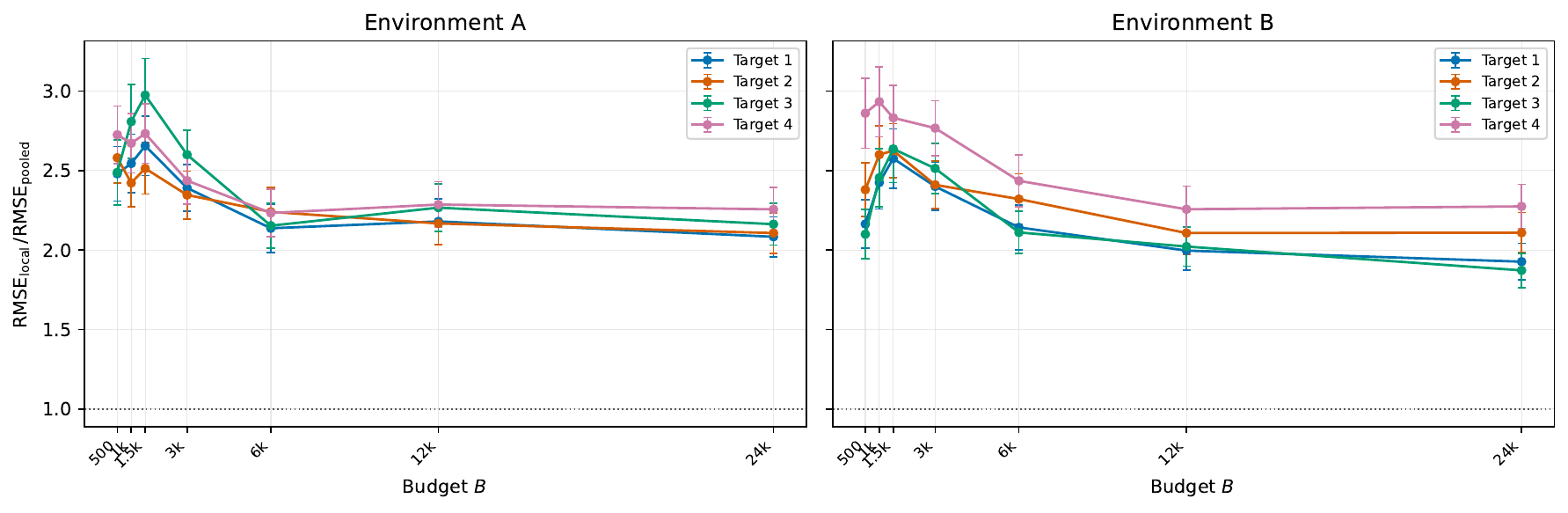}
    \caption{
Ratio of the RMSE of the logger-specific estimator to that of the pooled estimator in Environments
A and B. Values above one favor the pooled estimator. Error bars show $\pm1$ Monte Carlo standard
error.
}
    \label{fig-app-pooled-local}
\end{figure}

By \eqref{eq:aggregate-local-order}, the pooled estimator has a fixed-design covariance no larger
than that of the logger-specific estimator. To measure this difference, we apply both
estimators to the same purchased samples. Figure~\ref{fig-app-pooled-local} shows that the
logger-specific estimator has a larger RMSE in every reported setting.

\subsubsection*{Block-split estimator}\label{app-blocksplit}

We also evaluate the block-split estimator of Appendix~\ref{app:blocksplit} in Environment A, with
budgets $1500$ and $6000$ and $R=200$ replications. The first $\bar n=n_0=20$ observations form the first
block and are not scored, and the following blocks have lengths $20,40,80,\ldots$, so that
$\varkappa=2$. The regression for each block is fitted on all earlier observations, and each block
uses its own logger frequencies in the denominator. Table~\ref{tab-blocksplit} shows that the
block-split estimator is less accurate at $B=1500$ and that the gap narrows at $B=6000$, which
reflects the observations it discards and the less accurate regressions it uses in early blocks.

\begin{table}[t]
\centering
\caption{Worst-target RMSE of the final and block-split estimators in Environment A.}
\label{tab-blocksplit}
\begin{tabular}{rcc}
\toprule
Budget & Final estimator & Block-split estimator \\
\midrule
1500 & 0.02218 & 0.03898 \\
6000 & 0.01076 & 0.01353 \\
\bottomrule
\end{tabular}
\end{table}

\subsection{Intermediate Budgets and Runtime}\label{app-anytime-details}

The trajectory experiment uses Environment A with the standard initialization, budget $5000$, and
$R=50$ replications, and it records the design at the intermediate budgets $200$, $500$, $700$, $1000$,
$2000$, $3000$, $4000$, and $5000$. Adaptive Greedy and Reoptimized plug-in use the same random
numbers in each replication. The index of Adaptive Greedy does not involve the budget, since its
smoothing parameter depends only on the money spent so far. The budget affects only which loggers are
affordable and when the run stops. However, Reoptimized plug-in plans its total counts for the
budget, which is $5000$ throughout this experiment.

Table~\ref{tab-trajectory} reports the design diagnostics of Adaptive Greedy along these runs. The
difference in $\mathcal G/\mathcal G^\star$ between Adaptive Greedy and Reoptimized plug-in is negative at every
intermediate budget up to $4000$. At $5000$ it is $0.000847$, with interval $[0.000126,0.001567]$, so the
two procedures end at nearly the same criterion, with Reoptimized plug-in slightly closer.

\begin{table}[t]
\centering
\caption{Mean design diagnostics of Adaptive Greedy along runs with budget $5000$ in Environment A,
over $50$ replications.}
\label{tab-trajectory}
\begin{tabular}{rcc}
\toprule
Intermediate budget & $\mathcal G/\mathcal G^\star$ & $\|\theta-\theta^\star\|_2$ \\
\midrule
200  & 1.538371 & 0.464505 \\
500  & 1.179021 & 0.142007 \\
700  & 1.138549 & 0.125322 \\
1000 & 1.092325 & 0.101177 \\
2000 & 1.036573 & 0.062111 \\
3000 & 1.023061 & 0.048914 \\
4000 & 1.017046 & 0.040749 \\
5000 & 1.013295 & 0.034949 \\
\bottomrule
\end{tabular}
\end{table}

Table~\ref{tab-horizon-overlap} compares these runs with the separate runs of
Section~\ref{sec:experiments}, which use $R=200$ replications and a budget equal to the
intermediate budget. The Adaptive Greedy values in the two experiments agree closely, while the Reoptimized
plug-in values depend strongly on whether it plans for $5000$ or for the intermediate budget. When each
procedure is given the intermediate budget as its total budget, both reach similar criterion ratios, so the
difference between the two trajectories comes from the budget toward which each procedure plans.

\begin{table}[t]
\centering
\caption{
Criterion ratio $\mathcal G/\mathcal G^\star$ at three intermediate budgets. The trajectory columns follow runs with budget
$5000$, and the separate-run columns report separate runs whose budget equals the intermediate budget.
}
\label{tab-horizon-overlap}
\setlength{\tabcolsep}{5pt}
\begin{tabular}{rcccc}
\toprule
Intermediate budget &
\shortstack{Adaptive Greedy\\trajectory} &
\shortstack{Adaptive Greedy\\separate run} &
\shortstack{Reoptimized\\trajectory} &
\shortstack{Reoptimized\\separate run} \\
\midrule
500  & 1.179021 & 1.187451 & 1.414239 & 1.186214 \\
1000 & 1.092325 & 1.090244 & 1.722523 & 1.105435 \\
3000 & 1.023061 & 1.022266 & 1.285166 & 1.023932 \\
\bottomrule
\end{tabular}
\end{table}

The rise in the criterion ratio of Reoptimized plug-in early in the run occurs in almost every
replication. It increases between intermediate budgets $500$ and $700$ in $96\%$ of replications, and again
between $700$ and $1000$ in $96\%$, with mean increases of about $0.142$ and $0.166$. At these
budgets the procedure is still following plans made for budget $5000$. Its plans also change
substantially between early updates, which suggests that unstable early estimates contribute as
well.

\subsubsection*{Balanced costs}\label{app-balanced-horizon}

Under the costs $c=(1,2,4,8,16)$, the plan of Reoptimized plug-in for budget $5000$ is heavily
concentrated on Logger 1, and a partial run that follows it can be far from the plan for a smaller
budget. To check this explanation, we keep the statistical model and target policies of
Environment A and change only the costs to $c=(1,1,1.4,2.85,1.8)$, chosen so that the optimal
fractions are roughly equal over the four useful loggers,
$\theta^\star=(0.28243,0.22456,0.24673,0.24628,0)$.

Two findings support the explanation. First, consider a path computed with the true reward moments
that plans for budget $5000$. The ratio of its criterion at intermediate budget $B$ to that of the plan made
for budget $B$ is $1.018$, $1.009$, and $1.004$ at $B=500$, $700$, and $1000$ under the balanced
costs, compared with $1.569$, $1.778$, and $1.983$ under the original costs. Second,
Table~\ref{tab-balanced-horizon} shows that under the original costs the mean $\mathcal G/\mathcal G^\star$ of
Reoptimized plug-in rises from $1.414$ at $B=500$ to $1.723$ at $B=1000$, whereas under the
balanced costs it falls from $1.323$ to $1.213$, and its distance to $\theta^\star$ also decreases.
At $B=5000$, both runs are close to their optimal designs.

\begin{table}[t]
\centering
\small
\caption{
Plans for each intermediate budget and mean counts purchased by Reoptimized plug-in along runs with
budget $5000$, under the original and the balanced costs. The plans are deterministic, and the
purchased counts are means over $50$ replications. $\mathcal G/\mathcal G^\star$ and $\|\theta-\theta^\star\|_2$ use
the optimal design for the same costs.
}
\label{tab-balanced-horizon}
\setlength{\tabcolsep}{3pt}
\resizebox{\textwidth}{!}{
\begin{tabular}{rcccc}
\toprule
Intermediate budget & Plan for intermediate budget & Mean purchased counts & Mean $\mathcal G/\mathcal G^\star$ &
Mean $\|\theta-\theta^\star\|_2$ \\
\midrule
\multicolumn{5}{l}{\textbf{Original costs $c=(1,2,4,8,16)$}}\\
500  & $(168,4,15,25,4)$ & $(295.2,7.1,6.8,12.4,4.0)$ & 1.414 & 0.169 \\
700  & $(240,4,23,37,4)$ & $(488.7,8.5,6.8,12.8,4.0)$ & 1.556 & 0.205 \\
1000 & $(348,4,33,56,4)$ & $(782.7,8.5,6.9,13.6,4.0)$ & 1.723 & 0.225 \\
2000 & $(708,4,73,116,4)$ & $(1534.8,10.3,18.0,38.3,4.0)$ & 1.593 & 0.216 \\
3000 & $(1068,4,111,177,4)$ & $(1664.7,11.0,37.9,136.9,4.0)$ & 1.285 & 0.144 \\
4000 & $(1428,4,151,237,4)$ & $(1747.2,12.8,98.8,220.7,4.0)$ & 1.071 & 0.073 \\
5000 & $(1788,4,189,298,4)$ & $(1807.6,21.2,178.1,296.7,4.0)$ & 1.012 & 0.024 \\
\midrule
\multicolumn{5}{l}{\textbf{Balanced costs $c=(1,1,1.4,2.85,1.8)$}}\\
500  & $(89,75,80,76,4)$ & $(66.1,205.5,80.6,35.9,7.0)$ & 1.323 & 0.501 \\
700  & $(126,102,112,108,4)$ & $(112.0,257.1,111.1,56.9,7.0)$ & 1.282 & 0.446 \\
1000 & $(180,147,160,155,4)$ & $(180.9,290.7,166.5,98.9,7.0)$ & 1.213 & 0.349 \\
2000 & $(361,290,319,314,4)$ & $(408.7,347.9,311.2,278.6,7.0)$ & 1.028 & 0.149 \\
3000 & $(543,435,477,472,5)$ & $(583.9,449.3,459.8,459.4,7.0)$ & 1.004 & 0.070 \\
4000 & $(725,579,636,631,4)$ & $(745.2,588.5,610.9,630.8,7.0)$ & 1.001 & 0.041 \\
5000 & $(907,724,795,789,4)$ & $(910.3,733.6,774.6,792.6,7.0)$ & 1.001 & 0.027 \\
\bottomrule
\end{tabular}
}
\end{table}

\subsubsection*{Runtime}\label{app-runtime}

The runtimes refer to our implementation and are meant for comparison between procedures. In
Environment A, the mean runtime at $B=12000$ is about $742$ seconds for Adaptive Greedy and $6423$
seconds for Reoptimized plug-in, and at $B=24000$ about $1750$ and $16442$ seconds. The difference
comes mainly from the repeated reoptimization of Reoptimized plug-in.

\subsection{ACTG 175 Construction}\label{app-actg-details}

The ACTG 175 study \citep{hammer1996} was a randomized trial comparing nucleoside treatment regimens
in HIV-infected adults. We use the public data \citep{hammer1996uci}, which contain $2139$ subjects
with complete records. The actions are the four treatments, coded $0$ for zidovudine (ZDV), $1$ for
ZDV with didanosine (ddI), $2$ for ZDV with zalcitabine, and $3$ for ddI alone. The context is
\[
X=(\text{age},\text{wtkg},\text{cd40},\text{cd80},\text{karnof},\text{symptom},\text{str2})^\top,
\]
which records age, weight, baseline CD4 and CD8 counts, Karnofsky score, symptomatic status, and
prior antiretroviral therapy. Continuous covariates are standardized and binary covariates are
centered, using the means and standard deviations of the $2139$ subjects.

The response used to build the reward model is $Y=(\text{cd420}-\text{cd40})/100$, the change in
CD4 count from baseline to around week $20$. For each treatment $a$ we fit a least-squares
regression of $Y$ on the basis $\phi(x)=(1,x_1,\ldots,x_7,x_3^2,x_1x_3)^\top$, giving a fitted mean
$\widehat m_a(x)$ and a residual variance $\widehat s_a^2$. In the experiment, contexts are drawn
with replacement from the $2139$ subjects, and the reward is
\[
R=\min\{4,\max(-4,R^\circ)\},\qquad R^\circ=\widehat m_A(X)+\varepsilon_A,
\qquad\varepsilon_a\sim N(0,\widehat s_a^2).
\]
The design calculations use the exact moments of the clipped normal distribution. This
construction keeps the observed covariate distribution and the fitted treatment effects while
making the policy values and the optimal design known.

\subsubsection*{Loggers}

Let $s=(x_3,x_7,x_6)^\top$ and
\[
D=
\begin{pmatrix}
0.35 & -0.45 & -0.30\\
-0.25 & 0.20 & 0.25\\
-0.25 & 0.15 & 0.25\\
0.15 & 0.10 & -0.20
\end{pmatrix},
\qquad
B=
\begin{pmatrix}
1 & 1 & -1 & -1\\
-1 & 1 & 1 & -1\\
-1 & -1 & 1 & 1\\
1 & -1 & -1 & 1
\end{pmatrix}.
\]
For $k=1,\ldots,4$, Logger $k$ is the softmax policy with logits $B_{k,a}+(Ds)_a$, mixed with the
uniform distribution with weight $\epsilon_k$, where
$(\epsilon_1,\epsilon_2,\epsilon_3,\epsilon_4)=(0.05,0.05,0.08,0.08)$. Logger 5 is uniform, and the
costs are $c=(1,2,4,3,8)$. Averaged over the $2139$ subjects, the loggers choose the treatments with
the following probabilities.
\[
\begin{array}{c|cccc}
& a=0 & a=1 & a=2 & a=3\\
\hline
\text{Logger 1} & 0.433 & 0.431 & 0.069 & 0.066\\
\text{Logger 2} & 0.082 & 0.423 & 0.423 & 0.072\\
\text{Logger 3} & 0.081 & 0.075 & 0.424 & 0.420\\
\text{Logger 4} & 0.424 & 0.081 & 0.080 & 0.416\\
\text{Logger 5} & 0.250 & 0.250 & 0.250 & 0.250
\end{array}
\]
Loggers 1 to 4 therefore cover different pairs of treatments. The costs are units chosen for the
experiment and are not estimates of clinical recruitment costs.

\subsubsection*{Target policies}

The first target policy favors treatments with higher fitted rewards. With $q_{\mathrm{clip}}(x,a)$
the mean of the clipped reward and $\tau_1$ equal to half the standard deviation of
$q_{\mathrm{clip}}(X,a)$ over subjects and treatments, it is
\[
\pi_1(a\mid x)=0.9\,\frac{\exp\{q_{\mathrm{clip}}(x,a)/\tau_1\}}{\sum_b\exp\{q_{\mathrm{clip}}(x,b)/\tau_1\}}
+\frac{0.1}{4}.
\]
For $j=2,3,4$, target $j$ is the softmax policy with logits $b_{j,a}+\nu_j(Ds)_a$, where
$b_2=(-0.6,0.6,0.6,-0.6)$, $b_3=(-0.9,0.3,-0.3,0.9)$, $b_4=(0.7,0.3,-0.5,-0.5)$, and
$\nu=(1,0.7,1.2)$, mixed with the uniform distribution with floors $(0.06,0.05,0.05)$. The optimal
design is
\[
z^\star=(0.576469,0,0.020001,0.114509,0),\qquad
\theta^\star=(0.810810,0,0.028132,0.161058,0),
\]
with $\mathcal G^\star=4.237904$, and targets 1 and 3 attain the maximum at this design.

\subsubsection*{Cost-based designs}

With $c=(1,2,4,3,8)$, Equal expenditure has limiting fractions
$\theta_{\mathrm{eq}}\propto(1,1/2,1/4,1/3,1/8)$, that is,
$\theta_{\mathrm{eq}}=(0.452830,0.226415,0.113208,0.150943,0.056604)$, which differ from the equal
counts $(0.2,0.2,0.2,0.2,0.2)$. With $z_\theta=\theta/c^\top\theta$ the design with sample fractions
$\theta$, their criterion ratios are
$\mathcal G(z_{\theta_{\mathrm{count}}})/\mathcal G^\star=2.499102$ and $\mathcal G(z_{\theta_{\mathrm{eq}}})/\mathcal G^\star=1.404870$, and
purchasing only from the cheapest logger gives $\mathcal G(z_{\theta_{\mathrm{cheap}}})/\mathcal G^\star=1.213666$.

\subsection{Additional ACTG 175 Results}\label{app-actg-results}

Oracle greedy, which uses the true gains, has worst-target RMSE ($\times10^{-3}$) of $120$, $55.9$,
$28.2$, and $18.6$ at budgets $500$, $1500$, $6000$, and $12000$, with Monte Carlo standard errors
$8.7$, $2.5$, $1.4$, and $0.83$. Table~\ref{tab-actg-allocation} reports the mean fractions of
Adaptive Greedy. As the budget grows, the mass on Loggers 2 and 5 decreases toward zero and the
fractions of Loggers 1, 3, and 4 approach their optimal values. At $B=12000$, the mean runtime is
about $93$ seconds for Adaptive Greedy and $705$ seconds for Reoptimized plug-in. Because the
rewards are simulated from a fitted model, these results describe the purchasing problem and should
not be read as a comparison of the clinical efficacy of the treatments.

\begin{table}[t]
\centering
\caption{Mean fractions and design diagnostics of Adaptive Greedy in the ACTG application.}
\label{tab-actg-allocation}
\setlength{\tabcolsep}{4pt}
\begin{tabular}{rccccccc}
\toprule
$B$ & $\theta_1$ & $\theta_2$ & $\theta_3$ & $\theta_4$ & $\theta_5$ & $\mathcal G/\mathcal G^\star$ &
$\|\theta-\theta^\star\|_2$ \\
\midrule
500   & 0.7957 & 0.0408 & 0.0130 & 0.1390 & 0.0115 & 1.11505 & 0.10277 \\
1500  & 0.7883 & 0.0269 & 0.0081 & 0.1729 & 0.0038 & 1.04539 & 0.06461 \\
6000  & 0.8046 & 0.0107 & 0.0169 & 0.1669 & 0.0009 & 1.01680 & 0.03043 \\
12000 & 0.8086 & 0.0057 & 0.0218 & 0.1634 & 0.0005 & 1.01045 & 0.01994 \\
\bottomrule
\end{tabular}
\end{table}

\end{document}